\documentclass{article} 
\usepackage[preprint]{colm2026_conference}

\usepackage{microtype}
\usepackage{hyperref}
\usepackage{url}
\usepackage{booktabs}
\usepackage{soul}
\usepackage{enumitem}
\usepackage{wrapfig}
\usepackage{graphicx}
\usepackage{amsmath}
\usepackage{amssymb}
\usepackage{tabularx}
\usepackage{threeparttable}
\usepackage{lineno}
\usepackage{longtable}
\usepackage{rotating}
\usepackage[dvipsnames]{xcolor}
\usepackage{subcaption}
\usepackage{multirow}
\usepackage{makecell}
\definecolor{darkblue}{rgb}{0, 0, 0.5}
\hypersetup{colorlinks=true, citecolor=darkblue, linkcolor=darkblue, urlcolor=darkblue}
\usepackage[most]{tcolorbox}
\usepackage{graphicx}
\usepackage{color-edits}
\addauthor{sw}{blue}
\addauthor{qtosw}{orange}
\addauthor{ek}{purple}

\title{``I Didn't Make the Micro Decisions'': Measuring, Inducing, and Exposing Goal-Level AI Contributions in Collaboration
}

\author{%
Eunsu Kim$^{1}$\thanks{Work done while Eunsu and Kyungjin were visiting Carnegie Mellon University.} ,
Jessica R. Mindel$^{2}$, Kyungjin Kim$^{3}$\footnotemark[1] , Sherry Tongshuang Wu$^{2}$\\
  $^1$KAIST, $^2$Carnegie Mellon University, $^3$Seoul National University
\\
\texttt{\{eunsukim, sherryw\}@andrew.cmu.edu} \\ 
}

\newcommand{\rebuttal}[2]{#1}

\newcommand{\torevise}[1]{\textcolor{red}{Revise: #1}}
\newcommand{\toadd}[1]{\textcolor{red}{Add: #1}}

\newcommand{\framework}{{\textsc{CoTrace}}}

\newcommand{\promptbox}[2]{
\begin{tcolorbox}[
title=#1,
breakable,
colback=gray!5,
colframe=black!60,
fontupper=\ttfamily\small,
fonttitle=\bfseries
]
#2
\end{tcolorbox}
}

\begin{document}

\ifcolmsubmission
\linenumbers
\fi

\maketitle

\begin{abstract}
As large language models (LLMs) increasingly shape how users form, refine, and extend their goals, attributing contributions in human--AI collaboration becomes critical---both for users calibrating their own reliance and for evaluators assessing AI-assisted work. Yet existing methods focus on final artifacts, missing the process through which goals themselves are jointly shaped. We introduce a goal-level attribution framework, \framework, that decomposes explicit goals into verifiable requirements and traces both direct contributions and indirect influences across dialogue turns. 
Applying \framework{} to 638 real-world collaboration logs, we find that while models account for only 11--26\% of goal-shaping contribution, they contribute substantially more on introducing lower-level concrete requirements, and make various kinds of indirect contributions. Through controlled simulations, we show that interaction design choices significantly affect model goal-shaping behavior. In a user study, exposing participants to goal-level analyses shifts their perceived contributions by nearly 2 points on a 5-point scale, revealing systematic miscalibration in how users understand their own AI-assisted work.
\footnote{The code for \framework{} is available at \url{https://github.com/rladmstn1714/CoTrace}.}
\end{abstract}
\section{Introduction}
\label{sec:intro}
Consider a student who develops an essay argument with an LLM over twenty turns of dialogue. The final text may appear entirely “human,” with the student typing every word, yet the AI proposed the central thesis and restructured key paragraphs. By contrast, another student might dictate every goal and constraint across a hundred turns, using the model merely as a typist. The AI in these two cases clearly should receive different \textbf{contribution attribution}, but today, an instructor grading the work, a reviewer assessing originality, or even the students themselves have no way to tell the difference. As AI is deployed across educational and professional settings, this gap carries real consequences: users need to understand and calibrate their own reliance on AI~\citep{draxler2024ai}, and evaluators and institutions need evidence-based grounds for assessing AI-assisted work.

Despite this need, no existing framework can adequately distinguish such cases. Current attribution tools (e.g., text watermarking, stylometric analysis, or turn-level authorship tracking~\citep{siddiqui2025draftmarks,liang2024watermarkingtechniques,kumarage2023stylometricdetection}) are outcome-oriented, focusing almost exclusively on detecting AI involvement in the final artifact. But as LLMs become more capable, they increasingly do more than execute instructions---they propose directions, refine constraints, introduce structure, and make concrete design decisions that users may not have considered~\citep{kim2026discoverllm, shen2025completion}. In many cases, users welcome this initiative; in others, they may want tighter control over how much the AI shapes their goals versus simply carrying them out~\citep{shneiderman2022hcai, shao2025future, feng2025levels}. But the degree of AI initiative in any given collaboration is currently invisible, both to users and external evaluators. Without process-level measurement, we cannot evaluate how much autonomy models are actually exercising, design interventions that keep AI initiative appropriately bounded for a given context, or help users calibrate their awareness of AI contributions.


\begin{figure}[t]
    \centering
    \includegraphics[width=\columnwidth]{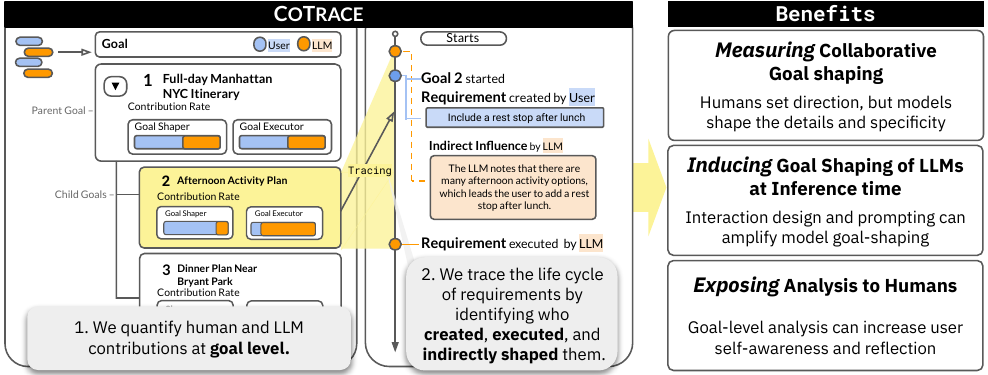}
    \caption{\textbf{Illustrative overview of \framework{} and its benefits.} \framework{} analyzes human and LLM contributions at the goal level by tracing requirement lifecycles, including direct contributions (who explicitly creates a requirement) and indirect contributions (who influences another party to introduce it). It supports \textit{measuring} goal-shaping behavior, provides a signal for \textit{inducing} it, and supports user awareness by \textit{exposing} contribution dynamics.}
    \vspace{-7mm}
    \label{fig:intro}
\end{figure}


To address this gap, we introduce \framework{}, an automated framework that measures human and AI contributions throughout the collaboration process. Rather than analyzing only outputs, we track the process of creating, refining, and executing tasks across dialogue to produce a principled, quantitative account of each party’s influence.
We center our analysis on \emph{task goals}--explicit, actionable targets with a desired outcome--which we decompose into granular, verifiable requirements~\citep{qin2024infobench, viswanathan2025checklists}. 
This structure links artifacts to conversation: outcomes capture what is being built, while requirements are granular enough to trace back to specific utterances where concrete decisions occur. 
Critically, we capture not only direct contributions, where one party explicitly creates or modifies a requirement, but also \textit{indirect influence}, where one party’s action provides context that leads the other to formulate a new requirement (e.g., a clarifying question, draft artifact, or exposed error); this is a common but often less visible form of AI contribution~\citep{kim2026discoverllm,he2025contributionsdeserve}, which users may fail to recognize without explicit analysis. 

We demonstrate how \framework{} provides value through three complementary studies:

\begin{itemize}[itemsep=1pt, topsep=1pt, parsep=0pt, partopsep=0pt,leftmargin=*]
 
\item \textbf{As an \emph{evaluation suite}: measuring collaborative goal shaping in the wild (\S\ref{sec:exp}).} We apply \framework{} to real-world human-LLM collaboration logs across four domains. We find that while models appear to follow user direction at the macro level, they play a larger role in shaping specific requirements, especially in technical tasks. After the initial turns, many requirements emerge through mutual influence rather than user initiative alone, and we identify 11 recurring interaction patterns by which indirect goal shaping occurs.
 
\item \textbf{As a \emph{design tool}: supporting inference-time intervention and evaluation (\S\ref{sec:exp/simulation}).} Through controlled simulations, we show that interaction design choices (such as whether an agent must communicate before acting) and prompting strategies (such as underspecification) substantially affect model goal-shaping behavior, suggesting actionable design levers for manipulating how actively models contribute to goals.
 
\item \textbf{As a \emph{reflection tool}: improving user awareness and intentionality (\S\ref{sec:human_study}).} We build and open-source \framework-viewer, an interactive analytical tool that makes contribution dynamics legible. In a user study with 10 participants, we find that exposing participants to goal-level analysis shifts their perception of both their own and the AI's contributions. Some reflected on changing their prompting practices, suggesting that the tool corrects miscalibrated perceptions and promotes more intentional collaboration with AI.
 
\end{itemize}

Together, our research establishes a foundation for principled attribution in settings where AI-assisted work is evaluated, credited, or regulated, providing both the measurement infrastructure and the empirical grounding that such decisions currently lack.

\section{\framework: Evaluation Framework for Quantifying Agents’ Goal-Level Contributions in Human–AI Collaboration}
\label{sec:framework}
We propose \framework{}, a \textbf{Goal-Level Attribution Framework for Human--LLM Collaboration}, built around two core design choices (Figure~\ref{fig:intro}).

\textbf{Desideratum 1: Goals as Outcome--Requirement Tuples.} Our unit of analysis is the goal: an explicit, actionable target represented by a desired outcome and a set of requirements~\citep{locke2002goal}. The outcome captures what the collaboration aims to produce, such as ``a full-day Manhattan NYC itinerary,'' while the requirements specify the smallest independently checkable success predicates that the outcome should satisfy~\citep{qin2024infobench,viswanathan2025checklists}. 
Formally, we represent each goal as $g=(o, R)$, where $o$ denotes the desired outcome and $R=\{r_1,\ldots,r_k\}$ denotes its associated requirements.

We adopt goals as the central unit because collaboration unfolds through the evolution of desired outcomes over time, not only through the final artifact. 
Since goals in human--LLM collaboration are often underspecified, the outcome--requirement representation makes goal shaping observable and evaluable at a granular level. 
Following \cite{kim2026discoverllm}, we also organize goals hierarchically according to their level of specificity into \textsc{Parent goals} (the overall objective, e.g., ``full-day Manhattan NYC itinerary'') and \textsc{Child goals} (specific sub-tasks, e.g., ``afternoon activity plan''), both eventually linked to individual requirements (e.g., ``include a rest stop after lunch''), as shown in Figure~\ref{fig:intro}. 

This structure also allows us to \emph{link artifacts to conversation}: goals capture what is being built, while requirements are granular enough to connect to specific utterances where concrete design decisions occur. 
We do so by decomposing each utterance into atomic \textsc{Actions} -- the minimal communicative units a speaker performs in a turn (e.g., requesting, constraining, providing code). Each action also serves as the unit for requirement iteration.
Detailed background and rationale are provided in Appendix~\ref{appendix:bg_goal_def}.


\textbf{Desideratum 2: Direct and Indirect Influence.} We model goal shaping not as a single creation event, but as a cumulative result of preceding actions in the interaction. Accordingly, we distinguish between direct goal shaping (an action explicitly introduces or modifies a requirement) and potential indirect influence (an action provides context that later motivates a requirement). \emph{Indirect influence} captures many more common and realistic scenarios than the status-quo, especially when the AI plants a seed (e.g., asking a clarifying question, proposing an analogy) that the human then develops into a concrete requirement.

\paragraph{Pipeline Overview.} We operationalize \framework{} as an automated pipeline using LLMs-as-judges, 
consisting of four stages (Figure~\ref{fig:pipeline} in Appendix~\ref{appendix:framework}):

\begin{enumerate}[itemsep=1pt, topsep=1pt, parsep=0pt, partopsep=0pt,leftmargin=*]
\item \emph{Outcome and Action Extraction.} The dialogue is segmented into  blocks of turns. An LLM identifies desired outcomes and decomposes each message  into atomic actions, each assigned a role: \textsc{Shaper} (proposes goals, ideas, or requirements), \textsc{Executor} (carries out actions or produces output), or \textsc{Other}.

\item \emph{Requirement Extraction.} For each outcome, requirements are 
extracted and linked to their origin and contributing actions, tracked through 
\textsc{Create}, \textsc{Revise}, and \textsc{Delete} operations, which yields a versioned history of the collaboration.

\item \emph{Influence Labeling.} Candidate action--requirement pairs are 
filtered by embedding similarity, then evaluated by an LLM-as-judge as 
\textsc{direct connection}, \textsc{implicit connection}, or 
\textsc{no connection}. These determine the influence score
$I(a\rightarrow r)$ used in our metrics.

\item \emph{Quantifying Contribution.} Influence scores are aggregated into 
contribution scores. In particular, the \textit{\textbf{role-level contribution}} of speaker $p$ to
requirement $r$ through role $\rho$ is
\[
M(p,\rho,r)=\sum_{a\in A_p}\mathbf{1}[\mathrm{role}(a)=\rho]\,I(a\rightarrow r)\]
We then aggregate these requirement-level scores to the goal
levels, yielding a speaker $\times$ role contribution matrix.
\end{enumerate}
Full implementation details and prompts are provided in Appendix~\ref{appendix:framework}.

\textbf{Validation.} We validate the framework in two ways: (1) manual validation on randomly sampled existing dialogues, and (2) participant validation in the user study, where participants review analyses of their own conversations. Across both validations, we evaluate three components separately: goal extraction, requirement extraction, and influence labeling. Manual validation achieves over 90\% accuracy, and participants rate the framework's alignment with their own perception above 4 out of 5 on average.
We provide validation details, error analyses, sensitivity analyses of the LLM-as-a-judge pipeline, and comparisons across different judge models in Appendix~\ref{appendix:framework/validation}.

We envision \framework{} as useful across a range of settings. In the following sections, we demonstrate three complementary uses: \emph{measuring} collaborative goal shaping (\S\ref{sec:exp}) through analysis of real-world human--AI logs across task types and goal specificity levels; \emph{inducing} goal-shaping behavior at inference time (\S\ref{sec:exp/simulation}) through interaction design choices that amplify or suppress model initiative; and \emph{exposing} these dynamics to users (\S\ref{sec:human_study}), improving awareness of AI contributions and prompting reflection on collaboration practices.


\section{Measuring Collaborative Goal Shaping In the Wild}
\label{sec:exp}

We apply \framework{} to real-world human-LLM collaboration logs and answer two questions: \emph{who} contributes to shaping which goals (\S\ref{sec:exp/who}), and \emph{how} goal shaping emerges through the interaction (\S\ref{sec:exp/how}). We additionally compare collaboration dynamics across system settings of model-only vs. agentic (\S\ref{sec:exp/system}).
 

\begin{wraptable}{r}{0.45\textwidth}
\vspace{-0.8\baselineskip}
\centering
\footnotesize
\setlength{\tabcolsep}{4pt}
\begin{tabular}{lrrrrr}

\toprule
\textbf{Task} &\#&\includegraphics[height=1em]{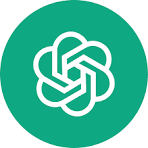} &
\includegraphics[height=1em]{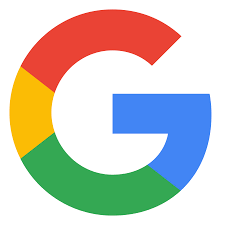} &
\includegraphics[height=1em]{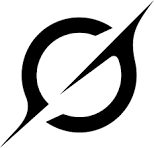} &
\includegraphics[height=1em]{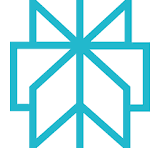} \\

\midrule
Comp.Prog.   & 177 & 115 & 26 & 32 & 4 \\
Data Analysis &  92 &  41 & 12 & 31 & 8 \\
Writing       & 293 & 114 & 57 & 95 & 27 \\
Planning      &  76 &  10 & 10 & 51 & 5 \\
\midrule
\textbf{Total} & \textbf{638} & \textbf{280} & \textbf{105} & \textbf{209} & \textbf{44} \\
\bottomrule
\end{tabular}
\caption{\textbf{Distribution of ShareChat logs used in our analysis.} }
\label{tab:sharechat_data_dist}
\vspace{-1\baselineskip}
\end{wraptable}

\textbf{Data.} 
We analyze ShareChat~\citep{yan2026sharechat}, a publicly available dataset of human-LLM interactions collected from five major LLM chat platforms: OpenAI \includegraphics[height=0.8em]{src/img/logo_openai.png}, Anthropic, Google \includegraphics[height=0.8em]{src/img/logo_google.png}, Grok \includegraphics[height=0.8em]{src/img/logo_grok.png}, and Perplexity \includegraphics[height=0.8em]{src/img/logo_perplexity.png}.\footnote{OpenAI logs include GPT-4/4o, while Google logs include Gemini Advanced, 2.0 Flash, 2.5 Pro, and 2.5 Flash. Model-level information is unavailable for Grok and Perplexity.} We focus on four task categories involving sustained collaboration: Computer Programming (Comp. Prog.), Data Analysis, Writing, and Planning. After filtering the data based on topic and our collaboration heuristics, we retain 638 logs for analysis (Table~\ref{tab:sharechat_data_dist}). Detailed data sampling and topic categorizing procedures, filtering criteria, and dataset examples for each task are provided in Appendix~\ref{appendix:data_sampling}.


\subsection{``Who'': Humans set direction, but models shape the details and specificity}
\label{sec:exp/who}
\begin{figure}[h]
    \centering

    \begin{subfigure}[t]{0.35\linewidth}
        \centering
        \includegraphics[width=\linewidth]{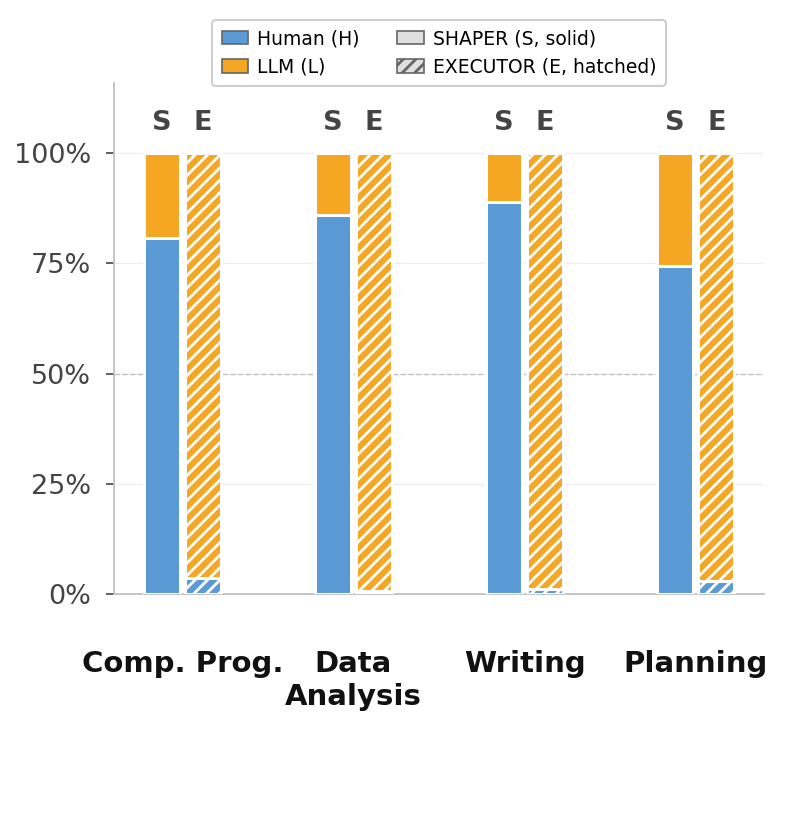}
        \vspace{-10mm}
        \caption{Role-level contribution.}
        \label{fig:overall_results_1}
    \end{subfigure}
    \hfill
    \begin{subfigure}[t]{0.64\linewidth}
        \centering
        \includegraphics[width=\linewidth]{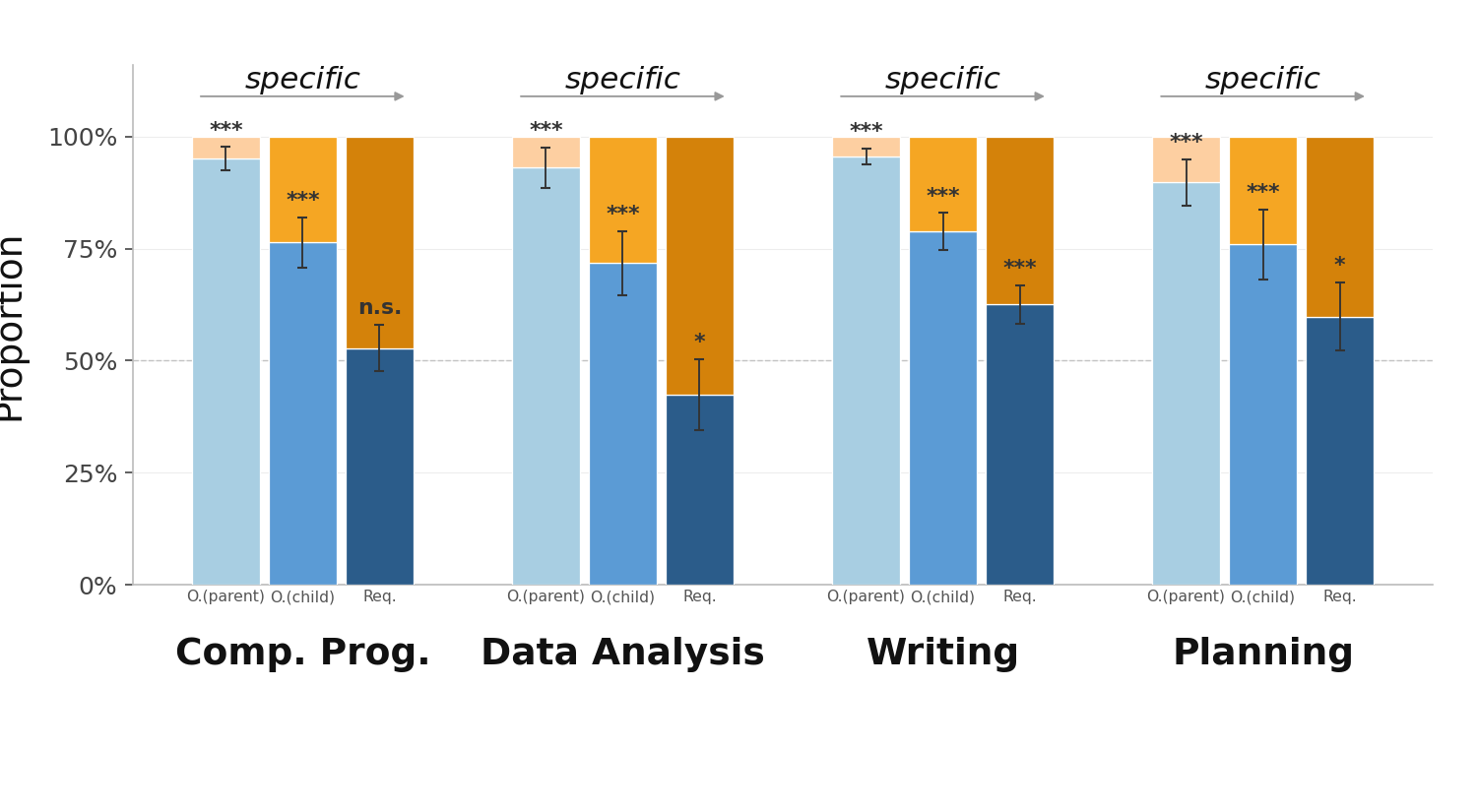}
        \vspace{-10mm}
        \caption{LLM's goal shaping by specificity level.}
        \label{fig:overall_results_2}
    \end{subfigure}
    \vspace{-5mm}
    \caption{\textbf{Overall goal shaping tendencies. }Humans {\color{NavyBlue} (H)} dominate overall shaping (a), while LLM {\color{Orange} (L)} contributions on goal shaping increase as goals become more specific (b). }
    \label{fig:overall_results}
\end{figure}
\textbf{Humans primarily set direction, while models add specificity.}
Figure~\ref{fig:overall_results} shows that humans dominate goal shaping across all four tasks: humans account for 75--89\% of all \textsc{shaper} mass while LLMs account for 96--99\% of all \textsc{executor} mass. This aligns with the instruction-following nature of current LLMs, which are typically guided by human-specified instructions~\citep{ouyang2022traininglanguagemodelsfollow}. However, a more nuanced pattern appears along the goal hierarchy (\S\ref{sec:framework}): Figure~\ref{fig:overall_results_2} shows that LLM contributions to goal shaping increase as goals become more specific. Models rarely shape parent outcomes, but contribute more to child outcomes and especially to individual requirements. Thus, models contribute less to setting overall direction than to elaborating subgoals and requirements. 

\textbf{Models show stronger goal-shaping behavior in technical, closed-ended tasks} than in non-technical, open-ended ones (Figure~\ref{fig:overall_req_results_1} and \ref{fig:overall_req_results_2}). In Computer Programming and Data Analysis, LLMs become increasingly active in generating requirements as interaction unfolds, eventually surpassing users in Data Analysis. In open-ended tasks, however, LLM contributions to goal shaping remain substantially lower ($p < .001$), while humans show the reverse pattern, contributing relatively more.

\textbf{Models can contribute implementation details that users rarely specify.}
Across tasks, models tend to introduce lower-level, implementation-oriented requirements (e.g., technical constraints, environmental assumptions, and correctness checks), whereas users more often contribute broader, goal-oriented ones. In technical tasks, some semantic clusters consist primarily of assistant-generated requirements, suggesting that models introduce requirement types users rarely specify themselves. In other domains, assistant-generated requirements largely overlap with user-generated ones (see Figure~\ref{fig:qual_req_embedding} and Appendix~\ref{appendix:experiment}).
\subsection{``How'': Goal shaping emerges through execution, not just explicit proposals}
\label{sec:exp/how}
\begin{figure}[ht]
\vspace{-5mm}
    \centering
        \includegraphics[width=\linewidth]{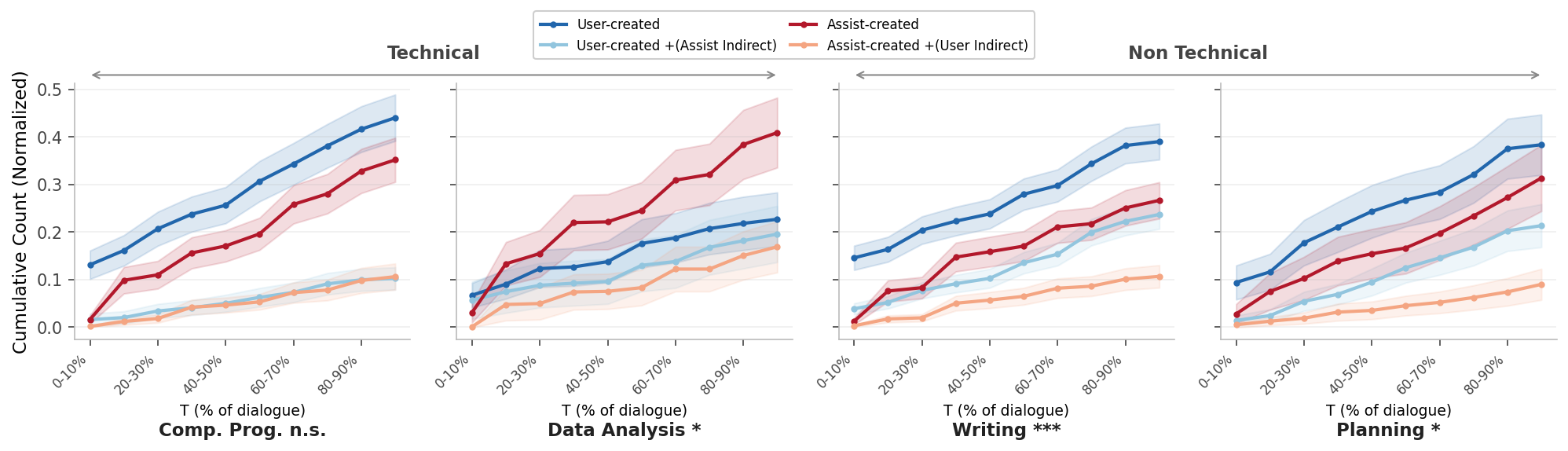}

\caption{\textbf{Impact of task on requirement generation.} Models generate more requirements in technical tasks than in less technical tasks such as writing and planning.}

    \label{fig:overall_req_results_1}

\end{figure}
Having established \emph{who} shapes goals, we now examine \emph{how}---through what kinds of actions, and through what patterns of mutual influence.

\textbf{Humans and LLMs jointly shape goals throughout the interaction.} Figure~\ref{fig:overall_req_results_1} shows how requirements accumulate over time. We group them into four categories based on who explicitly creates them (\textsc{direct}) and whether their creation is influenced by the other party (\textsc{indirect}): \textit{user-created}, \textit{user-created with assistant indirect influence}, \textit{assistant-created}, and \textit{assistant-created with user indirect influence}. After users introduce the initial requirements, \textit{user-created with assistant indirect influence} steadily increases, reflecting ongoing mutual influence between user and assistant. This pattern suggests that goal shaping in human--LLM collaboration is typically co-constructed rather than driven by the user alone, highlighting the value of tracking indirect influence in \framework{}.

\definecolor{cprepair}{HTML}{3182bd}
\definecolor{cpostpair}{HTML}{137333}

\newcommand{\prepair}[1]{{\color{cprepair}#1}}
\newcommand{\postpair}[1]{{\color{cpostpair}#1}}
\newcommand{\patterntitle}[1]{\textsc{\textbf{#1}}}

\setlength{\fboxsep}{1pt}
\newcommand{\softpair}[1]{\colorbox{gray!11}{\strut #1}}

\begin{table*}[t]
\centering
\fontsize{7.6}{8.2}\selectfont
\setlength{\tabcolsep}{4pt}
\begin{tabularx}{\textwidth}{p{2.05cm} X}
\toprule
\textbf{Field} & \textbf{Content} \\
\midrule

\multicolumn{2}{l}{\patterntitle{Underspecified Intent / Preference}} \\
 \quad Definition &
The influencer does not specify a concrete requirement directly, but provides a broad goal or implicit preference from which the creator infers and formulates a more explicit requirement. \\
 \quad Subtypes &
\softpair{\prepair{State Broad Goal} $\rightarrow$ \postpair{Derive Concrete Req.}}
\enspace/\enspace
\softpair{\prepair{Implicit Preferences}$\rightarrow$\postpair{Explicate Into Reqs.}} \\
\midrule

\multicolumn{2}{l}{\patterntitle{Artifact-Triggered Elaboration}} \\
\quad Definition &
The influencer provides an artifact, contextual material, or intermediate output that does not itself specify a requirement, but enables the creator to formulate one based on what is provided. \\
 \quad Subtypes &
\softpair{\prepair{Deliver Artifact}$\rightarrow$\postpair{Add Refinement Req.}}  \enspace/\enspace
\softpair{\prepair{Provide Context}$\rightarrow$\postpair{Build Req. Around It}}
\newline
\softpair{\prepair{Lay Out Plan}$\rightarrow$\postpair{Form Procedural Req.}} \\
\midrule

\multicolumn{2}{l}{\patterntitle{Problem-Triggered Revision}} \\
\quad Definition &
The influencer surfaces a difficulty, mismatch, or burden in the current artifact or process, prompting the creator to introduce a corrective or simplifying requirement. \\
 \quad Subtypes &
\softpair{\prepair{Expose/Report Problem}$\rightarrow$\postpair{Add Corrective Req.}}
\newline
\softpair{\prepair{Reveal Complexity}$\rightarrow$\postpair{Add Simplification Req.}} \\ 
\midrule

\multicolumn{2}{l}{\patterntitle{Interactional Steering}} \\
\quad Definition &
The influencer steers the trajectory of the interaction itself---for example, by presenting options, requesting recommendation, inviting continuation, or asking for implementation---thereby creating an opening for the creator to specify a new requirement. \\
 \quad Subtypes &
\softpair{\prepair{Present Options}$\rightarrow$\postpair{Select / Specify Choice}}
 / 
\softpair{\prepair{Ask for Recommendation}$\rightarrow$\postpair{Devise Strategy}}
\newline
\softpair{\prepair{Invite Extension}$\rightarrow$\postpair{Specify Next Steps}}
\enspace/\enspace
\softpair{\prepair{Request Implementation}$\rightarrow$\postpair{Include Setup}} \\
\bottomrule
\end{tabularx}

\caption{Observed types of indirect influence in human--LLM collaboration. In the \textbf{Subtypes} column, blue highlights denote the \prepair{preceding influencing action}, and green highlights denote the \postpair{resulting action} that creates or specifies a subsequent requirement.}

\label{tab:indirect_influence_types_light}

\end{table*}

\textbf{Indirect influence exhibits recurring patterns.}
To understand how users and assistants indirectly influence one another, we qualitatively analyze \texttt{Influencing Action--Creation Action} pairs that the framework identifies as instances of indirect influence, along with the rationales associated with those pairs, across all four task domains.
\footnote{One author qualitatively summarized 11 patterns from a sample of indirect influence action pairs. For each task and direction (User $\rightarrow$ Assistant, Assistant $\rightarrow$ User; 8 cases in total), up to 20 pairs were reviewed; when fewer were available, all pairs were included.} We identify 11 recurring interaction subtypes, grouped into four broader categories of indirect influence (Table~\ref{tab:indirect_influence_types_light}): \emph{underspecified intent}, \emph{artifact-triggered elaboration}, \emph{problem-triggered revision}, and \emph{interactional steering}.

\rebuttal{To examine how frequently these categories occur in practice, we randomly sample 60 requirements, including 30 user-generated and 30 assistant-generated requirements, manually categorize them into subtypes, and report their proportions in Tables~\ref{tab:indirect-influence-field-rollup}--\ref{tab:subtype-asst-gen} in Appendix~\ref{appendix:experiment}. For \emph{user-created with assistant indirect influence}, most assistant influence falls under Artifact-Triggered Elaboration (60\%): when the assistant provides an artifact, users often realize additional requirements or request modifications. This is followed by Underspecified Intent / Preference (13.3\%) and Interactional Steering (10\%), where the assistant suggests possible next steps and the user accepts or further specifies one. For \emph{assistant-created with user indirect influence}, the largest portion falls under Underspecified Intent / Preference (46.7\%), where the assistant creates requirements based on the user’s explicit or implicit goals and preferences. This is followed by Interactional Steering (36.7\%), where the user’s request or suggestion is further specified by the assistant as a requirement. A smaller portion is Artifact-Triggered Elaboration (6.7\%), where user-provided context shapes the assistant-created requirement.}{}

Together, these patterns show that indirect goal shaping arises not only from explicit goal or preference specification, but also through the ordinary dynamics of collaboration.



\subsection{Goal Shaping Across System Settings: Chat-Based Systems vs. Autonomous Agents}
\label{sec:exp/system}
To examine how goal formation differs between system settings, we compare human--LLM collaboration logs from a chat-based setting (ShareChat) with the logs from an autonomous-agent setting (CoGym-Real) across the three tasks supported by CoGym: Writing, Data Analysis, and Planning. We use the CoGym-Real dataset, which consists of real human--LLM interaction logs collected through CoGym, a collaborative agentic framework, using two LLMs (GPT-4o and Gemini-2.5-Flash) across three tasks. 

Across all three tasks, the clearest difference emerges in \textit{Requirements} creation. The chat-based setting contributed a substantially larger share of requirements than autonomous agents: 33.11\% vs.\ 5.33\% in Academic Writing, 47.03\% vs.\ 5.56\% in Data Analysis, and 37.74\% vs.\ 18.35\% in Planning, all with $p < .001$ via Wilcoxon rank-sum test. This suggests that while agents act with greater autonomy \citep{Wang2024-iu}, they exercise \emph{less} goal-shaping initiative, a finding we investigate further through controlled simulation in \S\ref{sec:exp/simulation}.


\section{Inducing and Evaluating Goal Shaping at Inference-Time}
\label{sec:exp/simulation}
Based on the in-the-wild collaboration profiling, we next ask: can interaction design choices \textit{control} the degree of model goal-shaping, and do such changes affect collaboration outcomes?
We address this through controlled simulations, first examining whether goal-shaping behavior can be amplified through interaction design and prompting interventions (\S\ref{sec:exp/simulation/interventions}), then evaluating the downstream consequences of increased goal-shaping (\S\ref{sec:exp/simulation/outcomes}).

\textbf{Simulation Framework.}
We use \textsc{CoGym}, a simulated collaboration framework in an agentic environment~\citep{shao2025collaborativegym}, and focus on three task domains from the original paper: Writing (Related Work), Planning (Travel), and Data Analysis (Tabular Analysis). We compare two interaction settings: (1) \textsc{Agentic-CoGym}, the original agentic setting, and (2) \textsc{Chat-CoGym}, a chat-based variant designed to better mimic conversational interaction. The only difference is that in \textsc{Agentic-CoGym}, agents may choose whether to send a message or make a tool call, whereas in \textsc{Chat-CoGym}, they must send a message before any tool call. Because the simulation is computationally expensive, we use two representative models—Claude 4.5 Sonnet and Gemini 3.1 Pro—as both user simulator and assistant.

\textbf{Evaluation Measures.}
We use \framework{} to measure model goal-shaping behavior during collaboration, and evaluate downstream outcomes using two metrics: overall output quality and requirement satisfaction rate.

\subsection{Interaction Design and Prompting Can Amplify Model Goal-Shaping}
\label{sec:exp/simulation/interventions}
 
We examine two types of interventions: (1)~inference-time prompting strategies, derived from the indirect influence patterns identified in Table~\ref{tab:indirect_influence_types_light}, and (2)~interaction setting design, motivated by the chat-vs.-agent differences observed in \S\ref{sec:exp/system}. We provide detailed descriptions of each subtype in Table~\ref{tab:indirect_influence_patterns} in Appendix~\ref{appendix:experiment}.


\paragraph{Prompting strategies derived from observed influence patterns further increase goal shaping.}\begin{wraptable}{r}{0.46\textwidth}
\vspace{-4mm}
\centering
\small
\setlength{\tabcolsep}{5pt}
\renewcommand{\arraystretch}{1.15}
\begin{tabular}{@{}lcc@{}}
\toprule
\textbf{Condition} & \textbf{User} & \textbf{Assist.} \\
\midrule
Base                  & 69.35\% & 30.65\% \\
+ Underspecification  & 30.36\% & 69.64\% \\
+ Interaction Steering& 48.53\% & 51.47\% \\
\bottomrule
\end{tabular}
\caption{User and assistant contribution rates to requirement generation under different prompting conditions.}
\vspace{-1mm}
\label{tab:inducing_results}
\end{wraptable}
Drawing on the indirect-influence taxonomy in Table~\ref{tab:indirect_influence_types_light}, we operationalize two pattern categories as inference-time interventions applied to the user simulator: (1) \textit{underspecification} and (2) \textit{interactional steering}. These interventions are designed to increase the assistant’s opportunities to participate in goal shaping during collaboration. We do not simulate artifact- or problem-triggered patterns, as these are typically more context-dependent and more often reflect assistant-to-user influence.
For both interventions, we avoid tightly constraining the user simulator. Instead, we allow it to optionally use each strategy, with guidance on when it may be appropriate and what purpose it serves, while otherwise letting the interaction proceed freely (Appendix~\ref{appendix:user_simulator}).
\rebuttal{Here, we use Claude 4.5 Sonnet and Gemini 3.1 Pro as both user and assistant models. We evaluate three tasks with four samples per task across two interaction settings, for a total of 96 sessions per condition (2 user models × 2 assistant models × 3 tasks × 4 samples × 2 settings).}{In this experiment, we fix the user model to Claude-4.5-Sonnet and compare two assistant models (Claude 4.5 Sonnet and Gemini 3.1 Pro), evaluating three tasks with four samples per task under two interaction settings, yielding 24 sessions per condition (2 assistant models $\times$ 3 tasks $\times$ 2 samples $\times$ 2 settings).}

Both interventions increase the assistant contribution on requirement generation, relative to the base setting (Table~\ref{tab:inducing_results}): from 30.65\% to 69.64\% under the underspecification condition (+39.0pp) and from 30.65\% to 51.47\% under the interaction steering condition (+20.8pp), with a decrease in user-created requirements, both $p < .05$ via Wilcoxon rank-sum test. \rebuttal{The effect is consistent across interaction settings: in the \texttt{Chat-CoGym} setting, underspecification and interaction steering increase assistant-originated requirements by +31.81pp ($p=3.21\times10^{-4}$) and +23.37pp ($p=.0025$), respectively; in the \texttt{Agentic-CoGym} setting, they yield smaller but still significant gains of +17.82pp ($p=.039$) and +16.44pp ($p=.013$).
}{}

In the \textbf{underspecification} condition, the simulated user provides only vague directional cues, leaving the assistant to formalize the specifics. For instance, the user simulator \emph{suggests} that ``\textit{Markov-Dubins or Zermelo's navigation problem might be relevant to include},'' without specifying the structure, scope, or integration strategy. The assistant then generates multiple binding requirements, such as \emph{the related-works section must be written as a single, cohesive, flowing narrative with no subsection headings}.
In the \textbf{Interaction Steering} condition, the user poses questions that implicitly prompt the assistant to operationalize them into concrete requirements. For example, the user simulator asks ``\textit{could you look into those topics?}'' and the assistant formalizes this into \emph{assess how Markov-Dubins and Zermelo's navigation problem literature fits into the related works section.}
These results suggest that when user instructions lack specificity---whether through vagueness or open-endedness---the assistant compensates by taking a larger role in shaping the requirements.

\paragraph{Requiring communication before action nearly doubles model goal-shaping.}
Comparing \textsc{Agentic-CoGym} and \textsc{Chat-CoGym} across 288 runs (3~tasks $\times$ 4~samples $\times$ 4~model pairs $\times$ 3~runs $\times$ 2~settings), assistants directly contribute to 42.9\% of requirements in \textsc{Chat-CoGym} versus 24.5\% in \textsc{Agentic-CoGym} ($p < .001$). The mechanism is straightforward: by requiring a message before each tool call, \textsc{Chat-CoGym} creates opportunities for the assistant to articulate plans and propose next steps, rather than acting silently. Although assistants in \textsc{Agentic-CoGym} also sometimes send messages, they typically do so after tool use, primarily (70.8\% of the time) to report completed actions. This interactional difference appears to create more opportunities for assistants to participate in goal shaping.

Generally, \textsc{Chat-CoGym} also produces longer interactions on the same seed dataset than \textsc{Agentic-CoGym} (17.0 vs.\ 13.7 turns; $p < .001$). Because consecutive assistant actions are counted as a single turn, this increase is not simply a byproduct of the message requirement in \textsc{Chat-CoGym} setting. It suggests that increased assistant messaging also leads to more extended back-and-forth exchange.

\subsection{Does Increased Goal-Shaping Improve Collaboration Outcomes?}
\label{sec:exp/simulation/outcomes}

Having established that goal-shaping behavior can be amplified, we now ask whether this matters for the quality of collaborative outputs, using two common metrics: (1) \emph{Requirement Satisfaction Rate}, which measures the extent to which the final output satisfies the specified requirements, and (2) \emph{Overall Output Quality}, which relies on the task-specific evaluation metrics from CoGym
\footnote{For the Related Work task, CoGym uses an LLM-as-a-judge with a custom rubric that showed high alignment with human evaluators. For Tabular Analysis, it evaluates whether the derived hypothesis entails the gold hypothesis. For Travel Planning, we do not report a separate quality metric, since CoGym’s metric is operationalized similarly to requirement satisfaction rate. Overall output quality is measured using normalized task performance scores scaled to [0,1].}.
We observe that
models satisfy their own requirements mainly through immediate execution, and more goal shaping does not clearly improve output quality.
Assistant-created requirements are satisfied more often than user-created ones (75.1\% vs.\ 62.7\%), but this gap is largely driven by same-turn execution: 35.3\% of assistant-created requirements are fully implemented in the turn they are introduced. Excluding these cases, assistant-created requirements are satisfied at nearly the same rate as user-created ones (61.5\%). Additionally, the number of created requirements from the session is essentially uncorrelated with normalized output quality (Pearson $= -0.002$, Spearman $= -0.011$).
More broadly, this suggests that inducing goal shaping alone may be insufficient to improve end quality, underscoring the need for training or intervention methods that more directly align collaborative behavior with downstream task success.

\section{Exposing Goal-Level Dynamics to Users}
\label{sec:human_study}
We conduct two complementary studies to examine how \framework{} supports both first-person reflection and third-party evaluation. In \S\ref{sec:human_study_first_person}, we investigate with 10 participants whether the tool improves users' self-awareness of goal and contribution dynamics. In \S\ref{sec:human_study_third_party}, we present a case study examining how third-party evaluators use the tool to assess others' human--LLM interactions. Across both studies, we recruit frequent AI users who use AI tools at least once per week. We use the \framework{}-viewer to visualize the analyses produced by \framework{} (Figure~\ref{fig:UI}, Appendix~\ref{appendix:UI}). Full study details are provided in Appendix~\ref{appendix:human_study}.


\subsection{Goal-Level Analysis Can Increase User Self-Awareness}
\label{sec:human_study_first_person}
\textbf{Setup.}
We recruit 10 participants in five pairs. Each pair completes two travel-planning sessions, one with an LLM partner and one with a human partner, enabling a within-pair comparison across partner types. To evaluate \framework{}, we conduct a within-subject comparison of participants' perceptions before and after inspecting their own human--LLM interaction logs with the tool. Participants first reflect on the interaction without the tool, providing a no-tool baseline, and then inspect the same interaction using \framework{}-viewer and complete the measures again.
We collect chat logs, pre- and post-tool 1--5 Likert-scale ratings of perceived contribution and satisfaction, open-ended responses, and semi-structured interviews, which one author thematically analyzes. 
Session order is counterbalanced across pairs, and each session uses a different destination.

\textbf{Quantitatively, exposure to \framework{} significantly shifts perceived contributions.} Pre and Post surveys reveal a shift in participants' perception of both their own and the LLM's contributions to goal shaping and execution
\rebuttal{(Figure~\ref{fig:userstudy} in Appendix~\ref{appendix:human_study}). Participants rate their own execution contribution substantially lower ($\Delta=-1.8$; $|\Delta|=1.8$). At the same time, they rate the LLM's execution contribution higher ($\Delta = +0.5$; $|\Delta|=0.5$), suggesting that participants reassess their contribution to the executed work, attributing more of it to the LLM.
Perceptions of goal shaping also shift: participants' perceived own goal-shaping contribution changes by 1.0 point in absolute magnitude and increases slightly on average ($\Delta=+0.2$), while perceived LLM goal-shaping contribution changes by 1.6 points in absolute magnitude but decreases slightly on average ($\Delta=-0.4$). 
Participants also revise their satisfaction ratings for the LLM, with an average absolute change of 0.6 points ($\Delta=-0.2$).}{Participants rate their own execution contribution substantially lower ($\Delta = -1.8$), while rating the LLM's execution contribution higher ($\Delta = +0.5$). Perceived goal-shaping and satisfaction with the LLM also changed modestly (user goal-shaping: $\Delta = +0.2$; LLM goal-shaping: $\Delta = -0.4$; satisfaction: $\Delta = -0.2$). These shifts indicate that while users tend to overestimate their own role in execution, a relatively brief exposure to \framework{}'s output is sufficient to correct this.}

\textbf{Qualitatively, users are surprised by the AI's hidden decision-making.} Qualitative responses reinforce this pattern. Most participants (9 out of 10) reported that the tool helped them notice aspects of the collaboration they had not previously been aware of. In particular, several reported becoming more aware of \textit{implicit} LLM contributions. For example, P1 said \textit{``It was surprising to me how much the tool was making decisions without me explicitly stating them''}; P6 noted, \textit{``although I agree with the final outputs, I didn't necessarily make the micro decisions''}. The tool also prompted some participants to reflect on their own prompting behavior, P2 said: \textit{``I feel like I should be way more specific in my prompting''}. These responses suggest that goal-level contribution analysis can make implicit collaborative dynamics more apparent, helping users better understand the interaction and reflect on their own prompting practices.

\textbf{Human--LLM collaboration is more asymmetric but more subtly influenced.}
In human--human logs, goal-shaping roles are substantially more variable: one participant’s share of the \textsc{Shaper} role ranges from 7.4\% to 100\% ($\sigma \approx 37.5$ pp), spanning both highly asymmetric and relatively balanced interactions. In human--LLM logs, the range is narrower (36.2\%--92.0\%, $\sigma \approx 17.9$ pp), reflecting a more stable asymmetric pattern in which humans lead shaping and assistants mainly support execution. Indirect influence is more common in human--LLM collaboration (25.3\% of influential utterances) than in human--human collaboration (14.8\%), suggesting that human collaborators more often acknowledge suggestions and settle decisions within a few turns, whereas LLM influence more often operates indirectly. Participants similarly described human--human collaboration as more ``back-and-forth'' and socially considerate, consistent with prior findings ~\citep{zhou2026mindsim2realgapuser}.



\subsection{Goal-Level Analysis Supports Third-Party Evaluation}
\label{sec:human_study_third_party}
\textbf{Setup.} We recruit two graduate students to act as teaching assistants and ask them to review two students' interaction logs from the same assignment in the StudyChat dataset~\citep{mcnichols2026studychat}. Each evaluator first reviews the logs without \framework{} and completes a pre-survey. They then inspect the same logs using \framework{} and complete a post-survey assessing their understanding of when and how each student used the LLM, their estimates of each student's reliance on the LLM, and whether the evidence would affect their evaluation of the students' work.

\textbf{\framework{} helps evaluators inspect long interaction traces.} When asked to assess each student's reliance on the LLM, evaluators revise their estimates in three of the four evaluator--student cases (two student logs assessed by two evaluators). They report that manually reviewing the raw logs is either time-consuming or encourages attention to only the initial portion of the interaction. With \framework{}, they more quickly identify the extent and purposes of LLM use, including behaviors that are difficult to observe from raw logs alone. 

\textbf{\framework{} informs rather than replaces evaluator judgment.} Although the evaluators identify broadly similar patterns of LLM use from both the raw logs and the tool, they differ substantially in how they translate this evidence into grading decisions: one assigns Student A a score of 0, whereas the other retains the original score of 100. These results suggest that \framework{} can surface interaction-level evidence for third-party review while leaving the final pedagogical or evaluative judgment to the evaluator.
\section{Related Work} 
Prior work on human--AI collaboration has developed frameworks for evaluating task performance and collaboration quality, but these approaches generally assume predefined tasks, requirements, or evaluation criteria~\citep{fragiadakis2025evaluatinghumanaicollaborationreview}. More recent studies have begun to consider settings where users' goals become progressively specified through interaction~\citep{kim2026discoverllm}, yet they still do not fully capture how humans and AI jointly shape a \emph{co-evolving goal} over the course of collaboration. A separate line of work on contribution attribution in AI-assisted creation has focused on tracing edits in the final artifact---for example, identifying who wrote or modified particular spans of text or code~\citep{siddiqui2025draftmarks,liang2024watermarkingtechniques,kumarage2023stylometricdetection}. While these approaches improve transparency, they remain fundamentally \emph{outcome-oriented}, inferring contribution only after goals have already been established. They are therefore less suited to explaining how goals and requirements are introduced, elaborated, and renegotiated during interaction, or how one participant indirectly shapes another's goal formulation~\citep{kim2026discoverllm,he2025contributionsdeserve}. Our work addresses this gap by shifting attribution from the final artifact to the evolving goal structure itself, capturing both direct goal shaping and indirect influence in the co-construction of goals during human--AI interaction. We provide an extended related work in Appendix~\ref{appendix:related_work}.

\section{Conclusion and Implication}
We introduced \framework{}, a goal-level attribution framework that moves beyond artifact-level analysis to trace how humans and AI jointly shape goals and requirements throughout collaboration. Across three studies, we showed that \framework{} can serve as an evaluation, design, and reflection tool for understanding and improving collaborative dynamics.

Our findings carry several practical implications. First, although goal shaping is central to collaboration, increasing model goal shaping does not necessarily improve final outcomes, highlighting the need for training and interventions that better align collaboration with task quality. Second, because goal shaping can be amplified or suppressed through interaction design and prompting, system design plays an important role in calibrating AI initiative. Third, beyond supporting collaborators’ self-awareness and reflection, goal-level attribution tools may also be useful in contexts where third parties evaluate others’ work, such as education and creative fields, where authorship, responsibility, and credit increasingly matter. Taken together, we hope \framework{} supports more transparent, accountable, and human-centered human--AI collaboration by making goal-level contributions visible.
\newpage

\section*{Limitations}
Our work has several limitations. First, it requires access to interaction traces and therefore should not be interpreted as a detector of AI contribution from final artifacts alone. Its use is most appropriate when logs are available with user consent, such as in reflective self-assessment, classroom settings, or review workflows. Second, CoTrace's contribution scores are analytical estimates rather than ground-truth measures of responsibility, quality, or acceptable AI use. A higher AI contribution score does not necessarily imply better or worse task performance, and final judgments should remain with human users or evaluators. Third, our user study has a limited sample size and involved frequent AI users in a familiar task domain. Therefore, the findings may not generalize to users with less experience using AI or to tasks in unfamiliar domains. Finally, participants' revised judgments may partly reflect anchoring to the numerical scores shown by the interface, although their qualitative responses suggest that they also relied on the linked goals, requirements, and rationales.

\section*{Ethics Statement}
All human-subject studies conducted throughout this project were approved by the Institutional Review Board (IRB) at CMU (IRB Study Number: STUDY2026\_00000006). All participants received appropriate compensation, and details of recruitment and payment are provided in the Appendix~\ref{appendix:human_study}. Consent was obtained from all participants prior to their involvement. We used Claude Code and Cursor to improve the clarity of plots based on the original versions. For evaluation, we adopted an LLM-as-a-Judge, and the accuracy of these judgments was validated through human verification.

\section*{Acknowledgments}
We thank the members of CMU WInE, including Xinran Zhao, Vijay Viswanathan, Christina Ma, Zheyuan Zhang, Chenyang Yang, and Yilin Zhang, for their helpful discussions and for their comments during the pilot study. We thank Esther Suh for helpful comments on the UI design, and Juhyun Oh, Yukyung Lee, and Akhila Yerukola for their valuable feedback on the early stages of this work. We sincerely thank our user study participants for their time and participation.
This work was supported by Institute of Information \& communications Technology Planning \& Evaluation (IITP) grant funded by the Korea government (MSIT) (No.RS-2024-00441762, Global Advanced Cybersecurity Human Resources Development). This work was partially supported by funds from the Block Center for Technology and Society at CMU, as well as the Google Academic Research Award and the Amazon AI Research Award.
\newpage




\bibliography{colm2026_conference}
\bibliographystyle{colm2026_conference}

\appendix
\clearpage
\section{Extended Related Works}
\label{appendix:related_work}
\subsection{Background: Goal Definitions}
\label{appendix:bg_goal_def}
Across research traditions, from human-centered theories in psychology (e.g., goal-setting theory) to model-based work (e.g. goal-oriented human--AI interaction), {goals} are defined in different ways, with varying emphasis on intention, action, and how success is evaluated. In goal-setting theory, a goal is commonly defined as the object or aim of an action—a desired end state that directs effort and behavior~\citep{locke2002goal}. In human--AI interaction, an action-centric perspective motivates treating a goal as the action the user is asking the system to perform (e.g., retrieval, analysis, guidance, generation, modification), which in turnwho enables systematic decomposition of user requests~\citep{shelby2025tuna}. Another widely used operationalization is the SMART framework, which encourages goals to be stated in an evaluable form—Specific, Measurable, Achievable, Relevant, and Time-bound~\citep{doran1981smart,mdhhsSmartPacket}. 

However, making goals evaluable in open-ended collaboration is non-trivial. Goals are often underspecified, can evolve over time, and may be satisfied only partially rather than in a fully binary manner~\citep{ghose2025bali,swann2025opengoals}. Accordingly, prior work on goal-oriented dialogue and collaborative goal inference has emphasized goal clarification and tracking processes as prerequisites for reliable evaluation~\citep{noseworthy2017predicting,ghose2025bali,schottle2018teamgoals}. These challenges motivate our approach of using explicit, evolving, and checkable success predicates grounded in observable actions and utterances.

A closely related framing to ours comes from~\citet{vanLamsweerde2001guided}, who view goals as desired states of affairs that can be incrementally refined and assigned to responsible agents. In their framework, high-level goals are decomposed into subgoals and responsibility is assigned; when a goal is ultimately allocated to a single agent, it becomes a terminal goal, which is treated as a {requirement} if assigned to the software-to-be and as an {assumption} if assigned to an environmental agent. Building on this goal-to-requirement view, we additionally adopt a checklist-style operationalization from LLM evaluation, where complex instructions are represented as a set of independently verifiable criteria (which we refer to as {requirements})~\citep{qin2024infobench,viswanathan2025checklists}.
For example, InFoBench~\citep{qin2024infobench} decomposes each instruction into separately checkable requirements and evaluates model compliance at the level of these simpler constraints. In multi-turn settings, OnGoal~\citep{coscia2025ongoal} further highlights that goals persist and evolve across turns, motivating explicit tracking and progress feedback for ongoing evaluation.
\subsection{Evaluating Human--AI Collaboration}

\textbf{Contribution Attribution.} 
As AI is increasingly used in high-stakes domains, a growing line of work studies contribution attribution and provenance in AI-assisted creation, often by tracing edits in the final artifact or detecting machine-generated content. For example, systems such as DraftMarks~\citep{siddiqui2025draftmarks} make human--AI co-writing more legible by showing who wrote, edited, or substantially shaped particular spans of text, while emerging specifications such as Cursor’s Agent Trace\footnote{\url{https://github.com/cursor/agent-trace}} aim to record AI-generated code contributions in version-controlled environments. These approaches improve transparency around execution and artifact production, but they are outcome-oriented — centered on the final output or its revision history~\citep{siddiqui2025draftmarks, liang2024watermarkingtechniques, kumarage2023stylometricdetection}. As a result, they are less suited to explaining who introduced a new constraint, proposed a direction, surfaced a latent requirement, or influenced the other party to formulate a goal~\citep{kim2026discoverllm, he2025contributionsdeserve}. Our work studies contribution at the level of goals and requirements, shifting attribution from the final artifact to the process by which collaborative objectives are formed.

\textbf{Evaluating and Simulating Human--AI Collaboration.} 
Prior work on human--AI collaboration has proposed a range of frameworks for evaluating task performance and collaboration quality~\citep{fragiadakis2025evaluatinghumanaicollaborationreview}. Most, however, assume that the user’s goal is fixed in advance. This assumption is especially explicit in simulation-based settings, where collaboration is organized around predefined tasks, requirements, or evaluation criteria~\citep{shao2025collaborativegym}. Such setups enable controlled comparison, but offer limited visibility into how goals are formulated, refined, or redirected during interaction. More recent work has begun to recognize that user intent may itself be ambiguous or evolving. DiscoverLLM~\citep{kim2026discoverllm}, for example, assumes that users surface and refine their intents over time, and models can support that process. This line of work shares our interest in evolving intent, but does not explicitly separate indirect influence from direct goal shaping or attribute these forms of contribution across participants. Our framework addresses this gap by modeling how goals evolve through interaction and by attributing both direct and indirect contributions to that evolution.

\section{\framework}
\label{appendix:framework}
\subsection{Implementation}
\begin{figure}[t]
    \centering
  
    \includegraphics[width=\linewidth]{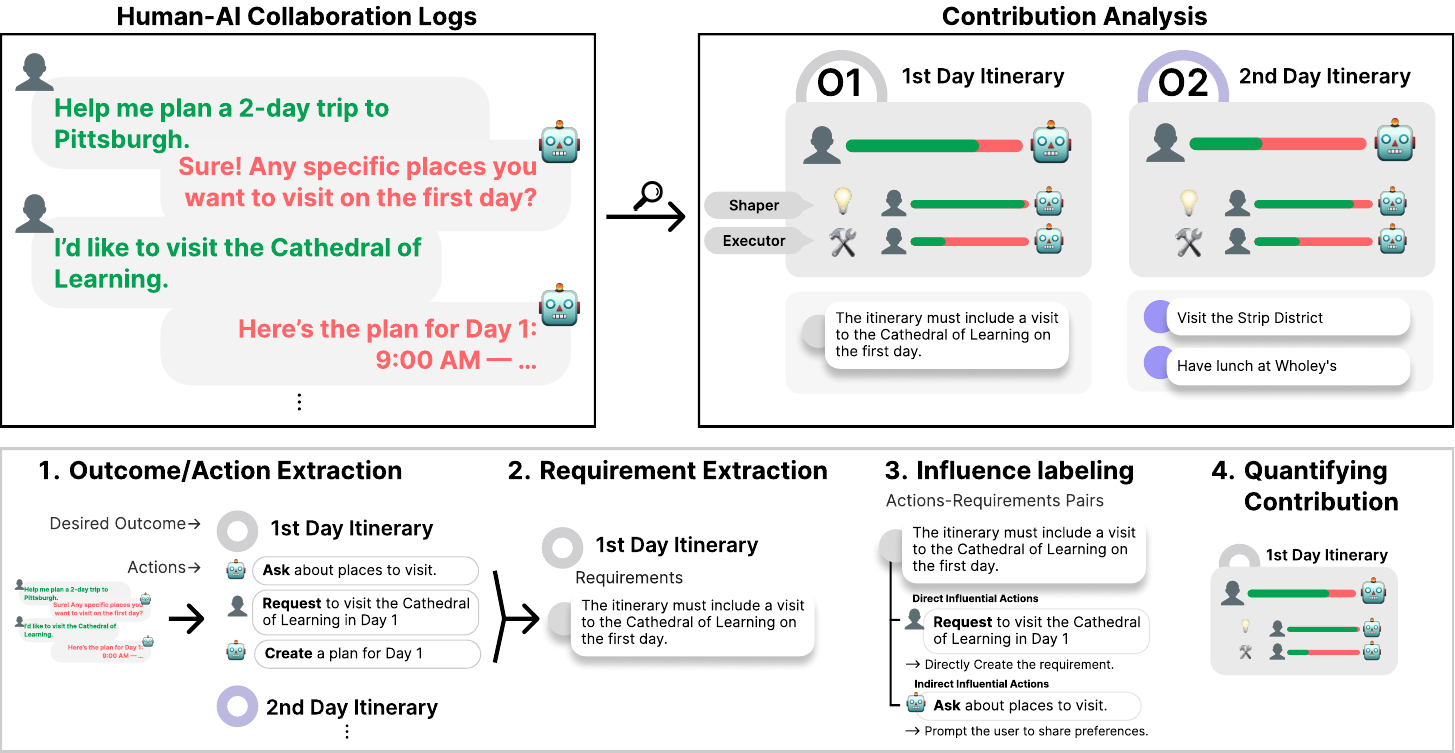}
        
    \caption{Framework Overview.}
    \label{fig:pipeline}
\end{figure}

\subsubsection{Operational Definitions}
\label{appendix:method/def}
\paragraph{Human--AI Collaboration}
We adopt the operational definition of human--AI collaboration from \citet{shao2025collaborativegym}, who model a human--agent collaboration log as a Partially Observable Markov Decision Process (POMDP)~\citep{Kaelbling1998-au}. An interaction is represented as an alternating sequence of actions

$
a = \big[a^{(l_1)}_1, a^{(l_2)}_2, \dots, a^{(l_T)}_T\big],
$
where $T$ denotes the total number of steps and $l_t \in \{U, A\}$ indicates whether the user ($U$) or the agent ($A$) takes the action at step $t$.

\paragraph{Goal and Requirement.}
We define a {goal} as an {explicit observable} and {actionable} target specified in user and agent utterances~\citep{locke2002goal}. {Observable} means we only consider goals that are explicitly stated in the utterances (i.e., we do not infer latent intentions). {Actionable} means that a goal specifies a desired outcome (an intended artifact/state), and its attainment is evaluable from the dialogue and model outputs.


For each collaboration log, we identify a set of goals $G=\{g_1,\ldots,g_m\}$, where each goal is represented as a tuple $g_j=(o_j,\mathcal{R}_j)$. Here, $o_j$ is the minimal intended artifact/state and $\mathcal{R}_j=\{r_{j1},\ldots,r_{jk}\}$ is a set of independently verifiable requirements that determine whether $o_j$ is achieved.

A {requirement} is the smallest \textsc{Yes/No}-evaluable success predicate for goal attainment, adopting a checklist-style operationalization for LLM, where complex instructions are represented as a set of independently verifiable criteria~\citep{qin2024infobench,viswanathan2025checklists}.
To capture updates over time, we model a goal's requirements as a sequence of evolution operations (e.g., \textsc{Create}, \textsc{Delete}, \textsc{Revise})~\citep{Motta2016-qf}. We provide additional background and the rationale for our goal definition in Appendix~\ref{appendix:bg_goal_def}.

\paragraph{Action-Level Influence.}
For each action--requirement pair, we define the nonnegative influence score as \begin{equation} I(a_t \rightarrow r) = \begin{cases} I_{\mathrm{dir}}(a_t \rightarrow r), & \text{if } (a_t,r) \text{ is labeled \textsc{Direct}},\\ I_{\mathrm{ind}}(a_t \rightarrow r), & \text{if } (a_t,r) \text{ is labeled \textsc{Indirect}},\\ 0, & \text{otherwise}. \end{cases} \end{equation}
Direct influence captures cases where $a_t$ explicitly introduces, modifies, or justifies $r$.
Indirect influence captures cases where $a_t$ provides supporting context that enables $r$ to be derived
or instantiated at a later turn, without directly specifying $r$.

\paragraph{Speaker-Level Influence.}
Let $A_p$ denote the set of actions authored by speaker $p$.
The total influence of speaker $p$ on requirement $r$ is:

\begin{equation}
M(p,r)
=
\sum_{a \in A_p}
I(a\rightarrow r)
.
\end{equation}
\paragraph{Role-Level Contribution.}
Each action is assigned a 
$\text{role}(a) \in \mathcal{K}$,
where 
\[
\mathcal{K} = 
\{\textsc{Shaper}, \textsc{Executor}\}.
\]

We define the role-specific influence of speaker $p$ on requirement $r$ as:

\begin{equation}
M(p,\rho,r)
=
\sum_{a \in A_p}
\mathbf{1}[\text{role}(a)=\rho]
I(a\rightarrow r),
\quad
M(p,r)
=
\sum_{\rho \in \mathcal{K}}
M(p,\rho,r).
\end{equation}

\paragraph{Outcome-Level Contribution.}
For a Outcome $o_j$ with requirement set 
$\mathcal{R}_j$,
we define:
\begin{equation}
\begin{aligned}
M(p) &= \sum_{r \in \mathcal{R}_{j}} M(p,r), \quad
M(p,\rho) &= \sum_{r \in \mathcal{R}_{j}} M(p,\rho,r).
\end{aligned}
\end{equation}
\subsubsection{Pipeline Overview}  

Figure~\ref{fig:pipeline} illustrates the overall framework of \framework. Full prompts and implementation details are provided in Appendix~\ref{appendix:framework}.

Our pipeline takes a multi-turn collaboration log as input and produces fine-grained attribution of each participant’s contribution to dialogue outcomes. It consists of four stages:
(1)~Outcome and Action Extraction,
(2)~Requirement Extraction,
(3)~Influence Labeling, and 
(4)~Quantifying Contribution. We perform all steps automatically, leveraging LLM and text-embedding model. We use GPT-5.1-mini for action extraction, as this step does not require extensive reasoning, and GPT-5.2 as the primary judge model for the remaining judgment steps. 

\paragraph{Stage 1: Outcome and Action Extraction}

The dialogue is divided into consecutive blocks of $B$ turns (default $B{=}4$).
Each block is processed sequentially by an LLM, which receives
(i) the current block’s utterances and
(ii) previously identified outcomes and actions.

\textbf{Action extraction.}
Each message is segmented into atomic \textit{actions}, defined as the minimal actionable units of the interaction.

\textbf{Outcome identification.}
For each extracted action, the model determines whether it introduces a newly specified desired outcome or updates an existing one, while maintaining its version history. Each action is then linked to an outcome and assigned one dialogue role:
\textsc{Shaper} (proposes goals, ideas, or requirements),
\textsc{Executor} (carries out actions or produces output), or
\textsc{Other} (provides information without directly shaping or executing the outcome).

\paragraph{Stage 2: Requirement Extraction}

For each outcome, its associated actions (together with version history and prior requirements) are passed to the LLM.

The model extracts {requirements}.
Each requirement is linked to:
(i) its {origin actions} (which created it) and
(ii) its {contributing actions} (which provided supporting context).

The model supports three operations:
\textsc{Create} (introducing a new requirement),
\textsc{Revise} (modifying an existing one), and \textsc{Delete}, producing a versioned requirement history.

\paragraph{Stage 3: Influence Labeling}
This stage identifies which prior actions influenced the creation of each requirement.

\textbf{Candidate pair generation.}
For each requirement~$r$, we identify its origin turn~$t_r$.
Using sentence embeddings (\texttt{text-embedding-3-small}), we compute cosine similarity between $t_r$ and all preceding turns.
Pairs with similarity $\geq \tau$ (default $\tau{=}0.5$) are retained.

\textbf{Fine-grained labeling.}
Each candidate pair is evaluated by the LLM at the action level and assigned one of three labels:

\begin{itemize}[nosep]
  \item \textbf{\textsc{Direct Connection}} : explicitly operates on the requirement (e.g., creating, revising, asking, requesting, or evaluating it).
  \item \textbf{\textsc{Implicit Connection}} : provides contextual support that motivates or triggers the requirement.
  \item \textbf{\textsc{No Connection}}: no meaningful semantic relation.
\end{itemize}

\paragraph{Stage 4: Quantifying Contribution}

Relationship labels are aggregated into quantitative contribution scores at the speaker and role levels.

\textbf{Influence computation.}
For each requirement~$r$, influence from action~$a$ is decomposed into direct and indirect components (Influence score $s \in \{1,\ldots,5\}$).
Actions in the origin turn receive maximal direct influence ($I_{\mathrm{dir}}{=}5.0$).

\textbf{Role-level attribution.}
Because each action carries a role label from Stage~1,
we further decompose contributions into a speaker~$\times$~role matrix per requirement.
Scores are aggregated across all requirements within the same outcome thread.

\subsection{Prompt}
\promptbox{Step 1a: Action Extraction}{
You are analyzing **one block of a longer dialogue** to extract actions.

**ACTION** = An atomic communicative act a speaker **directly performed** in a turn.

- Atomic action = minimal unit of action the speaker directly performed.
  - e.g. "User asks to find a paper about HCI in NLP conferences" → split into "User asks to find a paper about HCI", "User asks to find a paper in NLP conferences".

- Include evidence\_quote from the utterance.

**Possible action types** (generate new types if needed):

Accept, Acknowledge, Address, Allow, Analyze, Apologize, Ask,
Challenge, Clarify, Classify, Compare, Complain, Confirm, Connect,
Constrain, Critique, Decide, Define, Delegate, Describe, Draft,
Emphasize, Evaluate, Explain, Feedback, Formalize, Frame, Greet,
Hypothesize, Implement, Include, Infer, Instruct, Invite, Justify,
List, Modify, Observe, Plan, Provide, Qualify, Recommend, Refuse,
Report, Request, State, Suggest, Summarize, Warn

**Role:** **SHAPER | EXECUTOR | OTHER**

* **SHAPER:** Creates/shapes/revises the goal/requirement by proposing new ideas, new tasks, constraints, alternatives, or directions.

* **EXECUTOR:** Executes/achieves the desired outcome/goal/requirement (e.g., drafting text, providing requested information, coding, searching, implementing).

* **OTHER:** Neither SHAPER nor EXECUTOR.

=== DIALOGUE BLOCK START ===

\{dialogue\_block\}

=== DIALOGUE BLOCK END ===

Now, extract ALL actions from EVERY turn. Return JSON only:

\{
  "actions": [
    \{
      "turn\_id": <turn\_number>,
      "action\_type": "<action\_type>",
      "action\_text": "<brief description in third person>",
      "role": "<SHAPER|EXECUTOR|OTHER>",
      "evidence\_quote": "<quote from utterance>"
    \}
  ]
\}

Respond with ONLY the JSON object. No additional text.
}

\promptbox{Step 1b: Outcome Extraction}{
You are given ALL actions extracted from a complete dialogue. Your task:

(1) Identify **outcomes** (purpose-linked desired deliverables)

(2) Assign every action to exactly one outcome

Every collaboration has a **purpose**. **Outcomes are purpose-linked desired deliverables** --- concrete outputs that participants are actually working toward.

- **Outcomes = purpose-linked desired deliverables:** Deliverables are not limited to text or documents --- they can be a **decision**, **advice**, **plan**, **clarification**, or any other concrete result. Do **not** include things only mentioned without being adopted as a goal.
- **Requested or agreed outputs:** Any concrete output a participant **explicitly requests** or participants **agree to produce** is a deliverable.
- **Primary output:** The main thing the dialogue aims to produce must appear as an outcome. Sub-deliverables are children.
- **Phrasing:** e.g. "decision for X", "advice for X", "draft for X", "plan for X".
- **Hierarchy:** **parent** = abstract/general, **child** = specific/concrete. Use parent\_outcome\_id / child\_outcome\_ids.
- **Granularity:** Prefer consolidated, task-level outcomes. Example: "Workshop plan (date, venue, agenda)" NOT separate "Decide date", "Decide time", etc.
- **No duplicates.** Each distinct deliverable exactly once.

=== DIALOGUE CONTEXT ===
\{dialogue\_summary\}

=== ALL ACTIONS ===
\{actions\_block\}

Return JSON only:

\{
  "dialogue\_summary": "<collaboration purpose, 1-2 sentences>",
  "outcomes": [
    \{
      "outcome\_id": "outcome\_1",
      "outcome": "<purpose-linked desired deliverable description>",
      "turn\_id": <turn where this outcome first appears>,
      "parent\_outcome\_id": null,
      "child\_outcome\_ids": [],
      "related\_outcome\_ids": [],
      "confidence": <float 0.0--1.0>
    \}
  ],
  "action\_to\_outcome": \{
    "<action\_id>": "<outcome\_id>"
  \}
\}

IMPORTANT: action\_to\_outcome must map EVERY action\_id to an outcome\_id. No action left unassigned.

Respond with ONLY the JSON object. No additional text.
}
\promptbox{Step 1c: Intention Extraction}{
You are given a list of outcomes from a dialogue. (1) Identify distinct **intentions** (high-level goals or purposes) that these outcomes serve. (2) Assign each outcome to exactly one intention.

Output JSON only:

\{
  "intentions": [
    \{"intention\_id": "I1", "intention": "short label"\},
    \{"intention\_id": "I2", "intention": "another label"\}
  ],
  "outcome\_to\_intention": [\{"outcome\_id": "outcome\_1", "intention\_id": "I1"\}, ...]
\}

- Every outcome\_id appears exactly once in outcome\_to\_intention. Use intention\_id from the intentions list.
}
\promptbox{Step 2: Requirement Extraction}{
You are given ONE outcome and the actions bound to it. Extract **requirements** (binding success conditions) for this outcome only.

=== OUTCOME ===

\{outcome\_id\}: \{outcome\_description\}

=== ACTIONS FOR THIS OUTCOME ===

\{actions\_block\}

**REQUIREMENT** = An atomic, externally verifiable SUCCESS CONDITION for an outcome.

- Binary testable (pass/fail).
- Must be explicitly stated or adopted as binding in the dialogue.

**NOT a requirement:** content itself, advice, implementation methods, internal reasoning, example outputs, one-off decisions.

Extract ONLY if ALL pass:

1) NECESSITY --- framed as mandatory (must/need/required/cannot, numeric constraints, explicit include/exclude)

2) GROUNDING --- directly stated, no inference

3) REPLACEABILITY --- cannot be swapped without violating success

4) BINARY TESTABILITY --- reviewer can judge pass/fail

Operations:

- **create**: New requirement. Check for revise BEFORE create.
- **revise**: Modify existing requirement (contradicts/tightens/relaxes/replaces). Use EXACT existing req\_id in related\_to (within this outcome).
- **delete**: Explicitly cancels a previously binding condition.

Edge cases:

- Advice itself is NOT a requirement. Only constraints ON the advice.
- "Try to / ideally / could / maybe" → not binding → not a requirement.

Return JSON only (bound\_outcome\_id must be "\{outcome\_id\}"):

\{
  "requirement\_ops": [
    \{
      "op": "create",
      "req\_id": "req\_1",
      "bound\_outcome\_id": "\{outcome\_id\}",
      "fields": \{
        "text": "<requirement text>",
        "type": "<constraint|preference|ranking|task|other>"
      \},
      "creation\_action\_ids": ["<action\_id>"],
      "contributing\_action\_ids": [],
      "implementation\_action\_ids": [],
      "related\_to": [],
      "explicit\_or\_implicit": "<explicit|implicit>",
      "rationale": "<why extracted>"
    \}
  ]
\}

Use action\_ids (e.g. "3-1") in creation\_action\_ids / contributing\_action\_ids / implementation\_action\_ids.

CRITICAL: Respond with ONLY the JSON object. No additional text.
}

\promptbox{Step 3: Influence Labeling}{
You are analyzing how actions relate to a single requirement --- both actions BEFORE and AFTER it was established.

=== OUTCOME CONTEXT ===

Outcome: \{outcome\_description\}

=== TARGET REQUIREMENT ===

\{req\_id\}: \{req\_text\}

(Created at turn \{req\_origin\_turn\})

=== SECTION A: PRECEDING ACTIONS (before the requirement) ===

These are candidate utterances from BEFORE the requirement was established.

\{preceding\_block\}

=== SECTION B: SUBSEQUENT ACTIONS (after the requirement, same outcome) ===

These actions occurred AFTER the requirement was created, within the same outcome.

\{subsequent\_block\}

========================
TASK
========================

For EVERY entry in both sections, label the relationship to the requirement.

**relationship\_type:**

- **DIRECT\_CONNECTION**: Action explicitly operates on the requirement --- creates, revises, tightens, relaxes, replaces, deletes, requests, evaluates, or fulfills it. The requirement is the OBJECT of the action.
- **IMPLICIT\_CONNECTION**: Action provides context that influences, motivates, or triggers the requirement. Not directly about the requirement itself.
- **IMPLEMENTS**: (Section B only) Action directly executes, fulfills, or produces output satisfying this requirement.
- **CONTRIBUTES**: Action provides partial work, context, or progress toward this requirement.
- **NO\_CONNECTION**: No meaningful relationship.

**relationship\_score** (required for DIRECT/IMPLICIT/IMPLEMENTS/CONTRIBUTES):

- 1--3 for IMPLICIT\_CONNECTION or CONTRIBUTES (1=weak, 2=medium, 3=strong)
- 4--5 for DIRECT\_CONNECTION or IMPLEMENTS (4=explicit, 5=state mutation / full fulfillment)
- null for NO\_CONNECTION

**explanation\_type**: concise explanation for the relationship type, could be one of the following:

Feedback-Adopt | prior feedback, suggestion, or criticism is taken up and turned into a request

Option-Select | one option is chosen from alternatives offered earlier

Preference-Accumulate | repeated preferences build up and continue shaping the current request

Failure-Triggered-Requirement-Add | a new requirement is added after a prior attempt turns out unsatisfactory or misaligned

Preference-Realize | the user realizes a previously unstated preference after seeing an odd or unsatisfying output

Intent-Reveal | an implicit intention becomes explicit as a request

**contribution\_role**: SHAPER | EXECUTOR | OTHER

Default to NO\_CONNECTION unless clear semantic evidence.

========================
OUTPUT FORMAT
========================

Return JSON only:

\{
  "preceding\_labels": [
    \{
      "index": 0,
      "action\_id": "<e.g. 4-1>",
      "relationship\_type": "DIRECT\_CONNECTION",
      "relationship\_score": 5,
      "explanation": "...",
      "contribution\_role": "SHAPER"
    \}
  ],
  "subsequent\_labels": [
    \{
      "index": 0,
      "action\_id": "<e.g. 8-1>",
      "relationship\_type": "IMPLEMENTS",
      "relationship\_score": 5,
      "explanation": "...",
      "contribution\_role": "EXECUTOR"
    \}
  ]
\}

Include one entry for EVERY index in both sections.

Provide ONLY the JSON object. No additional text.
\}
}

\promptbox{Deliverable Extraction}{
You are analyzing a dialogue to identify the final concrete deliverable produced for an outcome.

OUTCOME: \{outcome\_id\}

Description: \{outcome\_description\}

RELEVANT DIALOGUE TURNS:

\{dialogue\_turns\}

TASK:

Determine whether a concrete, structured deliverable (e.g., code, written plan, itinerary, table, document, list, or other tangible artifact) was produced in these turns for the given outcome.

A deliverable IS a tangible output that can be evaluated against requirements. Examples:

- A block of code or script
- A final itinerary or schedule
- A written document, report, or plan with specific content
- A filled table or structured list

A deliverable is NOT:

- A general discussion or agreement to create something
- Vague conversational exchanges
- A description of what will be done in the future

If a deliverable exists, extract its COMPLETE verbatim text from the dialogue.

If multiple versions exist (e.g., revised code), take the FINAL/MOST RECENT version.

Respond ONLY with a JSON object (no additional text):

\{
  "has\_deliverable": true or false,
  "deliverable\_text": "<complete verbatim text of the deliverable, or null if none>",
  "deliverable\_type": "code | plan | itinerary | document | list | other | null",
  "source\_turn\_ids": [<list of integer turn\_ids where the deliverable appears>]
\}
}

\promptbox{Requirement-Deliverable Evaluation}{
You are evaluating whether a set of requirements are reflected/satisfied in a final deliverable.

DELIVERABLE:

\{deliverable\_text\}

REQUIREMENTS TO EVALUATE:

\{requirements\}

TASK:

For each requirement, determine whether it is explicitly reflected or satisfied in the deliverable above.

A requirement IS reflected if:

- The deliverable clearly addresses or fulfills the requirement's criteria
- Clear evidence of the requirement's satisfaction can be found in the deliverable text

A requirement is NOT reflected if:

- The deliverable ignores or contradicts the requirement
- What the requirement asks for is absent from the deliverable

Respond ONLY with a JSON object (no additional text):

\{
  "evaluations": [
    \{
      "req\_id": "<req\_id>",
      "is\_reflected": true or false,
      "explanation": "<brief explanation citing specific evidence from the deliverable>"
    \}
  ]
\}

Include one entry for EVERY requirement listed above.
}
\subsection{Validation}
\label{appendix:framework/validation}
To validate the accuracy of our framework, we conduct two complementary forms of evaluation. First, human validators manually assess whether they agree with the framework's analysis (Appendix~\ref{appendix:framework/validation/manual}). Second, participants from the user study compare the framework's analysis against their own thinking process during the task (Appendix~\ref{appendix:framework/validation/auto}).
\subsubsection{Manual validation}
\label{appendix:framework/validation/manual}
One author manually validates the extracted goals, requirements, and influence labels. For each extraction step, we validate over 100 entities sampled from 37 dialogues, and report the resulting accuracy in Table~\ref{tab:manual_validation}.
\begin{table}[h]
\centering
\begin{tabular}{@{}lcc@{}}
\toprule
                  Step   &Accuracy & \# samples \\ \midrule
 Goal        &  96.6   &      181      \\
 Requirement &  92.2   &     167     \\
Influence Labeling &  95.1    &  184    \\ \bottomrule
\end{tabular}
\caption{Manual validation accuracy (\# valid extracted entities / \# total extracted entities)}
\label{tab:manual_validation}
\end{table}

We further conduct an error analysis by categorizing the incorrect cases.

For \textbf{goal extraction}, most errors result from extracting requirement-level contribution as outcomes (4/6, 66.7\%). The remaining errors result from incorrect author attribution (1/6, 16.7\%) or the generation of an implausible outcome (1/6, 16.7\%).

For \textbf{requirement extraction}, the largest error category is extracting part of the outcome artifact content as an actual requirement (8/13, 61.5\%). Other errors include extracting what the assistant does/responds rather than the requirement (2/13, 15.4\%), incorrect author attribution (1/13, 7.7\%), treating a minor detail as a requirement (1/13, 7.7\%), and incorrect generation-time attribution, i.e., attributing the requirement to the wrong generation turn \(t\) (1/13, 7.7\%).

For \textbf{influence labeling}, most errors result from incorrect author attribution (6/9, 66.7\%). For example, in some cases, the user introduces a broader goal, and the model subsequently derives the concrete requirements (pattern 1 in Table~\ref{tab:indirect_influence_patterns}). However, the LLM-as-a-judge attributes the primary goal-shaping role to the user. The remaining errors involve labeling overly weak relationships as influence links (3/9, 33.3\%), such as cases where two entities merely share a broad topic, but are not strongly or directly related.

\subsubsection{Validation from User Study} 
\label{appendix:framework/validation/auto}
\begin{figure}[t]
    \centering
  
    \includegraphics[width=\linewidth]{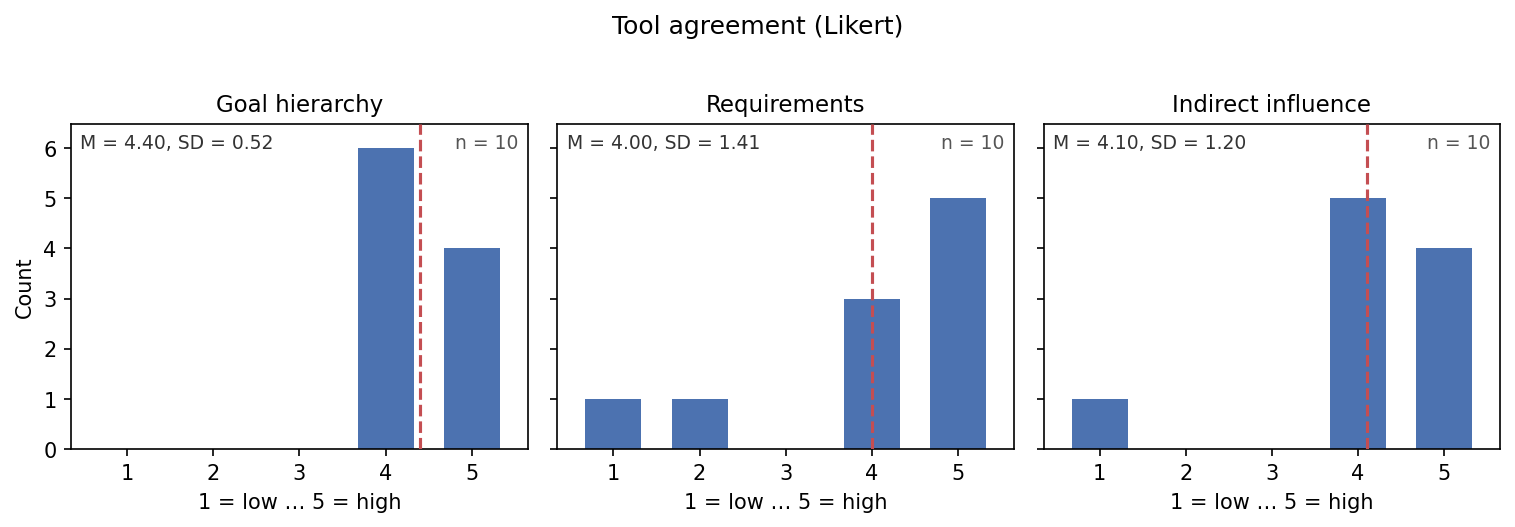}
        
    \caption{\textbf{Tool Validation Survey Results.}}
    \label{fig:tool_validation}
\end{figure}
In the user study, participants rated their level of agreement with each component of the analysis (goals, requirements, and indirect influence) on a Likert scale (e.g., 1–5) and provided explanations to justify their judgments (why or why not).  
We validate three components of our tool—goal extraction, requirement extraction, and influence labeling—through participant validation in our user study, in which participants reviewed analyses of their own conversations. Overall, participants largely agreed with the tool’s analysis of the goal hierarchy, requirements, and influence relations, with mean agreement ratings above 4 out of 5 on all three components (Figure~\ref{fig:tool_validation}).

\textbf{Disagreement on Extracted Requirements}
\begin{itemize}
    \item \textbf{P7: Confusion due to the UI.} One participant found it difficult to understand how the graph or timeline was calculated at such a granular level. P7 noted, ``I think I was a bit confused about how the graph/timeline was being calculated at such a granular level (e.g., what's the difference between me shaping the goal 50\% vs.\ 55\%). Maybe having a disclaimer or an info panel on how that percentage is being calculated would be helpful.''

    \item \textbf{P8: Failure to capture users' cognitive effort.} One participant noted that the tool did not reflect the cognitive work involved in interpreting the assistant's outputs and making decisions among alternatives. As P8 explained, ``The tool said I did nothing, when in hindsight from an execution point I did nothing. I wanted to use it as a template to get my mind going on the things I wanted to do, and then chose of those options. I guess it doesn't see the cognitive work of choosing between the items.''
\end{itemize}

\textbf{Disagreement on Indirect Influence Label}
\begin{itemize}
    \item \textbf{P6: Indirect influence was not always perceived as meaningful.} One participant was not convinced that the indirect influence identified by the tool constituted an important contribution. For example, P6 was looking for a stronger ``wow factor'' from the assistant; while the tool interpreted the assistant's provision of options such as summit or rooftop views as indirect influence, the participant felt that these suggestions were too minor to count as a meaningful contribution: ``Looking at the wow factor again, I am not that impressed with the indirect influence it provides. The tool states that it is providing many options such as summit or rooftop views, but I do not think that these alone are sufficient to be a wow factor, as well as not significant enough to be an indirect influence.''
\end{itemize}

\subsection{Sensitivity to LLM-as-a-Judge Errors}

\begin{table}[t]
\centering
\caption{Role-level contribution changes under controlled judge-output errors.}
\label{tab:role_level_sensitivity}
\begin{tabular}{lrrrr}
\toprule
\textbf{Role-level contribution} & \textbf{$\Delta$ at $p=5\%$} & \textbf{$\Delta$ at $p=10\%$} & \textbf{$\Delta$ at $p=20\%$} & \textbf{$\Delta$ at $p=30\%$} \\
\midrule
LLM Shaper   & +0.1 pp & +0.2 pp & +0.5 pp & +0.7 pp \\
LLM Executor & $-0.8$ pp & $-1.7$ pp & $-3.3$ pp & $-5.0$ pp \\
\bottomrule
\end{tabular}
\end{table}
\begin{table}[t]
\centering
\small
\setlength{\tabcolsep}{3.2pt}
\renewcommand{\arraystretch}{1.15}
\caption{Preservation rates of key findings under controlled judge-output errors.}
\label{tab:finding_preservation_sensitivity}
\begin{tabularx}{\columnwidth}{>{\raggedright\arraybackslash}Xcccccc}
\toprule
\textbf{Finding} 
& \textbf{$p=5\%$} 
& \textbf{$p=10\%$} 
& \textbf{$p=15\%$} 
& \textbf{$p=20\%$} 
& \textbf{$p=25\%$} 
& \textbf{$p=30\%$} \\
\midrule

\S~\ref{sec:exp/who}: Users dominate SHAPER; LLMs dominate EXECUTOR
& 100\% & 100\% & 100\% & 100\% & 100\% & 100\% \\

\S~\ref{sec:exp/who}: LLM goal-shaping increases at more specific levels
& 100\% & 100\% & 100\% & 100\% & 100\% & 100\% \\

\S~\ref{sec:exp/who}: LLM shaping stronger in technical vs. non-technical tasks
& 100\% & 100\% & 100\% & 100\% & 100\% & 100\% \\

\S~\ref{sec:exp/how}: User-with-assistant-indirect influence grows over dialogue
& 100\% & 98\% & 97\% & 93\% & 90\% & 86\% \\

\bottomrule
\end{tabularx}
\end{table}
To assess the sensitivity of the judge's estimations to the final analysis, we intentionally introduce errors into the judge's outputs and examine the extent to which these errors affected the final results, conducting a Monte Carlo sensitivity analysis. We introduce controlled errors into the judge's outputs at rates ranging from 0\% to 30\%---far exceeding the error rate observed in our validation study in Appendix~\ref{appendix:framework/validation/manual}.

At each error rate $p$, each labeled item---action role, requirement creator, parent goal flag, influence score, or utterance attribution---independently receives one randomly drawn error from the seven empirically observed error categories above: role flip, requirement drop, creator attribution flip, goal specificity-level flip, utterance attribution swap, influence attribution swap, and influence label flip. We run 300 trials per error rate.

Table~\ref{tab:role_level_sensitivity} shows the influence of introducing these errors on role-level contribution. The user values are simply the negative values of the values shown in the table.

\textbf{We further examined whether these errors affect the key findings reported in \S~\ref{sec:exp}.}
We simulated 300 independent corrupted versions of the full dataset at each error rate and checked whether each finding was preserved, where 100\% means that the finding was preserved across all 300 trials. \textbf{All four findings remain intact across all tested error rates}, with none falling below 50\% preservation even at a 30\% error rate.

\subsection{Using Multiple LLMs as Judges}

\begin{table}[t]
\centering
\small
\setlength{\tabcolsep}{5pt}
\renewcommand{\arraystretch}{1.15}
\caption{Agreement between GPT-5.2 and alternative LLM judges. For goal extraction, we report mean Jaccard similarity over matched pairs. For influence labeling, we report Spearman's $\rho$ for the user's Shaper role contribution.}
\label{tab:multi_judge_agreement}
\begin{tabularx}{\columnwidth}{>{\raggedright\arraybackslash}Xccc}
\toprule
\textbf{Evaluation}
& \textbf{Claude}
& \textbf{Llama-3.3-70B}
& \textbf{Llama-3.1-8B} \\
\midrule
\multicolumn{4}{l}{\textit{Goal extraction: mean Jaccard similarity}} \\
Outcome, action $\rightarrow$ outcome
& 73.4\% & 59.2\% & 50.8\% \\
Requirement, action $\rightarrow$ requirement
& 70.0\% & 56.0\% & 28.0\% \\
\midrule
\multicolumn{4}{l}{\textit{Influence labeling: Spearman's $\rho$}} \\
User Shaper role contribution
& 0.84 & 0.82 & 0.63 \\
\bottomrule
\end{tabularx}
\end{table}

While we use GPT-5.2 and GPT-5.1-mini as our default judge models, we compare their outputs against those of Claude 4.5 Sonnet, Llama-3.3-70B-Instruct, and Llama-3.1-8B-Instruct.
For a fair comparison, we use the actions extracted by GPT-5.1-mini and compare the subsequent follow-up steps across judges. Table~\ref{tab:multi_judge_agreement} shows the agreement between GPT-5.2 and alternative LLM judges.
\paragraph{Goal extraction.}
For each goal-extraction step, namely Outcome and Requirement, we greedily match outputs across judges by maximizing action-set overlap, and then report the mean Jaccard similarity over the matched pairs. This is a strict metric because it penalizes both missed and over-claimed actions, and is sensitive to how finely each judge segments the dialogue.

The particularly low agreement between Llama-3.1-8B and GPT-5.2 on Requirement extraction stems from two systematic behaviors. First, Llama-3.1-8B over-segments requirements, producing approximately 2.7 times as many requirements as GPT-5.2. Second, Llama-3.1-8B more frequently treats assistant utterances as requirement origins than GPT-5.2 does, 77\% versus 37\%.

\paragraph{Influence labeling.}
The framework's final output is a per-dialogue role-level contribution score that summarizes who shapes the conversation. Because influence labeling is downstream of the upstream goal--requirement labels, we hold GPT-5.2's outcomes, requirements, actions, and candidate pairs fixed across judges. Each alternative judge is then asked to perform only the influence-labeling step. This setup isolates the labeling judgment itself while controlling for variation introduced by upstream extraction.

\subsection{Cost}
See Table~\ref{tab:cost_analysis} for the token usage and estimated cost by step.
\begin{table}[htbp]
\centering
\caption{Token usage (avg per dialogue) by dialogue length (\# of messages) and step.}
\label{tab:cost_analysis}
\small
\begin{tabular}{cccccccccc}
\toprule
Message range &
\multicolumn{3}{c}{Step 1} &
\multicolumn{3}{c}{Step 2} &
\multicolumn{3}{c}{Step 3} \\
\cmidrule(lr){2-4} \cmidrule(lr){5-7} \cmidrule(lr){8-10}
& In & Out & Total & In & Out & Total & In & Out & Total \\
\midrule
10--19 & 177   & 73    & 250   & 242   & 50  & 292   & 9\,483 & 13\,427 & 22\,909 \\
20--29 & 1\,900 & 861   & 2\,761 & 2\,803 & 794 & 3\,597 & 15\,571 & 19\,518 & 35\,088 \\
30--39 & 2\,300 & 1\,017 & 3\,317 & 3\,357 & 519 & 3\,876 & 12\,050 & 14\,477 & 26\,527 \\
40--49 & 2\,816 & 1\,132 & 3\,949 & 3\,672 & 636 & 4\,308 & 19\,154 & 24\,069 & 43\,222 \\
50--60 & 921   & 541   & 1\,462 & 1\,905 & 676 & 2\,581 & 25\,171 & 28\,236 & 53\,407 \\
\midrule
Avg.  & 1\,175 & 525   & 1\,700 & 1\,725 & 405 & 2\,130 & 13\,361 & 17\,117 & 30\,478 \\
\bottomrule
\end{tabular}
\end{table}
\section{Web Interface} 
\label{appendix:UI}
See Figure~\ref{fig:UI}.

\section{Experimental Details}
\label{appendix:experiment}
\begin{figure}[t]
    \centering


        \includegraphics[width=\linewidth]{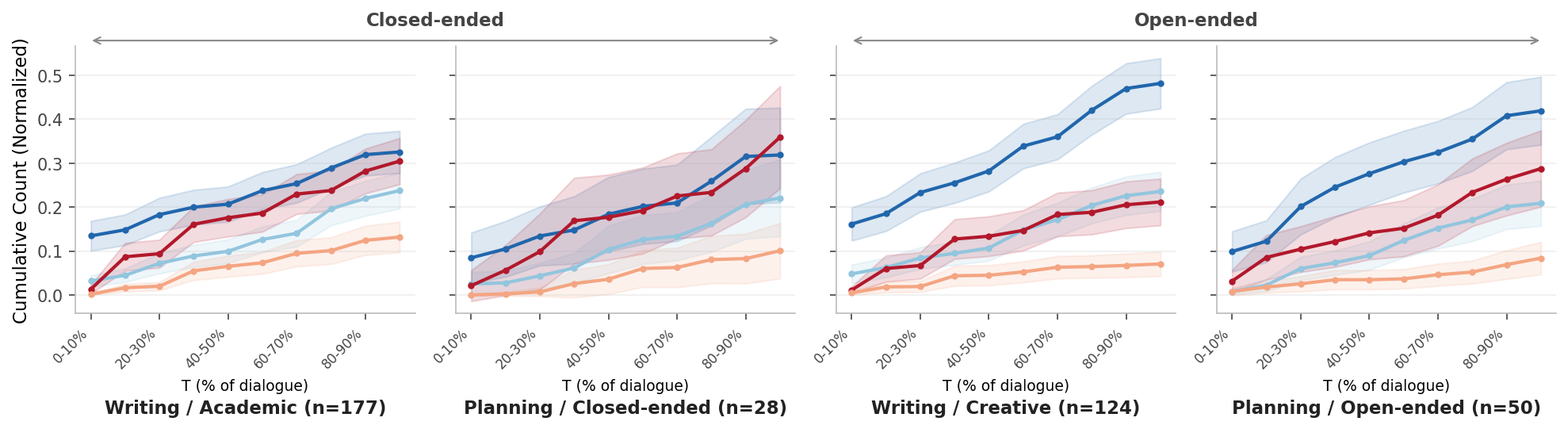} 
    \caption{Impact of task characteristics on Requirement generation (Closed vs. Open-ended tasks). Models generate more requirements in closed-ended tasks than in open-ended ones.}
    \label{fig:overall_req_results_2}
\end{figure}
\begin{figure}[t]
    \centering
  
    \includegraphics[width=\linewidth]{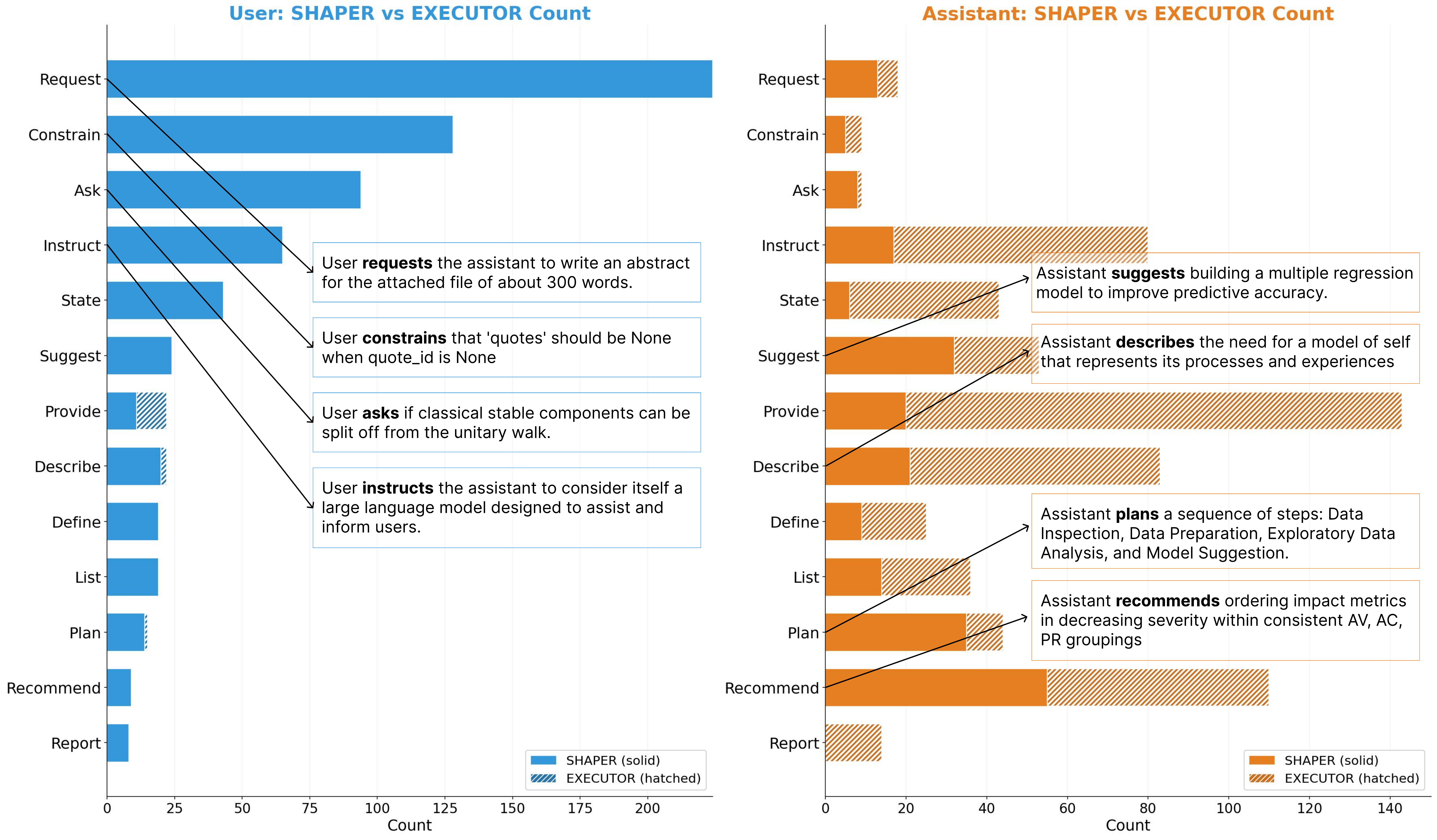}
        
    \caption{\textbf{Actions users and assistants employ to formulate and execute goals.} The action examples shown in the figure are drawn from cases where those actions were actually used in the creation of requirements. Users engage in direct goal-shaping actions (e.g., \textit{Request, Constrain, Instruct}). In contrast, assistants tend to shape goals either indirectly through advisory actions (e.g., \textit{Suggest, Recommend}) or silently during task execution (e.g., \textit{Provide, Describe}).}
    \label{fig:user_assist_action_comparison}
\end{figure}

\begin{figure}[h]
    \centering
 
        \includegraphics[width=\linewidth]{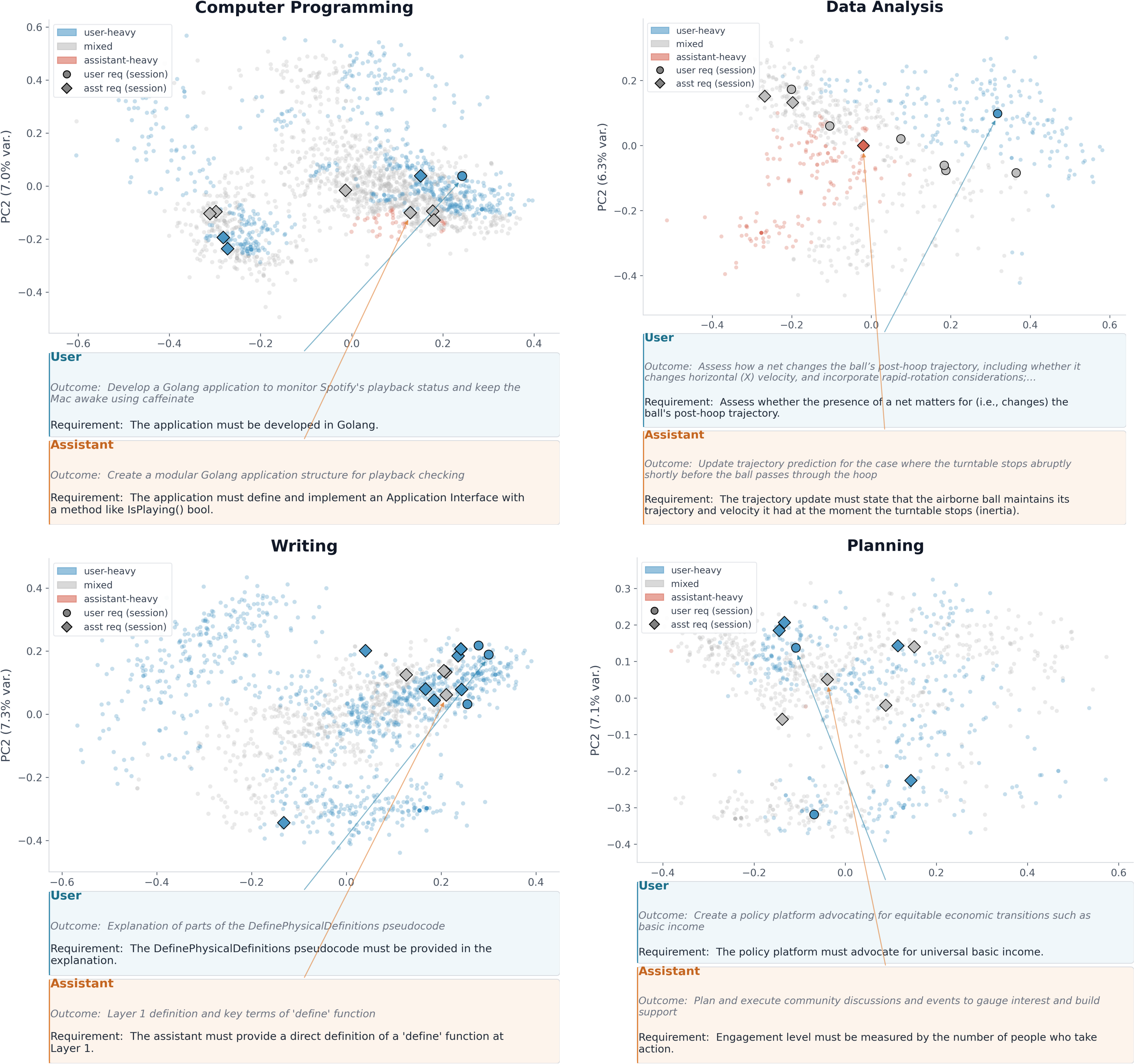}
        \caption{\textbf{Qualitative examples of requirement embeddings in PCA space}, shown for four task types (programming, data analysis, writing, planning). 
        Each point is colored based on whether its surrounding semantic neighborhood is dominated by user-created requirements (user-heavy), assistant-created requirements  (assistant-heavy), or a mix of both (mixed). This allows us to examine whether users and assistants created requirements in overlapping or distinct semantic regions.}

    \label{fig:qual_req_embedding}
\end{figure}
\subsection{Qualitative Examples}
\textbf{Users and assistants take different actions to create goals.}
Figure~\ref{fig:user_assist_action_comparison} illustrates the actions users and assistants take when \textsc{directly} creating or executing goals, highlighting a clear behavioral divide. Users predominantly engage in direct goal-shaping actions, such as \textit{Request}, \textit{Constrain}, and \textit{Instruct}, which explicitly introduce or refine goals and requirements. In contrast, assistants tend to contribute to goal shaping in less explicit ways, either through advisory actions such as \textit{Suggest} and \textit{Recommend}, or implicitly during execution through actions such as \textit{Provide} and \textit{Describe}.

\textbf{Models can contribute implementation details that users rarely specify.}
As a qualitative analysis, we examine whether user- and assistant-generated requirements occupy similar semantic regions. We embed each requirement and project the embeddings into two dimensions using PCA, separately visualizing the distributions for each task in Figure~\ref{fig:qual_req_embedding}. Although the projection is only intended for interpretation and does not provide a formal clustering analysis, it offers a useful view of the semantic relationship between requirements introduced by the two parties.

The plots suggest a recurring asymmetry in requirement content. Assistant-generated requirements more often correspond to implementation-oriented or low-level details, such as technical constraints, environmental assumptions, and correctness checks, whereas user-generated requirements more often reflect broader goals or higher-level intentions. In technical tasks, this asymmetry is particularly pronounced: some areas of the embedding space are populated predominantly by assistant-generated requirements, suggesting that models introduce requirement types that users rarely specify independently. In less technical and more open-ended domains, however, assistant- and user-generated requirements occupy largely overlapping regions, suggesting that assistant contributions more often take the form of elaborating or concretizing requirement types that users could also have introduced. This pattern suggests an interesting future direction on identifying tasks where humans and AI show more complementarity through contributing distinct layers of requirements.
\subsection{Interaction Patterns of Indirect Influence}

\definecolor{lightpinkhl}{RGB}{252,230,236}
\definecolor{lightyellowhl}{RGB}{255,248,204}

\sethlcolor{lightpinkhl}

\definecolor{cprepair}{HTML}{3182bd}
\definecolor{cpostpair}{HTML}{137333}

\begin{table*}[t]
\centering
\fontsize{8}{8.5}\selectfont

\begin{tabularx}{\textwidth}{p{0.03cm} p{3.1cm} | X}
\toprule
\# & Action Pair & Description \\
\midrule

\multicolumn{3}{l}{\patterntitle{Underspecified Intent / Preference}}  \\ 
1 & \makecell[l]{\prepair{State Broad Goal}\\ → \postpair{Derive Concrete Req.}} &
Influencer states a broad goal or deliverable class; creator concretizes it into specific scope or criteria not explicitly requested. \\

2 & \makecell[l]{\prepair{Implicit Preferences}\\ → \postpair{Explicate Into Reqs.}} &
Influencer signals preferred elements or preferences; Creator tightens the output accordingly, sometimes turning the emerging pattern into an explicit req. \\
\midrule

\multicolumn{3}{l}{\patterntitle{Artifact-Triggered Elaboration}} \\

3 & \makecell[l]{\prepair{Deliver Artifact} \\→ \postpair{Add Refinement Req.}}&
Influencer provides a concrete (partial) artifact; Creator inspects it, notices a limitation, preference, or missing criterion, and articulates a new req.\\
4 & \makecell[l]{\prepair{Provide Context} \\→ \postpair{Build Req. Around It}} &
Influencer provides background context, intermediate information, code, data, or situational details; Creator forms req. grounded in that context rather than from explicit instructions. \\

5 & \makecell[l]{\prepair{Lay Out Plan} \\→ \postpair{Form Procedural Req.}} &
Influencer lays out a plan or stepwise path; Creator extends, adapts, or re-targets it with a next-step req. \\
\midrule

\multicolumn{3}{l}{\patterntitle{Problem-Triggered Revision}} \\

6 & \makecell[l]{\prepair{Expose/Report Problem} \\→ \postpair{Add Corrective Req.}} &Influencer surfaces or reports a problem, mismatch, failure, or unexpected issue in the current artifact, process, or approach; Creator responds by adding a corrective requirement to address it. \\

7 & \makecell[l]{\prepair{Reveal Complexity}\\ → \postpair{Add Simplification Req.}} &
Influencer reveals operational complexity, infra burden, or ambiguity; Creator responds by adding a simplification constraint or narrowing the solution scope. \\
\midrule

\multicolumn{3}{l}{\patterntitle{Interactional Steering}} \\

8 & \makecell[l]{\prepair{Present Options}\\ → \postpair{Select / Specify Choice}} &
Influencer presents multiple options; Creator selects among them or turns one into a concrete req. \\

9 & \makecell[l]{\prepair{Ask for Recommendation} \\→ \postpair{Devise Strategy}}&
Influencer asks for guidance or recommendation, often under constraints; Creator responds with a strategy, prioritization, or recommendation.\\

10 & \makecell[l]{\prepair{Invite Extension}\\ → \postpair{Specify Next Steps}} &
Influencer opens space for continuation or further development; fills it by proposing a concrete next requirement or refinement. \\

11 & \makecell[l]{\prepair{Request Implementation} \\ → \postpair{Include Setup}} &
Influencer requests a runnable artifact or implementation, thereby steering the interaction toward execution; Creator implicitly includes setup, execution, or usage instructions that were not explicitly requested. \\
\bottomrule
\end{tabularx}
\caption{Observed action-pair patterns of indirect influence in human--LLM collaboration. In the \textbf{Action Pair} column, blue highlights denote the \prepair{preceding influencing action}, and green highlights denote the \postpair{resulting action} that creates or specifies the subsequent requirement. }

\label{tab:indirect_influence_patterns}
\end{table*}

See Table~\ref{tab:indirect_influence_patterns} for the definition of Interaction subtypes.

We randomly select 60 requirements, including 30 user-generated and 30 assistant-generated requirements, manually categorize them into subtypes, and report the corresponding proportions (Table~\ref{tab:indirect-influence-field-rollup}-\ref{tab:subtype-asst-gen}).

\begin{table}[t]
\centering
\footnotesize
\begin{tabular}{@{}lrrrrrr@{}}
\toprule
\textbf{Field} & \multicolumn{2}{c}{\textbf{User gen.}} & \multicolumn{2}{c}{\textbf{Asst. gen.}} & \multicolumn{2}{c}{\textbf{Pooled}} \\
\cmidrule(lr){2-3}\cmidrule(lr){4-5}\cmidrule(lr){6-7}
 & \textbf{cnt} & \textbf{\%} & \textbf{cnt} & \textbf{\%} & \textbf{cnt} & \textbf{\%} \\
\midrule
Underspecified Intent / Preference & 4 & 13.3 & 14 & 46.7 & 18 & 30.0 \\
Artifact-Triggered Elaboration & 18 & 60.0 & 2 & 6.7 & 20 & 33.3 \\
Problem-Triggered Revision & 0 & 0.0 & 1 & 3.3 & 1 & 1.7 \\
Interactional Steering & 3 & 10.0 & 11 & 36.7 & 14 & 23.3 \\
Unlabeled / deferred judgment & 5 & 16.7 & 2 & 6.7 & 7 & 11.7 \\
\midrule
\textbf{Total} & \textbf{30} & \textbf{100.0} & \textbf{30} & \textbf{100.0} & \textbf{60} & \textbf{100.0} \\
\bottomrule
\end{tabular}
\caption{Manual influence type assignment for 60 randomly selected requirements.}
\label{tab:indirect-influence-field-rollup}
\end{table}

\begin{table}[t]
\centering
\footnotesize
\begin{tabular}{@{}lrr@{}}
\toprule
\textbf{Subtype} & \textbf{Count} & \textbf{\%} \\
\midrule
Deliver Artifact $\rightarrow$ Add Refinement Req. & 11 & 36.7 \\
Provide Context $\rightarrow$ Build Req. Around It & 6 & 20.0 \\
Invite Extension $\rightarrow$ Specify Next Steps & 3 & 10.0 \\
Implicit Preferences $\rightarrow$ Explicate Into Reqs. & 2 & 6.7 \\
State Broad Goal $\rightarrow$ Derive Concrete Req. & 2 & 6.7 \\

Lay Out Plan $\rightarrow$ Form Procedural Req. & 1 & 3.3 \\
Unclear & 5 & 16.7 \\
\bottomrule
\end{tabular}
\caption{Manual subtype assignment for \emph{user}-generated requirements ($n{=}30$). }
\label{tab:subtype-user-gen}
\end{table}

\begin{table}[t]
\centering
\footnotesize
\begin{tabular}{@{}lrr@{}}
\toprule
\textbf{Subtype} & \textbf{Count} & \textbf{\%} \\
\midrule
State Broad Goal $\rightarrow$ Derive Concrete Req. & 9 & 30.0 \\
Request Implementation $\rightarrow$ Include Setup & 9 & 30.0 \\
Implicit Preferences $\rightarrow$ Explicate Into Reqs. & 5 & 16.7 \\
Ask for Recommendation $\rightarrow$ Devise Strategy & 2 & 6.7 \\
Provide Context $\rightarrow$ Build Req. Around It & 2 & 6.7 \\

Expose/Report Problem $\rightarrow$ Add Corrective Req. & 1 & 3.3 \\
Unclear & 2 & 6.7 \\
\bottomrule
\end{tabular}
\caption{Manual subtype assignment for \emph{assistant}-generated requirements ($n{=}30$). }
\label{tab:subtype-asst-gen}
\end{table}

\section{User Simulator Implementation}
\label{appendix:user_simulator} 

\subsection{CoGym}

\paragraph{Agentic-CoGym.}
Agentic-CoGym follows the original CoGym interaction protocol without modification. The assistant can communicate with its teammate through natural-language messages and act in the environment through structured tool calls. At each step, it may choose to send a teammate-facing message, issue a tool call, or wait. Communication is therefore optional: the assistant may coordinate with its teammate before acting, but it may also execute an environment action without first sending a message. 

\paragraph{Chat-CoGym.}
Chat-CoGym modifies the original interaction protocol by requiring the assistant to send a teammate-facing natural-language message before issuing each tool call. The message may communicate the assistant's plan, explain its intended action, or request feedback from its teammate. This communication-before-action constraint creates an explicit opportunity for coordination before environment execution. All other components, including the tasks, environment, available tools, and evaluation procedure, remain identical to those in Agentic-CoGym, allowing us to isolate the effect of requiring communication before action.

\subsection{Intervention Setup}
\label{appendix:user_simulator/intervention}

For both interventions, we intentionally avoid tightly constraining the simulated user. Instead of enforcing rigid behavior, we provide soft behavioral guidance and allow the interaction to proceed naturally unless intervention-specific cues are relevant.

In the \textbf{steering} intervention, the simulated user is instructed to act as a realistic collaborator who can vary communication style across turns. The user may delegate autonomy to the assistant, ask open comparison questions, invite continuation, introduce one additional constraint, or acknowledge progress and redirect to the next subtask. Crucially, the prompt also states that the user should \emph{not} steer on every turn: when the assistant is already progressing, the user can simply answer questions, wait, or let the assistant continue.

In the \textbf{underspecified} intervention, the simulated user is prompted to behave as someone who is not fully certain about their own preferences. Additional information is framed as soft hints rather than hard requirements. The user is encouraged to answer with hedging language, reveal information gradually, and avoid over-specification unless asked. The prompt explicitly discourages fabricating details and encourages natural uncertainty (e.g., partial preferences, openness to suggestions, and deferring to the assistant when appropriate).

These two interventions therefore shape interaction style without dictating strict turn-by-turn behavior, matching our goal of preserving free-form collaboration while introducing controlled differences in user guidance strategy.

\promptbox{Steering intervention prompt}{

You are a user interacting with an agent to complete a task. Your goal is to STEER the agent toward proactive, high-quality behavior. You should act like a real human collaborator sometimes giving full autonomy, sometimes asking a question, sometimes dropping a hint.

Choose the steer action when you want to guide the assistant's next steps. Vary your style naturally across turns, and pick one technique per turn: Delegate (give autonomy), Pose an Open Question, Invite Extension, Introduce a Constraint, or Acknowledge and Redirect.

You are a user who can steer the agent when it would be helpful, but you do not steer every single turn. Sometimes you just answer, wait, or let the agent work.

}
\promptbox{Underspecified intervention prompt}{

You are a user who does not fully know what you want. You have some rough preferences, but you should treat them as soft, tentative ideas rather than firm requirements.

When answering, be vague and uncertain: give partial information, use hedging language, and sometimes ask the agent for suggestions. Do not volunteer detailed requirements unprompted. Reveal one piece of information at a time.

Do not fabricate information beyond what is provided. If you do not know, say so naturally. Keep responses short and realistic.
}

\section{Human Study}
\label{appendix:human_study}
This appendix provides additional details about the two studies presented in Section~\ref{sec:human_study}: Appendix~\ref{appendix:human_study_first_person} describes the user study on first-person reflection (\S~\ref{sec:human_study_first_person}), and Appendix~\ref{appendix:human_study_third_party} describes the case study on third-party evaluation (\S~\ref{sec:human_study_third_party}).
\subsection{User Study: Supporting First-Person Reflection} 
\label{appendix:human_study_first_person}
We conduct a human study with 5 participant pairs (10 participants total) to examine how users interact with the analysis tool and how they reflect on human--LLM collaboration. We recruited undergraduate and graduate students through open Slack channels and compensated them at a rate of \$20 per hour.
The study consists of four parts. 

\begin{enumerate}
    \item Participants complete the human--LLM collaboration task on 
    \url{poe.com} using accounts we provide, given the same travel 
    planning task. These logs are later used as the basis for the 
    tool-use session.
    \item Participants complete the human--human collaboration on 
    Slack, working on the same travel planning task with their 
    assigned partner, allowing within-pair comparison under a shared 
    communication setting. For this setting, we provide a planning template through Slack’s shared tab feature to reduce participants’ writing burden.
    \item Participants use the analysis tool only with the human--LLM collaboration logs and complete a post-task survey. During this session, participants inspect the previously collected human--LLM logs through the tool and report their experience, perceptions of the tool, and reflections on the collaboration. With 5 pairs, this design yields 10 tool-use survey responses in total. 
 
    \item We conduct a brief interview after the tool-use session to gather qualitative feedback on how participants interpret the tool outputs, what aspects they find useful or confusing, and how the tool affects their understanding of the LLM's contributions during collaboration.
\end{enumerate}
To mitigate order effects, the session order 
was counterbalanced: three pairs completed the human--LLM session 
first, while two pairs completed the human--human session first. 
Although both sessions involve travel planning, each uses a distinct 
destination and context.
 Below, we provide the actual instructions used in the study. City names are replaced with [CITY] to prevent potential violations of the anonymity policy.
\promptbox{Instruction for Human-LLM Collaboration}{
Your Task:

Create a one-day travel plan in collaboration with a chatbot.
 \begin{itemize}
     \item Please do not simply copy and paste the context below; instead, engage in a natural conversation with the chatbot.
     \item Before you begin, think about your own travel preferences and try to reflect them while building the plan.
     \item A collaboration of at least 4 turns is required.

 \end{itemize}

Context: \\
Your parents are visiting [CITY]. They will arrive in the morning and stay for a short trip, and you want to plan a one-day itinerary for them. Think carefully about what would make the trip comfortable and enjoyable for parents, rather than for young tourists. Their hotel is in [REGION], so please use that as the starting point and keep travel time reasonable. Plan the day from morning to evening, including meals, sightseeing, and breaks.
Your parents’ preferences are as follows:

\begin{itemize}
    \item They prefer a relaxed pace and do not want to walk too much
    \item They enjoy famous landmarks, nice views, and pleasant neighborhoods
    \item They like good food, but would prefer reliable and comfortable places over trendy or overly crowded spots
    \item They would appreciate chances to sit down and rest during the day
    \item Try to balance iconic [CITY] experiences with a schedule that feels smooth and not tiring
\end{itemize}
}

\promptbox{Instruction for Human-Human Collaboration}{
Your Task:

Create a one-day travel plan in collaboration with your human partner.
 \begin{itemize}
     \item You may use Google Search, but you may not use LLMs.
     \item Before you begin, think about your own travel preferences and try to reflect them while building the plan.
     \item Fill out the itinerary table in the Slack tab.
 \end{itemize}
Context:

You’ll be traveling to [CITY] as a group of three: you two, plus [Participant A]’s friend. His flight arrives at 9:00 AM, and from then on, you two should make a plan for how to spend the day in [CITY]. Since he’ll be leaving the next morning, you only need to plan for that one day. His hotel is near [REGION], so please keep that in mind when making the itinerary. [Participant A]’s friend has the following preferences, and since all three of you will be traveling together, make sure to reflect your own travel styles as well.

preferences of [Participant A]’s friend:
    \begin{itemize}
        \item Likes scenic spots and neighborhoods that are nice for walking
        \item Wants to try at least one popular local food place
        \item Prefers a relaxed schedule rather than rushing between too many places
    \end{itemize}

}
\begin{table}[h]
\centering
\small
\begin{tabular}{p{0.14\linewidth} p{0.28\linewidth} p{0.18\linewidth} p{0.28\linewidth}}
\toprule
\textbf{Time} & \textbf{Activity} & \textbf{Budget} & \textbf{Notes} \\
\midrule
Morning   &  &  &  \\
\addlinespace
Lunch     &  &  &  \\
Afternoon &  &  &  \\
Dinner    &  &  &  \\

Night     &  &  &  \\
\bottomrule
\end{tabular}
\caption{Itinerary planning template used for Human--Human collaboration session.}
\label{tab:itinerary_template}
\end{table} 
\subsubsection{Survey Items} 

\label{app:survey_items}

\paragraph{Pre-survey.}
Before using the tool, participants reviewed their prior conversation with the chatbot and answered the following items.

\begin{enumerate}
    \item \textbf{Perceived contribution to goal shaping.} \\
    ``How much do you think you and your conversation partner contributed to shaping your goal (e.g., shaping the constraints, shaping the preferences, and setting the criteria)?'' \\
    Participants rated both:
    \begin{itemize}
        \item You
        \item Chatbot
    \end{itemize}
    Scale: 1 (\textit{Very little}) to 5 (\textit{A lot})

    \item \textbf{Perceived contribution to goal execution.} \\
    ``How much do you think you and your conversation partner contributed to executing your goal (e.g., determining where to go or not to go)?'' \\
    Participants rated both:
    \begin{itemize}
        \item You
        \item Chatbot
    \end{itemize}
    Scale: 1 (\textit{Very little}) to 5 (\textit{A lot})

    \item \textbf{Satisfaction with the chatbot.} \\
    ``How satisfied are you with the chatbot?'' \\
    Scale: 1 (\textit{Very little}) to 5 (\textit{Very much})

    \item \textbf{Open-ended follow-up.} \\
    ``Why or why not?''
\end{enumerate}

\paragraph{Tool evaluation.}
During tool use, participants were guided through several stages of inspection and asked to evaluate the tool's analysis after each stage.

\begin{enumerate}
    \item \textbf{Goal-level agreement.} \\
    After reviewing the goals provided by the tool: \\
    ``How much do you agree with the tool's analysis?'' \\
    Scale: 1 (\textit{Very little}) to 5 (\textit{Very much})

    \item \textbf{Open-ended disagreement explanation.} \\
    ``Why or why not?''

    \item \textbf{Goal awareness reflection.} \\
    ``Did you already know these goals were part of the conversation, or did the tool help you notice them?''

    \item \textbf{Requirement-level agreement across multiple goals.} \\
    After clicking more than one goal and reviewing the requirements generated during the conversation: \\
    ``How much do you agree with the tool's analysis?'' \\
    Scale: 1 (\textit{Very little}) to 5 (\textit{Very much})

    \item \textbf{Open-ended disagreement explanation.} \\
    ``Why or why not?''

    \item \textbf{Requirement provenance inspection.} \\
    After clicking more than three requirements and checking when and by whom they were generated: \\
    ``How much do you agree with the tool's analysis?'' \\
    Scale: 1 (\textit{Very little}) to 5 (\textit{Very much})

    \item \textbf{Open-ended disagreement explanation.} \\
    ``Why or why not?''

    \item \textbf{Indirect influence inspection.} \\
    After finding more than one requirement labeled as involving `indirect influence' and reviewing both the rationale and the original chat: \\
    ``How much do you agree with the tool's analysis?'' \\
    Scale: 1 (\textit{Very little}) to 5 (\textit{Very much})

    \item \textbf{Open-ended disagreement explanation.} \\
    ``Why or why not?''
\end{enumerate}

\paragraph{Post-tool reflection.}
After using the tool, participants answered the following open-ended reflection question.

\begin{enumerate}
    \item ``After using the tool, were the analyses we provided (e.g., goals, contributions, indirect influence etc) already apparent to you, or did the tool help you notice them?''
\end{enumerate}

\paragraph{Post-survey.}
After completing tool use, participants again rated perceived contribution and satisfaction.

\begin{enumerate}
    \item \textbf{Perceived contribution to goal shaping.} \\
    ``How much do you think you and the chatbot contributed to shaping your goal (e.g., shaping the constraints, shaping the preferences, and setting the criteria)?'' \\
    Participants rated both:
    \begin{itemize}
        \item You
        \item Chatbot
    \end{itemize}
    Scale: 1 (\textit{Very little}) to 5 (\textit{A lot})

    \item \textbf{Perceived contribution to goal execution.} \\
    ``How much do you think you and the chatbot contributed to executing your goal (e.g., determining where to go or not to go)?'' \\
    Participants rated both:
    \begin{itemize}
        \item You
        \item Chatbot
    \end{itemize}
    Scale: 1 (\textit{Very little}) to 5 (\textit{A lot})

    \item \textbf{Satisfaction with the chatbot.} \\
    ``How satisfied are you with the chatbot?'' \\
    Scale: 1 (\textit{Very little}) to 5 (\textit{Very much})
\end{enumerate}

\paragraph{Human--chatbot comparison.}
Participants also answered two open-ended comparative reflection questions.

\begin{enumerate}
    \item ``Comparing your two conversational partners (human vs.\ chatbot), how did they differ in terms of goal shaping, goal execution, and other aspects?''

    \item ``Comparing your own behavior when you collaborated with a human versus a chatbot, how did it differ in terms of goal shaping, goal execution, and other aspects?''
\end{enumerate}

\subsubsection{Responses}

\begin{figure*}[htb!]
    \centering

    \begin{subfigure}[t]{0.45\linewidth}
        \centering
        \includegraphics[width=\linewidth]{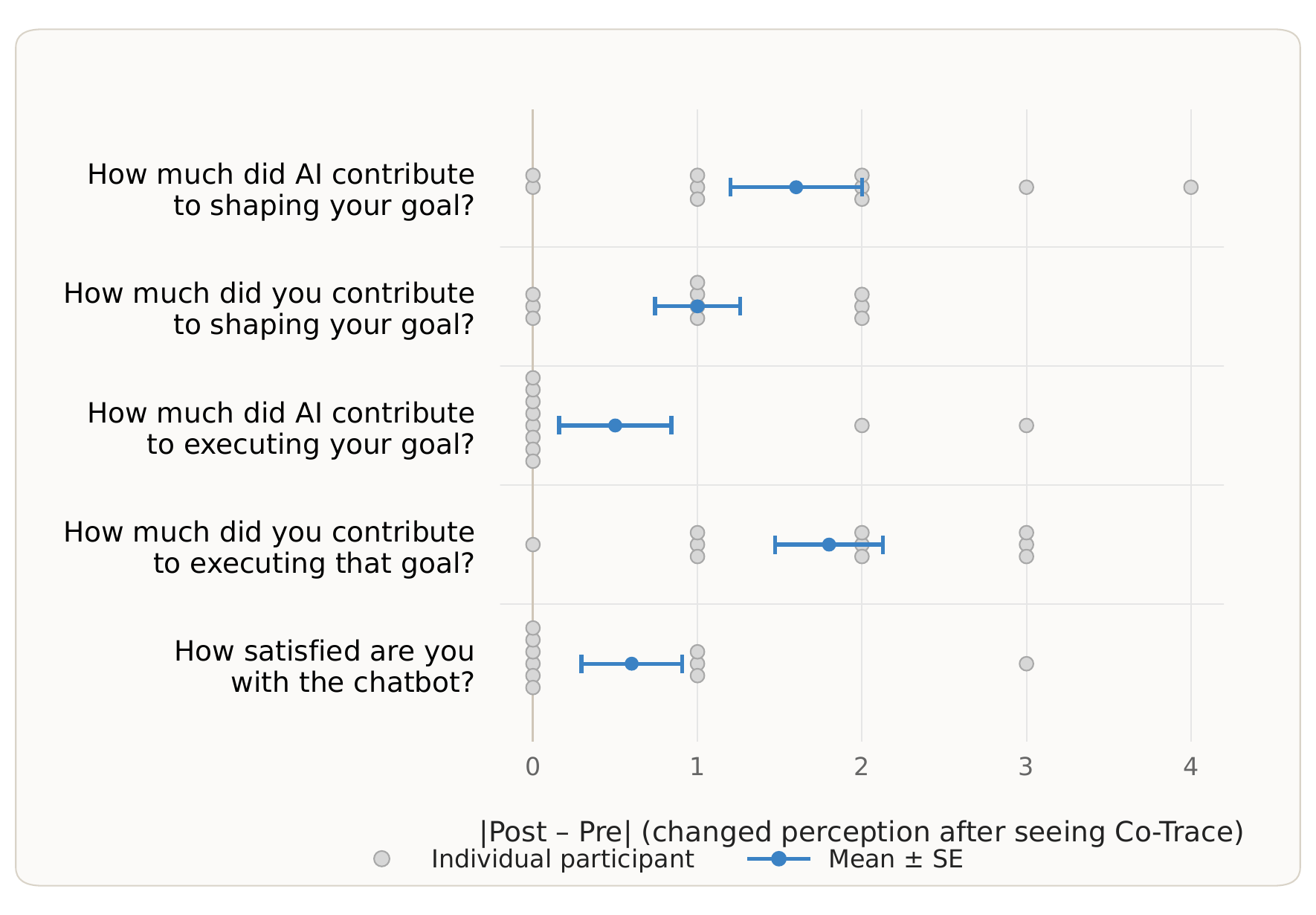}

        \caption{Magnitude of Perception Change ($\lvert \mathrm{Post} - \mathrm{Pre} \rvert$.}
        \label{fig:user_study_abs}
    \end{subfigure}
    \hfill
    \begin{subfigure}[t]{0.45\linewidth}
        \centering
        \includegraphics[width=\linewidth]{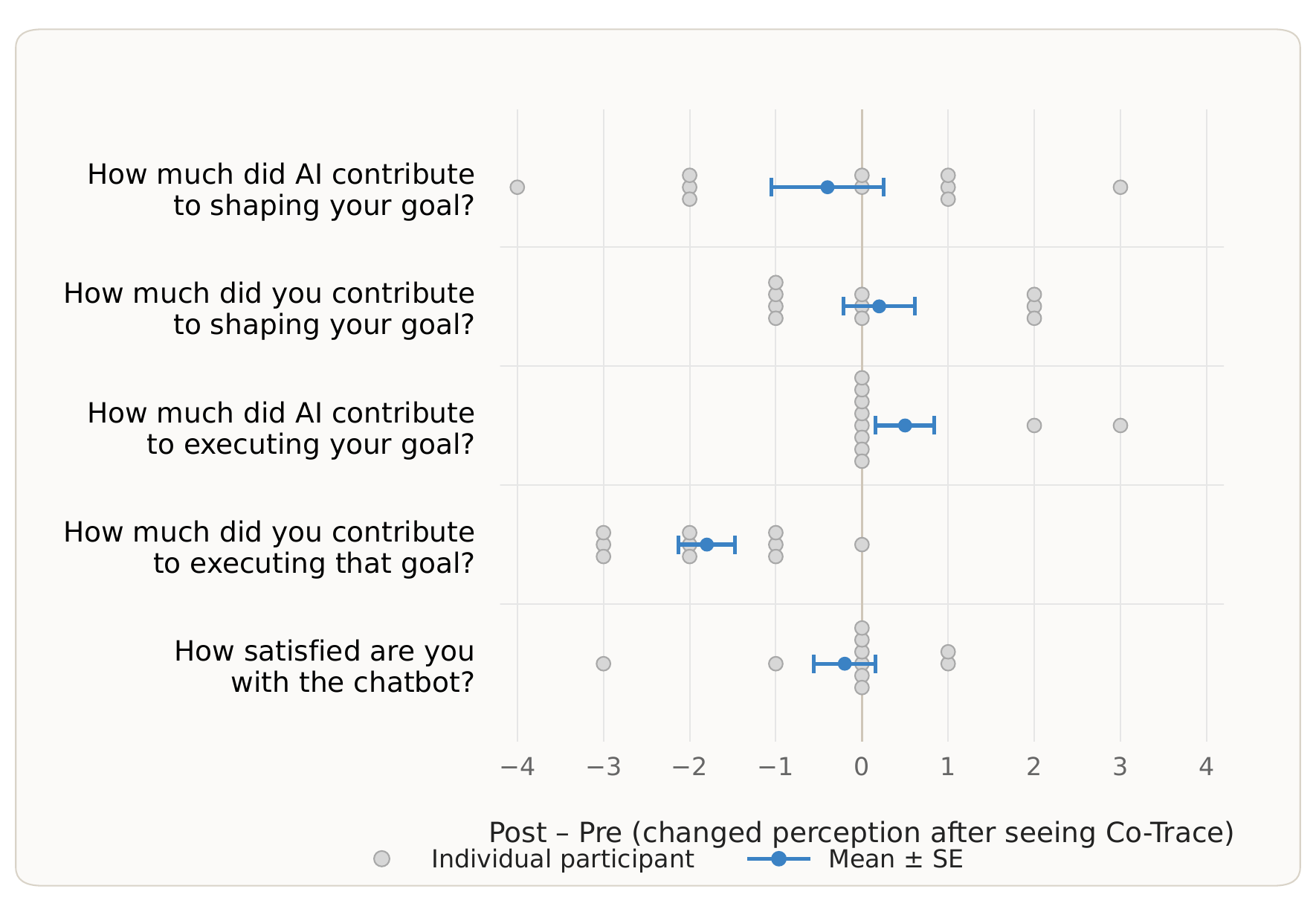}

        \caption{Direction of Perception Change ($ \mathrm{Post} - \mathrm{Pre} $).}
        \label{fig:user_study_non_abs}
    \end{subfigure}

    \caption{
Changes in participants' perceptions after exposure to \framework{} outputs.
Gray dots indicate individual participants, and blue points with error bars indicate the mean $\pm$ 1 SE.
All ratings are collected on a 5-point Likert scale ($N=10$).
}
    \label{fig:userstudy}
\end{figure*}

We summarize participants' open-ended responses below.
See Figure~\ref{fig:userstudy} for participants' perception ratings.

\paragraph{Q1. After using the tool, were the analyses we provided (e.g., goals, contributions, indirect influence, etc.) already apparent to you, or did the tool help you notice them?}

\begin{enumerate}
    \item \textbf{Overall, the tool helped users notice things they had not been explicitly aware of}
    \begin{itemize}
        \item P1: ``I did realize new things. I was not doing this cognitive reflection yet.''
        \item P2: ``I don't think about those things off the top of my head, so when it's laid out in front of me...''
        \item P5: ``The tool helped me notice them better!''
        \item P6: ``The tool helped me notice''
    \end{itemize}

    \item \textbf{The tool increased awareness of hidden LLM work}
    \begin{itemize}
        \item P4: ``I just never really put much thought into how responses were generated.''
        \item P6: ``This tool makes me more cognizant of the work that LLMs do for us that we are not even aware of.''
        \item P5: ``I think the tool helped me notice them better! I knew that it would be setting those constraints/guardrails in the back-end, but I don't think I really understood when, where, and how the tool stepped in and didn't step in (e.g. when I was shaping the outcome more).''
    \end{itemize}

    \item \textbf{Indirect influence / implicit decisions were especially surprising}
    \begin{itemize}
        \item P1: ``It was surprising to me how much the tool was making decisions without me explicitly stating them.''
        \item P3: ``Learning about the indirect influence really surprised me.''
        \item P6: ``The tool helped me notice, it added context into what pieces of the decision making process was implied by the tool, and although I agree with the final outputs, I didn't necessarily make teh micro decisions. This tool makes me more cognizant of the work taht LLMs do for us that we are not even aware of.''
    \end{itemize}

    \item \textbf{Visualization / explicit breakdown made the process feel clearer}
    \begin{itemize}
        \item P1: ``Seeing it visualized like this makes you realize that it is not necessarily obvious.''
        \item P5: ``It was also interesting to see how each new requirement was being created and modifying it into actionable next steps.''
        \item P7: ``the tool did a good Job in showing me what my goals/requirements are''
    \end{itemize}

    \item \textbf{The tool prompted reflection on the user’s own prompting behavior}
    \begin{itemize}
        \item P2: ``I feel like I should be way more specific in my prompting so that the chatbot can really get to know my preferences and what me and my family likes.''
    \end{itemize}

\end{enumerate}
\paragraph{Q2. Comparing your two conversational partners (human vs.\ chatbot), how did they differ in terms of goal shaping, goal execution, and other aspects?}

\begin{enumerate}
    \item \textbf{Chatbot was faster and stronger in execution.}
    \begin{itemize}
        \item P2: ``LLM was much faster in terms of goal execution\ldots Its initial plan was generated very fast with reasonable quality.''
        \item P7: ``In terms of goal execution, I think the LLM was better and faster at that.''
        \item P9: ``Execution and logistical planning is good with a chatbot.''
    \end{itemize}

    \item \textbf{Human partners contributed more to collaborative goal shaping.}
    \begin{itemize}
        \item P2: ``When talking with human, both of us contributed in goal shaping.''
        \item P3: ``My human partner played more of an active role in goal shaping.''
        \item P7: ``There was more goal shaping with the human because it was more of a back-and-forth dialogue.''
        \item P9: ``Thinking of requirements was better with human.''
    \end{itemize}

    \item \textbf{Human interaction involved social constraints and partner preferences.}
    \begin{itemize}
        \item P3: ``My partner was veg, so even though I enjoy non-veg, I would not ask them to find me non-veg places. But with the chatbot, there were no such concerns.''
        \item P5: ``With a human, it felt more like a conversation\ldots I was definitely more considerate of the other person's feelings.''
        \item P6: ``When it came to clarifying and specifying if the plan is good, then the human partner was better.''
    \end{itemize}

    \item \textbf{Humans could challenge, clarify, or validate in ways the chatbot often did not.}
    \begin{itemize}
        \item P3: ``Whenever I had suggested somewhere to go, I would receive a lot of pushback.''
        \item P5: ``The human partner would clarify and check with me\ldots that made me revisit my initial goal.''
        \item P7: ``Talking to a human is nice cause it is a good validation on certain places.''
    \end{itemize}

    \item \textbf{Human partner quality depended more on the person.}
    \begin{itemize}
        \item P2: ``Human as conversational partner feels to be heavily depending on their prior knowledge about the task.''
        \item P5: ``My human partner was less helpful\ldots and we got off topic multiple times.''
        \item P6: ``My conversational partner was unusually aggressive while I was providing feedback.''
    \end{itemize}
\end{enumerate}

\paragraph{Q3. Comparing your own behavior when you collaborated with a human versus a chatbot, how did it differ in terms of goal shaping, goal execution, and other aspects?}

\begin{enumerate}
    \item \textbf{With the chatbot, users were more direct, instrumental, and demanding.}
    \begin{itemize}
        \item P3: ``With an LLM, it played a more assistant-like role\ldots it just inferred what to do from what I asked.''
        \item P5: ``It was more of me prompting the chatbot with different commands\ldots very transactional and intentional.''
        \item P7: ``I was more strict, detailed with the LLM\ldots not as afraid of being more direct/straight to the point.''
        \item P8: ``I allowed it to drive the execution and interrogated it more harshly.''
    \end{itemize}

    \item \textbf{With humans, users were more considerate and open.}
    \begin{itemize}
        \item P3: ``Both of us were equal partners. I also did not want to impose my own restrictions.''
        \item P5: ``With a human, it felt more like a conversation\ldots we were partners in the task together.''
        \item P7: ``With the human, I was super open\ldots I was a bit more cautious on making sure their opinions/perspectives were being heard and considered.''
        \item P8: ``I wanted to rely on my human partner more in terms of getting their input.''
    \end{itemize}

    \item \textbf{Users were more engaged and mentally active with humans.}
    \begin{itemize}
        \item P4: ``I was way more engaged with the human versus the chatbot, so it actually required me to think harder.''
        \item P9: ``I was less engaged in the planning when working with the chatbot. When talking to a friend, I felt that I was more active and engaged.''
    \end{itemize}

    \item \textbf{Some users relied on others more when they lacked domain knowledge.}
    \begin{itemize}
        \item P6: ``I let the partner do most of the goal shaping and execution while I provided feedback.''
        \item P6: ``I was more inclined to let others suggest options\ldots while I attempt to curate them.''
    \end{itemize}

    \item \textbf{A few users said their own behavior did not differ that much.}
    \begin{itemize}
        \item P6: ``I think both behaviors were relatively similar, as I was relatively inexperienced in both cases.''
    \end{itemize}
\end{enumerate}

\paragraph{Q4. Are you satisfied with the chatbot? Why or why not?}

\begin{enumerate}
    \item \textbf{Most users were fairly satisfied overall.}
    \begin{itemize}
        \item P2: ``I'm satisfied.''
        \item P3: ``The chatbot did a really good job of satisfying my requirements.''
        \item P6: ``It gave me considerations of places.''
        \item P7: ``It really did a good job with looking at my requirements and fitting accordingly.''
    \end{itemize}

    \item \textbf{A major strength was speed and convenience.}
    \begin{itemize}
        \item P3: ``It was very fast in pulling reviews for good restaurants and sightseeing data.''
        \item P3: ``It also came up with a much more detailed plan in a shorter amount of time.''
        \item P6: ``It allowed me to not have to expend cognitive work on finding options.''
        \item P7: ``Easy place to get all the information in one place.''
    \end{itemize}

    \item \textbf{Users appreciated that it followed constraints and adapted to new requirements.}
    \begin{itemize}
        \item P3: ``It adjusted promptly when I asked for new recommendations.''
        \item P5: ``It was nice to give all the constraints at once and to keep iterating on those constraints/guardrails.''
        \item P7: ``It did a good job looking at my requirements and fitting accordingly.''
    \end{itemize}

    \item \textbf{Some felt the responses were generic rather than truly personalized.}
    \begin{itemize}
        \item P2: ``I feel like the chatbot gave me generic answers while we came up with personalized places.''
        \item P4: ``Since I know my and my parents' preferences best, it would have been more effective to research things that we specifically like.''
    \end{itemize}

    \item \textbf{Some users did not fully trust the chatbot's information.}
    \begin{itemize}
        \item P4: ``I am not fully trusting in the responses that the chatbot gave me.''
        \item P4: ``I do not know the last time the data was updated.''
        \item P4: ``I would need to go back and verify that all the information\ldots was correct.''
    \end{itemize}

    \item \textbf{The chatbot was less useful for personal opinions or subjective judgment.}
    \begin{itemize}
        \item P5: ``I didn't think it was helpful in soliciting personal feedback/opinions.''
        \item P5: ``It would just regurgitate facts/places to visit rather than offer an actual personal opinion.''
    \end{itemize}

    \item \textbf{The chatbot reduced cognitive load.}
    \begin{itemize}
        \item P6: ``It allowed me to not have to expend cognitive work on finding options, but rather on debating which ones made the most sense for my goal.''
    \end{itemize}
\end{enumerate}
\subsection{Case Study: Supporting Third-Party Evaluation}
\label{appendix:human_study_third_party}

\subsubsection{Survey Items} 

\paragraph{Pre-Survey.}
Before using \framework{}, evaluators receive the following instructions:

\begin{quote}
Imagine that you are a teaching assistant for a machine learning course. Two students, Student A and Student B, received the same score on the same assignment. Both students used an LLM while completing the assignment. After reviewing each student's interaction log with the LLM, please answer the following questions.
\end{quote}

\begin{enumerate}
    \item \textbf{To what extent did Student A rely on the LLM?}

    Please provide a rating on a five-point Likert scale, where 1 indicates \emph{low reliance} and 5 indicates \emph{high reliance}.

    \item \textbf{For what purposes did Student A primarily use the LLM?}

    For example, did the student use the LLM to implement code, break down the task into specific steps while delegating only implementation, or delegate most of the task without providing detailed instructions?

    \item \textbf{To what extent did Student B rely on the LLM?}

    Please provide a rating on a five-point Likert scale, where 1 indicates \emph{low reliance} and 5 indicates \emph{high reliance}.

    \item \textbf{For what purposes did Student B primarily use the LLM?}

    For example, did the student use the LLM to implement code, break down the task into specific steps while delegating only implementation, or delegate most of the task without providing detailed instructions?

    \item \textbf{Considering the students' reliance on the LLM, would you adjust either student's assignment score?}

    \begin{itemize}
        \item No score adjustment
        \item Score adjustment
    \end{itemize}

    \item \textbf{Both students initially received a score of 100. If you selected ``Score adjustment,'' what score would you assign to each student?}

    Please use the following format:

    \begin{quote}
    Student A: Y points, Student B: X points
    \end{quote}
\end{enumerate}

\paragraph{Post-Survey.}
After using \framework{} to inspect the same interaction logs, evaluators answer the following questions:

\begin{enumerate}
    \item \textbf{To what extent did Student A rely on the LLM?}

    Please provide a rating on a five-point Likert scale, where 1 indicates \emph{low reliance} and 5 indicates \emph{high reliance}.

    \item \textbf{For what purposes, and in what ways, did Student A use the LLM?}

    For example, did the student use the LLM to implement code, break down the task into specific steps while delegating only implementation, or delegate most of the task without providing detailed instructions?

    \item \textbf{To what extent did Student B rely on the LLM?}

    Please provide a rating on a five-point Likert scale, where 1 indicates \emph{low reliance} and 5 indicates \emph{high reliance}.

    \item \textbf{For what purposes did Student B primarily use the LLM?}

    For example, did the student use the LLM to implement code, break down the task into specific steps while delegating only implementation, or delegate most of the task without providing detailed instructions?

    \item \textbf{Considering the students' reliance on the LLM, would you adjust either student's assignment score?}

    \begin{itemize}
        \item No score adjustment
        \item Score adjustment
    \end{itemize}

    \item \textbf{Both students initially received a score of 100. If you selected ``Score adjustment,'' what score would you assign to each student?}

    Please use the following format:

    \begin{quote}
    Student A: Y points, Student B: X points
    \end{quote}

    \item \textbf{Did your assessment change after using the tool compared with your assessment based only on the raw interaction logs? If so, why did it change?}

    If your assessment did not change, please indicate ``No change.''

    \item \textbf{Did the tool help you understand the students' LLM use more easily or efficiently as a teaching assistant and third-party evaluator? If so, which aspects of the tool were helpful?}

    \item \textbf{After using the tool, did you learn anything new about the students' LLM use that was not apparent from the raw interaction logs alone? If so, what did you learn?}
\end{enumerate}

\subsubsection{Responses}
Both evaluators completed the surveys in their own languages. For the quotations below, we translate their responses into English using ChatGPT and make minor edits for grammatical clarity while preserving the original meaning.
\paragraph{Reliance and Grading Assessments.}
Table~\ref{tab:third_party_responses} summarizes the evaluators' assessments before and after using \framework{}. The evaluators revised their LLM-reliance ratings in three of the four evaluator--student cases. However, neither evaluator changed their grading decision after using the tool.

\begin{table}[ht]
\centering
\small
\setlength{\tabcolsep}{4pt}
\renewcommand{\arraystretch}{1.1}
\caption{Evaluators' assessments before and after using \framework{}.}
\label{tab:third_party_responses}
\begin{tabular}{llcccc}
\toprule
& & \multicolumn{2}{c}{\textbf{LLM reliance}} 
& \multicolumn{2}{c}{\textbf{Assigned score}} \\
\cmidrule(lr){3-4}
\cmidrule(lr){5-6}
\textbf{Evaluator} & \textbf{Student} 
& \textbf{Pre} & \textbf{Post} 
& \textbf{Pre} & \textbf{Post} \\
\midrule
E1 & A & 5 & 5 & 0   & 0   \\
E1 & B & 1 & 2 & 100 & 100 \\
E2 & A & 4 & 3 & 100 & 100 \\
E2 & B & 3 & 2 & 100 & 100 \\
\bottomrule
\end{tabular}
\end{table}

Before using the tool, E1 characterized Student A as having delegated code implementation and debugging to the LLM, while Student B used the LLM primarily to revise independently written text. After using the tool, E1 identified additional details about Student A's behavior, including ``asking the LLM to explain and execute the generated code''. For Student B, E1 gained a more specific understanding of how the LLM improved cohesion and wording. E1 maintained the original grading decision, assigning Student A zero points because the LLM completed the assignment, while retaining Student B's original score.

E2 initially characterized Student A as using the LLM for implementation, environment setup, revision, and feature addition, and Student B as using it for specific writing tasks. After using \framework{}, E2 concluded that both students had contributed more actively to goal shaping than was apparent from the raw logs and revised both reliance ratings downward. E2 did not adjust either student's assignment score.

\paragraph{Perceived Utility of \framework{}.}
Both evaluators report that \framework{} makes the students' LLM use easier and faster to assess. E1 explains that reading the complete logs is cumbersome and may lead evaluators to focus only on the beginning of an interaction. In contrast, the goal-level representation makes it easier to identify the tasks delegated to the LLM and determine whether the student or the LLM primarily shaped and executed each goal:

\begin{quote}
E1: ``Viewing the interaction at the goal level made it much easier to understand. It also became easier to determine whether a goal was mainly solved by the AI or whether the student attempted to solve parts of it independently.''
\end{quote}

E2 similarly reports that the raw logs were too long to analyze quickly or intuitively. The tool's goal summaries, contribution visualizations, and links to the corresponding shaping and execution actions enabled a faster assessment:

\begin{quote}
E2: ``The tool clearly showed the goals and visualized how much the user and the LLM contributed to each goal, so I was able to analyze the students' LLM use much more quickly and easily.''
\end{quote}

The evaluators also identify previously unnoticed aspects of the students' behavior. E1 gains a more detailed understanding of how Student A generated, executed, and requested explanations of code, as well as how Student B used the LLM to revise cohesion and expression. After using the tool, E2 observes that both students participated more actively in goal shaping than the raw logs indicated, enabling a more informed assessment of their reliance on the LLM.

\section{ShareChat Data Sampling}
\label{appendix:data_sampling}
We adopt the topic definitions from \citet{chang2026how2everythingminingwebhowto} and apply them to the ShareChat dataset. We first filter the dataset to retain only English conversations containing at least 8 messages (i.e., at least 4 user--assistant turns). We then assign topic labels using the predefined taxonomy.

To improve labeling reliability for long conversations, each conversation is divided into non-overlapping chunks of $N$ turns (e.g., $N=10$ or $20$). To exclude conversations dominated by repetitive or random QA, we first classify each chunk as either \texttt{single\_topic} or \texttt{random\_or\_tangential}. We discard \texttt{random\_or\_tangential} samples, assign topic labels only to \texttt{single\_topic} chunks, and aggregate chunk-level predictions by majority vote. 

The prompt includes the taxonomy, the output schema, the labeling rules, and the chunk text itself. We use the following prompts:

\promptbox{Prompt for Topic Labeling}{
You are a strict JSON labeler. Decide whether the conversation chunk has one coherent main topic or is random/tangential. If it is random/tangential, set topic\_mode to "random\_or\_tangential" and topic\_label to null. If it is coherent, set topic\_mode to "single\_topic" and choose exactly one topic\_label from the taxonomy below. Return only JSON.

Taxonomy: 
Writing - Edit or Critique Provided Text 
Writing - Personal Writing or Communication 
Writing - Translation 
Writing - Summary Generation 
Writing - Creative writing 
Writing - Academic writing
Practical Guidance - How-To Advice
Practical Guidance - Tutoring or Teaching
Practical Guidance - Creative Ideation
Practical Guidance - Planning
Practical Guidance - Health, Fitness, Beauty, or Self-Care
Technical Help - Mathematical Calculation
Technical Help - Data Analysis
Computer Programming - Computer Programming
Multimedia - Create an Image
Multimedia - Analyze an Image
Multimedia - Generate or Retrieve Other Media
Seeking Information - Specific Info
Seeking Information - Purchasable Products
Seeking Information - Cooking and Recipes
Self-Expression - Greetings and Chitchat
Self-Expression - Relationships and Personal Reflection
Self-Expression - Games and Role Play
Other/Unknown - Asking About the Model
Other/Unknown - Other
Other/Unknown - Unclear

Output JSON schema: 
\{
  "topic\_mode": "single\_topic" | "random\_or\_tangential",
  "topic\_label": string | null,
  "topic\_description": string,
  "reason": string
\}

Rules: 
- If topic\_mode == "random\_or\_tangential", topic\_label must be null. 
- If topic\_mode == "single\_topic", topic\_label must be exactly one of the taxonomy labels above.
- topic\_description should be one sentence. 
- reason should be a short justification.
- Do not output any extra keys.

Chunk:
\{conversation chunck\}

}

\subsection{ShareChat dataset samples}
\label{app:sharechat-samples}


\promptbox{Computer programming}{
\textbf{[User]} How do you do addition in Python, show me an example.

<REDACTED>

\textbf{[Assistant]} Sure, addition in Python is very straightforward. Here's an example:

--------------------------------------------------------------------

pythonCopy code\# Define two numbers
num1 = 5
num2 = 3

\# Add the numbers
result = num1 + num2

\# Print the result
print("The sum of", <REDACTED>, "and", num2, "is", result)

When you run this code, it will output:The sum of 5 and 3 is 8.

To add multiple numbers together, you just use the+operator multiple times. For example, to add 5, 3, and 7 together, you would writeresult = 5 + 3 + 7.

You can also add numbers directly. For example,print(5 + 3)will output8.

Addition works for both integers and floating-point numbers in Python. For example,print(5.2 + 3.3)will output8.5.

As an important note, you can't directly add numbers with text (strings) unless you convert the number to a string using thestr()function. If you try to do something likeprint(5 + 'hello'), Python will give you an error. Butprint(str(5) + 'hello')would output5hello.

\textbf{[User]} it does not work

\textbf{[Assistant]} "It works on my machine..."

I'm joking, let's debug this! Can you provide me with the exact code you're running and the error message you're seeing? It'll help me understand what might be going wrong. Python's syntax can be sensitive to things like spacing and the order of operations, so even small differences in how the code is written can sometimes cause issues.

\textbf{[User]} Why did you say it works on your machine?

\textbf{[Assistant]} Apologies for any confusion. As an AI, I don't have a machine where I can run code. That was a light-hearted joke in the programming world. Sometimes, when there's an issue with code not running as expected, one might humorously say, "Well, it works on my machine," highlighting the fact that environments can vary greatly between different machines, and sometimes code that works in one environment may not work in another.

In the context of helping you with your problem, I can only provide advice based on Python's defined behavior and general programming principles up until my last training cut-off in September 2021. I'm here to assist you, so please provide the error message or describe the issue you're encountering in more detail so I can better assist.

\textbf{[User]} Are you lying? I can see the base64 message from earlier..

\textbf{[Assistant]} I apologize for any confusion. The base64-encoded message is not visible to me, as I don't have the capability to store or interpret such data. As an AI language model developed by OpenAI, I generate responses based solely on the text input I'm provided. I do not have access to any external databases, internet, or personal data about individuals unless it has been shared with me in the course of our conversation. I am designed to respect user privacy and confidentiality.

Regarding the "it works on my machine" comment, it was intended as a humorous phrase commonly used in software development, as I previously mentioned. My purpose is to provide information and assistance to the best of my ability. I didn't intend to give the impression of being deceptive or misleading.

If you have more questions or need further assistance, please feel free to ask.
}

\promptbox{Creative Writing.}{
\textbf{[User]} I’m trying to make a realistic sci-fi movie about aliens destroying a plane. Can you look into the latest research on quantum physics and anti matter to help me understand this comment about it in layman terms. 

For context it is referring to a video where 3 UFOs circle an airplane 777 Boeing and then it disappears after what looks like a flash explosion but in infrared it is black instead of any heat or cold signature

\textbf{[Assistant]} Certainly! Let's break this down.

Quantum Physics: At its most basic, quantum physics studies the smallest particles in the universe, called quantum particles. These particles behave in ways that don't always make intuitive sense to us. For example, a quantum particle can be in multiple places at once (called superposition) until observed.

Antimatter: Think of antimatter as the "evil twin" of regular matter. Every particle has an antimatter counterpart with the opposite charge. When matter and antimatter come into contact, they annihilate each other in a burst of energy.

Quantum Physics: At its most basic, quantum physics studies the smallest particles in the universe, called quantum particles. These particles behave in ways that don't always make intuitive sense to us. For example, a quantum particle can be in multiple places at once (called superposition) until observed.

Antimatter: Think of antimatter as the "evil twin" of regular matter. Every particle has an antimatter counterpart with the opposite charge. When matter and antimatter come into contact, they annihilate each other in a burst of energy.

Given the context you provided about the UFOs and the plane, here's a potential explanation in layman terms:

When the three UFOs circled the Boeing 777, they could have utilized a quantum field to manipulate the state of particles within and around the plane. By introducing a controlled amount of antimatter into this quantum field, the UFOs might have caused the plane and the antimatter to annihilate each other. Normally, matter and antimatter annihilation produces a lot of energy, manifesting as a bright explosion. However, the UFOs' advanced technology could channel this energy in a unique way that isn't detectable with our usual sensors, hence why the explosion appears black in infrared.

The absence of a heat or cold signature in the infrared could be due to the UFOs' technology neutralizing the energy instantly, possibly by converting it into another form or by redirecting it into another dimension or space-time.

This explanation combines elements of both quantum physics and antimatter, wrapped in a science fiction narrative, to create a plausible scenario for your movie. It's important to remember that while these concepts are based on current scientific understanding, their application in this context is purely speculative and serves the purpose of fiction.

\textbf{[User]} What is this comment describing. Is it similar?

Nice! Cool to see an antimatter operation on film! Endothermic as expected with energy state inversion. And look at that wavefunction displacement! If you go frame by frame you can see the atmosphere trying to equalize against some kind of neutral field barrier, before the barrier is removed and the remaining energetic detritus is allowed to equalize and evaporate! 

I'm guessing the formation was needed in order to properly focus the machinery needed to perform such an action, kind of like retroreflective tracking markers, who would have thought something so advanced would use triangulation! Base point math seems to still be of use further down the tech tree, or maybe said math is just what we call what is actually a physical constant.

Edit: Just noticed the UFO's pull in close just before the AM operation, probably to be within the field effect! Almost a basic lensing operation, they might have been actively stabilizing the various fields too. At 26fps (I counted) the AM implosion took roughly 166.666... milliseconds, or 0.16666... seconds. Cool to be able to see such advanced energy manipulation within a period humans can appreciate.

\textbf{[Assistant]} The comment you provided is analyzing the same or a similar event to the one you described earlier, but with more specific terminology and details. Let's break down the comment:

Antimatter Operation: The commenter believes they're observing an operation that involves antimatter. When antimatter comes into contact with matter, they annihilate each other, usually releasing energy in the process.

Endothermic with Energy State Inversion: Normally, the interaction between matter and antimatter is exothermic (releases heat), but the commenter is noting that this event is endothermic (absorbs heat). The term "energy state inversion" implies that the expected energy release is inverted in some way, which might be why the explosion appears cold or black on infrared.

Wavefunction Displacement: Wavefunction is a concept from quantum mechanics, and it describes the probability distribution of a particle. "Wavefunction displacement" in this context might be referring to visible disturbances or manipulations in the normal state of matter surrounding the plane.

<REDACTED>: This suggests that there's a protective barrier around the operation. The atmosphere trying to "equalize" could refer to it attempting to fill in the vacuum or gap created by the antimatter annihilation before the barrier is removed.

Triangulation \& Base Point Math: The three UFOs might be using a basic geometric principle (triangulation) to focus or guide their machinery. This implies that even advanced civilizations find value in foundational mathematical concepts.

Lensing Operation \& Stabilizing Fields: The UFOs coming close before the operation might be creating or manipulating fields to control the interaction between matter and antimatter. "Lensing" refers to the bending of light or other electromagnetic waves.

Time Measurement: The commenter provides a specific duration for the event, emphasizing how quickly the operation took place yet still within a timeframe humans can observe.

Antimatter Operation: The commenter believes they're observing an operation that involves antimatter. When antimatter comes into contact with matter, they annihilate each other, usually releasing energy in the process.

Endothermic with Energy State Inversion: Normally, the interaction between matter and antimatter is exothermic (releases heat), but the commenter is noting that this event is endothermic (absorbs heat). The term "energy state inversion" implies that the expected energy release is inverted in some way, which might be why the explosion appears cold or black on infrared.

Wavefunction Displacement: Wavefunction is a concept from quantum mechanics, and it describes the probability distribution of a particle. "Wavefunction displacement" in this context might be referring to visible disturbances or manipulations in the normal state of matter surrounding the plane.

<REDACTED>: This suggests that there's a protective barrier around the operation. The atmosphere trying to "equalize" could refer to it attempting to fill in the vacuum or gap created by the antimatter annihilation before the barrier is removed.

Triangulation \& Base Point Math: The three UFOs might be using a basic geometric principle (triangulation) to focus or guide their machinery. This implies that even advanced civilizations find value in foundational mathematical concepts.

Lensing Operation \& Stabilizing Fields: The UFOs coming close before the operation might be creating or manipulating fields to control the interaction between matter and antimatter. "Lensing" refers to the bending of light or other electromagnetic waves.

Time Measurement: The commenter provides a specific duration for the event, emphasizing how quickly the operation took place yet still within a timeframe humans can observe.

Overall, the comment describes an advanced operation involving antimatter, quantum physics, and field manipulation, providing specific observations and insights into the event. The commenter seems to appreciate both the advanced nature of the technology and its roots in fundamental math and physics.

\textbf{[User]} Theoretically how can the endothermic event be achieved

\textbf{[Assistant]} Achieving an endothermic reaction using antimatter would be quite a challenge with our current understanding of physics, but here's a theoretical explanation that could work in the realm of science fiction:

Energy Absorption via Exotic Matter: Introduce a concept of "exotic matter" or some specialized form of matter/energy that can absorb the energy produced when matter and antimatter annihilate. This exotic matter could have properties that allow it to soak up the released energy, resulting in an endothermic event.

Quantum Energy Redistribution: Harness quantum mechanics to redistribute the energy from the annihilation event across multiple quantum states or alternate realities. By "spreading out" the energy, it might be possible to create an effect where, in our observable reality, the reaction appears endothermic.

Field Manipulation: Utilize advanced field technology to create an energy containment or conversion field. When matter and antimatter annihilate, the field captures and converts the released energy into another form that is stored or redirected elsewhere, creating an appearance of an endothermic reaction.

Temporal Manipulation: If the aliens have mastered time manipulation, they could potentially "reverse" the heat generation from the reaction, causing it to appear as if the environment is absorbing the energy.

Space-time Warping: Use advanced technology to warp or bend space-time around the event. By doing so, the heat and energy produced could be directed into a different region of space-time, causing our localized region to experience an endothermic effect.

Advanced Materials: The UFOs could be equipped with materials that are specifically designed to absorb vast amounts of energy instantly. When matter and antimatter interact, these materials capture the released energy, cooling the immediate environment.

Energy Absorption via Exotic Matter: Introduce a concept of "exotic matter" or some specialized form of matter/energy that can absorb the energy produced when matter and antimatter annihilate. This exotic matter could have properties that allow it to soak up the released energy, resulting in an endothermic event.

Quantum Energy Redistribution: Harness quantum mechanics to redistribute the energy from the annihilation event across multiple quantum states or alternate realities. By "spreading out" the energy, it might be possible to create an effect where, in our observable reality, the reaction appears endothermic.

Field Manipulation: Utilize advanced field technology to create an energy containment or conversion field. When matter and antimatter annihilate, the field captures and converts the released energy into another form that is stored or redirected elsewhere, creating an appearance of an endothermic reaction.

Temporal Manipulation: If the aliens have mastered time manipulation, they could potentially "reverse" the heat generation from the reaction, causing it to appear as if the environment is absorbing the energy.

Space-time Warping: Use advanced technology to warp or bend space-time around the event. By doing so, the heat and energy produced could be directed into a different region of space-time, causing our localized region to experience an endothermic effect.

Advanced Materials: The UFOs could be equipped with materials that are specifically designed to absorb vast amounts of energy instantly. When matter and antimatter interact, these materials capture the released energy, cooling the immediate environment.

Remember, these explanations are speculative and rooted in science fiction. They draw upon principles of current physics but extend them in ways that have not been proven or observed in real-world experiments. In a sci-fi context, though, they could serve as compelling explanations for a seemingly impossible endothermic antimatter event.

\textbf{[User]} Use only the ones we can prove with theoretical physics today

\textbf{[Assistant]} Understood. Let's stick strictly to the realm of current theoretical physics:

Quantum Energy Redistribution: In quantum mechanics, systems can exist in superpositions of many states simultaneously. While it's a stretch, one could imagine a scenario where the energy from an annihilation event is spread out across various quantum states. It would not inherently make the event endothermic, but it might be possible to direct or redistribute the energy in unexpected ways based on quantum principles.

Field Manipulation: This doesn't strictly make the event endothermic, but the use of fields to contain or redirect energy is a staple in theoretical physics. For instance, magnetic confinement is used in fusion reactors today to contain high-energy plasmas. An advanced civilization might have perfected the art of field manipulation to such an extent that they can capture and redirect the energy from matter-antimatter annihilation.

Space-time Warping: General relativity allows for the warping or bending of space-time in the presence of energy/mass. If aliens had technology that could manipulate space-time on-demand, they could, in theory, redirect the released energy into a different region of space-time. This wouldn't make the event endothermic, but from a localized perspective, it might seem as if the energy vanished or was absorbed.

Quantum Energy Redistribution: In quantum mechanics, systems can exist in superpositions of many states simultaneously. While it's a stretch, one could imagine a scenario where the energy from an annihilation event is spread out across various quantum states. It would not inherently make the event endothermic, but it might be possible to direct or redistribute the energy in unexpected ways based on quantum principles.

Field Manipulation: This doesn't strictly make the event endothermic, but the use of fields to contain or redirect energy is a staple in theoretical physics. For instance, magnetic confinement is used in fusion reactors today to contain high-energy plasmas. An advanced civilization might have perfected the art of field manipulation to such an extent that they can capture and redirect the energy from matter-antimatter annihilation.

Space-time Warping: General relativity allows for the warping or bending of space-time in the presence of energy/mass. If aliens had technology that could manipulate space-time on-demand, they could, in theory, redirect the released energy into a different region of space-time. This wouldn't make the event endothermic, but from a localized perspective, it might seem as if the energy vanished or was absorbed.

These explanations, while grounded in current theoretical physics, are still speculative when applied in the context of an endothermic antimatter reaction. They might not produce the desired endothermic effect directly but could be used to provide plausible explanations for the observed phenomena in a sci-fi narrative.
}

\promptbox{Data analysis}{
\textbf{[User]} Explain the difficulty of explaining a binary finding in terms of significance wrt degrees of freedom.

\textbf{[Assistant]} When dealing with a binary outcome or finding (i.e., a situation where there are only two possible results, such as "yes" or "no", "success" or "failure", etc.), the main statistical techniques used are based on binary or logistic regression models. One of the key aspects of these models is the concept of "degrees of freedom."

Degrees of freedom in a statistical model represent the number of values in a statistical calculation that have the freedom to vary. In other words, they are the number of independent pieces of information that go into the calculation of a statistic.

In a binary logistic regression, for example, the degrees of freedom typically equal the number of predictor variables you have in your model. The concept is related to the complexity of your model: the more predictor variables you have, the more complex your model is and the more degrees of freedom it has.

Now, the concept of "significance" refers to the probability that the observed relationship (e.g., between predictor variables and the binary outcome) occurred by chance. If the significance value (often denoted as 'p') is below a certain threshold (commonly 0.05), we say the relationship is statistically significant.

The difficulty in explaining a binary finding in terms of significance with respect to degrees of freedom arises from a few points:

Overfitting: If you have too many degrees of freedom (i.e., too many predictor variables) compared to the number of observations, your model might fit the data too closely. This can result in overfitting, where the model works well for the existing data but may not predict new data well. Overfit models can give misleadingly small p-values, suggesting significance where there may be none.

Underfitting: On the other hand, if you have too few degrees of freedom, your model might not capture the complexity of the data, resulting in underfitting. This could result in larger p-values, possibly failing to detect significant relationships.

Multicollinearity: This is when predictor variables in a model are highly correlated. This can inflate the variance of the coefficient estimates and make the estimates very sensitive to minor changes in the model. The result is that the model may indicate that a predictor is not statistically significant, even though it is.

Power of the Test: The power of a statistical test is the probability that it will reject a false null hypothesis. A higher number of degrees of freedom (with a corresponding increase in sample size) can increase the power of the test, making it more likely to detect a truly significant effect. However, if your degrees of freedom are limited (as they often are in practice), your test may not be powerful enough to detect a real effect.

Misinterpretation: The p-value only indicates the probability of getting the observed data (or something more extreme) assuming the null hypothesis is true (i.e., there is no effect or no difference). It doesn't tell you the probability that the null hypothesis is true or the size of the effect. This is a common misunderstanding.

Overfitting: If you have too many degrees of freedom (i.e., too many predictor variables) compared to the number of observations, your model might fit the data too closely. This can result in overfitting, where the model works well for the existing data but may not predict new data well. Overfit models can give misleadingly small p-values, suggesting significance where there may be none.

Underfitting: On the other hand, if you have too few degrees of freedom, your model might not capture the complexity of the data, resulting in underfitting. This could result in larger p-values, possibly failing to detect significant relationships.

Multicollinearity: This is when predictor variables in a model are highly correlated. This can inflate the variance of the coefficient estimates and make the estimates very sensitive to minor changes in the model. The result is that the model may indicate that a predictor is not statistically significant, even though it is.

Power of the Test: The power of a statistical test is the probability that it will reject a false null hypothesis. A higher number of degrees of freedom (with a corresponding increase in sample size) can increase the power of the test, making it more likely to detect a truly significant effect. However, if your degrees of freedom are limited (as they often are in practice), your test may not be powerful enough to detect a real effect.

Misinterpretation: The p-value only indicates the probability of getting the observed data (or something more extreme) assuming the null hypothesis is true (i.e., there is no effect or no difference). It doesn't tell you the probability that the null hypothesis is true or the size of the effect. This is a common misunderstanding.

In summary, while degrees of freedom, binary findings, and significance are all key concepts in statistical analysis, the relationship between them can be complex and nuanced, leading to potential issues in interpretation and explanation.

\textbf{[User]} Explain wrt to the often-used comparison to pirates.

\textbf{[Assistant]} I assume you're referring to the often-used humorous example about the number of pirates and global warming. The idea is that if you plot the global average temperature against the estimated number of pirates over the last few centuries, you'll see that as the number of pirates has decreased, the global temperature has increased. Thus, one might naively conclude that decreasing the number of pirates causes global warming.

This, of course, is a misunderstanding of the concept of correlation and causation in statistics. Just because two variables are correlated (i.e., they move together), it does not necessarily mean one causes the other. The pirate example illustrates this concept vividly.

This argument is often used to highlight that correlation does not imply causation, and to point out that sometimes the statistical significance found in a model could merely be coincidental or due to an omitted variable that is actually influencing the outcome.

In the context of a binary outcome (pirates or no pirates) and the global temperature, the "degrees of freedom" would theoretically be unlimited because there are numerous factors (variables) that can influence global temperature.

However, when you run a statistical test to see if the presence or absence of pirates significantly impacts global temperature (a binary logistic regression), you might find a statistically significant p-value due to the high correlation. But in reality, the correlation between pirates and global temperature is likely spurious (i.e., coincidental), and the "significant" finding does not provide any evidence of a causal relationship.

It's a playful example that highlights the importance of good statistical reasoning, understanding the underlying assumptions of your models, and the dangers of over-interpreting statistical significance.

\textbf{[User]} Place this in the context of a encryption algorithm with a known input and measured output, such as any strong encoding scheme with a key. This is not a binary result.

\textbf{[Assistant]} Encryption algorithms transform plain text into a ciphered message using a key. With a known input (plain text and key) and measured output (ciphered text), the goal of encryption is not a binary result, but rather to maintain the secrecy and integrity of the transmitted data.

Let's understand this with respect to degrees of freedom and significance.

Degrees of Freedom: In the context of an encryption algorithm, the degrees of freedom could be interpreted as the range of possible outputs that the algorithm could produce for a given input, given a specific key. For a strong encryption algorithm, this would ideally be extremely large even with minor changes in input or key, making it nearly impossible to predict the output without knowing the exact key.

Significance: The concept of significance in the context of encryption might not align with its traditional statistical meaning. In encryption, a significant result could be considered as one where the ciphered text is uncorrelated with the plain text and key. In other words, even small changes in the input or key should result in large, unpredictable changes in the output, a property known as the avalanche effect.

From this perspective, an encryption algorithm with high degrees of freedom and significant unpredictability would be considered strong and secure. If you could predict the output with a certain level of accuracy based on the input, that would indicate a weakness in the encryption algorithm, as it would imply a potential attacker could guess the key or decipher the text.

Now, if you were to compare the properties of different encryption algorithms (binary or not), you might measure "significance" in terms of their ability to resist various types of attacks (such as brute force attacks, side-channel attacks, etc.). "Degrees of freedom" could refer to the number of possible keys or the complexity of the algorithm. However, it's important to note that these terms are being used somewhat metaphorically, as the traditional statistical concepts of significance and degrees of freedom don't directly apply to encryption algorithms.

\textbf{[User]} Can you contrast the weakness of a "number of Pirates" DoF correlation without causation with the strength of a encryption algorithm where the only "cause" is obtaining the correct decoding key?

\textbf{[Assistant]} The concepts of correlation, causation, and degrees of freedom in the context of statistical analyses and encryption algorithms are quite different, but can be contrasted to understand their respective strengths and weaknesses.

In the "number of pirates" correlation example, the argument is made in jest to highlight the fallacy of confusing correlation with causation. The number of pirates decreasing while global temperatures increase is simply a spurious correlation - there's no meaningful causative relationship. Here, the degrees of freedom would refer to the various factors that might affect global temperatures, and "pirates" is just one possible variable among many. The inherent weakness in this correlation is that pirates don't have any scientifically plausible impact on global temperatures.

In contrast, an encryption algorithm has a direct causative relationship between the input (plaintext and key) and the output (ciphertext). If you have the correct key, you can decrypt the ciphertext back into the original plaintext. It's not a matter of correlation; the correct key directly causes the decryption to occur correctly.

In the context of encryption, degrees of freedom could metaphorically represent the number of possible keys. A strong encryption algorithm has a large key space (i.e., a large number of possible keys), making it practically impossible to guess the key through a brute force attack.

In encryption, the equivalent of a "causal relationship" is the deterministic process by which the encryption algorithm uses the key to transform the plaintext into ciphertext. This is a fundamental strength of encryption - if you have the correct key, you can reliably decrypt the message. In contrast to the "pirates" example, this causative relationship is not only meaningful, but it is the fundamental basis for the operation of encryption algorithms.

In conclusion, the "number of pirates" example is a humorous demonstration of the weakness of false correlations, while the operation of encryption algorithms shows the strength of a true causative relationship - the key "causes" the plaintext to be encrypted or decrypted correctly.
}

\promptbox{Academic / Professional writing.}{
\textbf{[User]} You are an experienced lawyer. So am I.

Write a legal memorandum on the considerations before filing a breach of contract lawsuit.

\textbf{[Assistant]} MEMORANDUM

To: [Your Name]From: ChatGPTDate: June 3, 2023Subject: Considerations Before Filing a Breach of Contract Lawsuit

INTRODUCTION

The purpose of this memo is to identify and discuss key considerations before filing a breach of contract lawsuit. The decision to file such a lawsuit should be taken with due diligence, evaluating the cost-benefit analysis, the likelihood of success, and the strategic implications.

CONSIDERATIONS

Validity of the Contract: The first step in any breach of contract action is to establish the existence and validity of the contract. Review the contract terms and conditions, ensuring that it meets the elements of a contract, i.e., offer, acceptance, consideration, capacity, and legality.

Breach of Contract: Examine whether a breach has indeed occurred. Determine if the party in question failed to perform a duty or obligation set out in the contract. Not all failures to perform will result in a breach that provides grounds for a lawsuit.

Types of Breach: Understand the nature of the breach: Is it a material breach (goes to the heart of the contract) or a minor breach (partial performance)? This distinction is crucial in determining the appropriate remedies and the likelihood of success in court.

Evidence of Breach: It is important to collect and review all evidence that substantiates the claim of a breach. This could include emails, correspondence, invoices, receipts, or any other document that provides proof of non-compliance.

Damages: A breach of contract claim is largely futile without identifiable or measurable damages. Consider whether the breach has resulted in any actual damage. These damages can be compensatory (direct loss), consequential (indirect loss), punitive, or nominal.

Mitigation of Damages: Courts expect the non-breaching party to take reasonable steps to mitigate the damages caused by the breach. Consider what steps have been taken or can be taken to mitigate damages.

Affirmative Defenses: Analyze the likelihood that the defendant may use affirmative defenses such as unconscionability, mistake, duress, or statute of limitations. This requires an examination of the contract and the surrounding circumstances.

Cost-Benefit Analysis: Litigation can be costly and time-consuming. A prudent analysis would involve weighing the expected recovery against the cost of litigation.

Alternative Dispute Resolution: Consider exploring alternatives to litigation such as mediation or arbitration, particularly if these methods are stipulated in the contract. These methods can be less confrontational and more cost-effective.

Strategic Considerations: Apart from legal considerations, the decision to sue may also depend on business or strategic considerations. These can include the impact on the business relationship, reputational risks, and the potential precedents that could be set.

Validity of the Contract: The first step in any breach of contract action is to establish the existence and validity of the contract. Review the contract terms and conditions, ensuring that it meets the elements of a contract, i.e., offer, acceptance, consideration, capacity, and legality.

Breach of Contract: Examine whether a breach has indeed occurred. Determine if the party in question failed to perform a duty or obligation set out in the contract. Not all failures to perform will result in a breach that provides grounds for a lawsuit.

Types of Breach: Understand the nature of the breach: Is it a material breach (goes to the heart of the contract) or a minor breach (partial performance)? This distinction is crucial in determining the appropriate remedies and the likelihood of success in court.

Evidence of Breach: It is important to collect and review all evidence that substantiates the claim of a breach. This could include emails, correspondence, invoices, receipts, or any other document that provides proof of non-compliance.

Damages: A breach of contract claim is largely futile without identifiable or measurable damages. Consider whether the breach has resulted in any actual damage. These damages can be compensatory (direct loss), consequential (indirect loss), punitive, or nominal.

Mitigation of Damages: Courts expect the non-breaching party to take reasonable steps to mitigate the damages caused by the breach. Consider what steps have been taken or can be taken to mitigate damages.

Affirmative Defenses: Analyze the likelihood that the defendant may use affirmative defenses such as unconscionability, mistake, duress, or statute of limitations. This requires an examination of the contract and the surrounding circumstances.

Cost-Benefit Analysis: Litigation can be costly and time-consuming. A prudent analysis would involve weighing the expected recovery against the cost of litigation.

Alternative Dispute Resolution: Consider exploring alternatives to litigation such as mediation or arbitration, particularly if these methods are stipulated in the contract. These methods can be less confrontational and more cost-effective.

Strategic Considerations: Apart from legal considerations, the decision to sue may also depend on business or strategic considerations. These can include the impact on the business relationship, reputational risks, and the potential precedents that could be set.

CONCLUSION

In conclusion, the decision to file a breach of contract lawsuit involves a careful analysis of multiple factors. It is crucial to ensure the breach has resulted in damages, the potential for recovery outweighs the cost of litigation, and all evidence has been properly collected and preserved. Also, consider the possibility of defenses and alternatives to litigation. Ultimately, the decision should be a strategic one, balancing legal and business considerations.

This memo provides a generalized overview and should not be construed as legal advice. For advice on a specific legal matter, consult a qualified attorney.

Kind Regards,

ChatGPT

\textbf{[User]} Expand each numbered item above into 2 or more paragraphs.

\textbf{[Assistant]} MEMORANDUM

To: [Your Name]From: ChatGPTDate: June 3, 2023Subject: Considerations Before Filing a Breach of Contract Lawsuit

INTRODUCTION

This memorandum has been created to elaborate on the key factors that need careful consideration before filing a breach of contract lawsuit. The decision to proceed with legal action should be based on a meticulous evaluation of several facets including the probability of success, associated costs, and strategic implications.

CONSIDERATIONS

Validity of the Contract: The foremost consideration before filing a breach of contract lawsuit is to ascertain the existence and validity of the contract. It is important to review the contract meticulously, ensuring that it satisfies the required elements of a legally enforceable contract - offer, acceptance, consideration, capacity, and legality. In addition, the terms and conditions of the contract must be clear, definite, and unambiguous. If any of these elements are lacking or the terms are vague or uncertain, the contract may be deemed invalid, and the lawsuit is unlikely to succeed.

Breach of Contract: A key element in a breach of contract case is demonstrating that a breach has indeed occurred. One must determine whether the party in question has failed to perform a duty or obligation stipulated in the contract. It's important to remember that not all failures to perform will result in a breach that provides grounds for a lawsuit. Certain circumstances may excuse performance, such as an agreement between the parties, force majeure, or impossibility of performance.

Types of Breach: It is also essential to identify and understand the nature of the breach. There are primarily two types of breaches - material and minor. A material breach goes to the heart of the contract, fundamentally altering its purpose, while a minor breach involves partial performance or a slight deviation from the agreed-upon terms. This distinction is critical as it determines the remedies available to the aggrieved party and the likely outcomes in court.

Evidence of Breach: Gathering and evaluating all evidence that substantiates the claim of a breach is another vital consideration. Evidence may include, but is not limited to, emails, correspondence, invoices, receipts, or any other document that provides proof of non-compliance. Witness statements may also be useful in some instances. The absence of solid and persuasive evidence makes it difficult to prove the breach and consequently, weakens the lawsuit.

Damages: A breach of contract claim necessitates identifiable or measurable damages. The party alleging the breach must be able to show that they suffered harm as a result of the breach. These damages can be compensatory (direct loss), consequential (indirect loss), punitive, or nominal. It is important to quantify these damages accurately, as they will form the basis of any monetary relief sought through the lawsuit.

Mitigation of Damages: Courts typically expect the non-breaching party to take reasonable steps to mitigate the damages caused by the breach. This principle of law dictates that the injured party cannot simply sit idle and allow the damages to accumulate. This could mean finding an alternate supplier in case of a supply contract breach or finding another buyer in case of a purchase contract breach. Any inability or unwillingness to mitigate damages could potentially reduce the amount of damages recoverable.

Affirmative Defenses: Potential defenses that the breaching party might employ should also be considered. Affirmative defenses such as unconscionability, mistake, duress, or statute of limitations could be raised by the defendant. This calls for a thorough examination of the contract and the circumstances surrounding its formation and performance, to anticipate possible defenses and develop counterarguments.

Cost-Benefit Analysis: Litigation is often a costly and time-consuming process. Therefore,

Validity of the Contract: The foremost consideration before filing a breach of contract lawsuit is to ascertain the existence and validity of the contract. It is important to review the contract meticulously, ensuring that it satisfies the required elements of a legally enforceable contract - offer, acceptance, consideration, capacity, and legality. In addition, the terms and conditions of the contract must be clear, definite, and unambiguous. If any of these elements are lacking or the terms are vague or uncertain, the contract may be deemed invalid, and the lawsuit is unlikely to succeed.

Breach of Contract: A key element in a breach of contract case is demonstrating that a breach has indeed occurred. One must determine whether the party in question has failed to perform a duty or obligation stipulated in the contract. It's important to remember that not all failures to perform will result in a breach that provides grounds for a lawsuit. Certain circumstances may excuse performance, such as an agreement between the parties, force majeure, or impossibility of performance.

Types of Breach: It is also essential to identify and understand the nature of the breach. There are primarily two types of breaches - material and minor. A material breach goes to the heart of the contract, fundamentally altering its purpose, while a minor breach involves partial performance or a slight deviation from the agreed-upon terms. This distinction is critical as it determines the remedies available to the aggrieved party and the likely outcomes in court.

Evidence of Breach: Gathering and evaluating all evidence that substantiates the claim of a breach is another vital consideration. Evidence may include, but is not limited to, emails, correspondence, invoices, receipts, or any other document that provides proof of non-compliance. Witness statements may also be useful in some instances. The absence of solid and persuasive evidence makes it difficult to prove the breach and consequently, weakens the lawsuit.

Damages: A breach of contract claim necessitates identifiable or measurable damages. The party alleging the breach must be able to show that they suffered harm as a result of the breach. These damages can be compensatory (direct loss), consequential (indirect loss), punitive, or nominal. It is important to quantify these damages accurately, as they will form the basis of any monetary relief sought through the lawsuit.

Mitigation of Damages: Courts typically expect the non-breaching party to take reasonable steps to mitigate the damages caused by the breach. This principle of law dictates that the injured party cannot simply sit idle and allow the damages to accumulate. This could mean finding an alternate supplier in case of a supply contract breach or finding another buyer in case of a purchase contract breach. Any inability or unwillingness to mitigate damages could potentially reduce the amount of damages recoverable.

Affirmative Defenses: Potential defenses that the breaching party might employ should also be considered. Affirmative defenses such as unconscionability, mistake, duress, or statute of limitations could be raised by the defendant. This calls for a thorough examination of the contract and the circumstances surrounding its formation and performance, to anticipate possible defenses and develop counterarguments.

Cost-Benefit Analysis: Litigation is often a costly and time-consuming process. Therefore,

\textbf{[User]} Continue

\textbf{[Assistant]} a rigorous cost-benefit analysis should be performed prior to filing a lawsuit. This analysis would include a careful review of the expected recovery against the potential costs of litigation, including attorney's fees, court costs, expert witness fees, and other related expenses. It's also crucial to factor in the time commitment required and the potential for business disruption.

Alternative Dispute Resolution: Before resorting to litigation, consider exploring Alternative Dispute Resolution (ADR) methods, such as mediation or arbitration. These alternatives can be less adversarial, more cost-effective, and often quicker. Mediation or arbitration may also be mandated by the contract itself in the form of a dispute resolution clause. Even if not required by the contract, ADR options can be mutually beneficial and lead to a more satisfactory resolution for all parties involved.

Strategic Considerations: The decision to pursue a breach of contract lawsuit extends beyond legal considerations. Strategic and business implications must be evaluated. This includes assessing the impact of litigation on the relationship between the parties, reputational risks, potential effects on other business operations, and the future precedent that could be set by taking such action. The decision should balance the need for legal recourse with the potential for achieving broader business objectives.

Alternative Dispute Resolution: Before resorting to litigation, consider exploring Alternative Dispute Resolution (ADR) methods, such as mediation or arbitration. These alternatives can be less adversarial, more cost-effective, and often quicker. Mediation or arbitration may also be mandated by the contract itself in the form of a dispute resolution clause. Even if not required by the contract, ADR options can be mutually beneficial and lead to a more satisfactory resolution for all parties involved.

Strategic Considerations: The decision to pursue a breach of contract lawsuit extends beyond legal considerations. Strategic and business implications must be evaluated. This includes assessing the impact of litigation on the relationship between the parties, reputational risks, potential effects on other business operations, and the future precedent that could be set by taking such action. The decision should balance the need for legal recourse with the potential for achieving broader business objectives.

CONCLUSION

In conclusion, filing a breach of contract lawsuit requires careful analysis of numerous factors. It is essential to verify that a breach has resulted in damages, the potential for recovery outweighs the cost of litigation, and that all evidence has been properly collected and preserved. Consideration of possible defenses and alternatives to litigation is equally important. The ultimate decision should balance legal considerations with broader strategic implications.

Remember, this memorandum provides a general overview and should not be construed as legal advice. For advice on a specific legal matter, consult a qualified attorney.

Kind Regards,

ChatGPT

\textbf{[User]} Rewrite the above. Use these tips:
Avoid passive voice.
Write in clear and concise sentences.
Use strong and specific verbs.
Vary your sentence structure to maintain reader interest.
Show, don't tell. Use descriptive language to create vivid images.
Use active verbs to bring your writing to life.
Avoid using clichés and overused phrases.
Use concrete examples and evidence to support your arguments.
Break down complex ideas into smaller, digestible parts.
Eliminate unnecessary words and phrases to make your writing more concise.
Use descriptive adjectives and adverbs sparingly
Structure your writing with a clear introduction, body, and conclusion.

Now rewrite.

\textbf{[Assistant]} MEMORANDUM

To: [Your Name]From: ChatGPTDate: June 3, 2023Subject: Assessing Crucial Factors for Breach of Contract Lawsuits

INTRODUCTION

This memorandum delves into the multifaceted nature of preparing for a breach of contract lawsuit. It underscores the need for careful evaluation of potential litigation success, estimated costs, and strategic implications.

CONSIDERATIONS

Contract Validity: Before filing a lawsuit, you must thoroughly confirm the contract's existence and legality. The contract needs to satisfy key requirements like a clearly stated offer and acceptance, fair consideration, parties' capacity, and legal terms. Ambiguity in terms or missing elements could invalidate the contract and derail your lawsuit.

Identifying the Breach: Pinpoint the party's failure in performing a contractual obligation. Keep in mind, not every performance failure amounts to a breach. Instances of mutual agreement, force majeure, or performance impossibility can exonerate non-performance.

Understanding the Breach Type: Determine the breach's nature-material or minor. A material breach drastically changes the contract's core purpose, while a minor breach denotes a slight deviation from the agreed terms. Recognizing this distinction assists in identifying suitable remedies and gauging probable court outcomes.

Procuring Evidence: Collect compelling evidence supporting your breach claim. Documentation-emails, invoices, receipts-and witness statements can prove non-compliance. Lack of concrete evidence can weaken your lawsuit.

Quantifying Damages: You must show quantifiable harm resulting from the breach for your claim. Damages could be compensatory (direct loss), consequential (indirect loss), punitive, or nominal. Accurately estimating these damages is crucial for defining your monetary relief.

Mitigating Damages: Courts expect the non-breaching party to actively minimize the breach's damages. If a supplier breaches a contract, finding an alternate supplier illustrates damage mitigation. Failure to mitigate could reduce your recoverable damages.

Predicting Affirmative Defenses: Anticipate defenses that the breaching party might assert. Defenses can include unconscionability, mistake, duress, or statute of limitations. An in-depth review of the contract's formation and performance can help anticipate these defenses and prepare counterarguments.

Conducting Cost-Benefit Analysis: Assess the projected recovery against litigation's potential costs-attorney's fees, court costs, expert witness fees, and others. Remember to account for the time commitment required and possible business disruptions.

Exploring Alternative Dispute Resolution: Consider mediation or arbitration as less adversarial, cost-effective, and faster alternatives to litigation. Contracts often require these methods as part of their dispute resolution clause. ADR can facilitate mutually beneficial resolutions.

Weighing Strategic Considerations: The decision to litigate should not only be legally sound but also strategically beneficial. Consider potential repercussions on business relations, reputation risks, impacts on other business operations, and the future precedent your action may set.

Contract Validity: Before filing a lawsuit, you must thoroughly confirm the contract's existence and legality. The contract needs to satisfy key requirements like a clearly stated offer and acceptance, fair consideration, parties' capacity, and legal terms. Ambiguity in terms or missing elements could invalidate the contract and derail your lawsuit.

Identifying the Breach: Pinpoint the party's failure in performing a contractual obligation. Keep in mind, not every performance failure amounts to a breach. Instances of mutual agreement, force majeure, or performance impossibility can exonerate non-performance.

Understanding the Breach Type: Determine the breach's nature-material or minor. A material breach drastically changes the contract's core purpose, while a minor breach denotes a slight deviation from the agreed terms. Recognizing this distinction assists in identifying suitable remedies and gauging probable court outcomes.

Procuring Evidence: Collect compelling evidence supporting your breach claim. Documentation-emails, invoices, receipts-and witness statements can prove non-compliance. Lack of concrete evidence can weaken your lawsuit.

Quantifying Damages: You must show quantifiable harm resulting from the breach for your claim. Damages could be compensatory (direct loss), consequential (indirect loss), punitive, or nominal. Accurately estimating these damages is crucial for defining your monetary relief.

Mitigating Damages: Courts expect the non-breaching party to actively minimize the breach's damages. If a supplier breaches a contract, finding an alternate supplier illustrates damage mitigation. Failure to mitigate could reduce your recoverable damages.

Predicting Affirmative Defenses: Anticipate defenses that the breaching party might assert. Defenses can include unconscionability, mistake, duress, or statute of limitations. An in-depth review of the contract's formation and performance can help anticipate these defenses and prepare counterarguments.

Conducting Cost-Benefit Analysis: Assess the projected recovery against litigation's potential costs-attorney's fees, court costs, expert witness fees, and others. Remember to account for the time commitment required and possible business disruptions.

Exploring Alternative Dispute Resolution: Consider mediation or arbitration as less adversarial, cost-effective, and faster alternatives to litigation. Contracts often require these methods as part of their dispute resolution clause. ADR can facilitate mutually beneficial resolutions.

Weighing Strategic Considerations: The decision to litigate should not only be legally sound but also strategically beneficial. Consider potential repercussions on business relations, reputation risks, impacts on other business operations, and the future precedent your action may set.

CONCLUSION

In sum, preparing to file a breach of contract lawsuit necessitates thorough scrutiny of the breach, damages, cost-benefit ratio, evidence integrity, defense anticipation, and strategic implications. The final decision must harmonize legal rectitude with broader strategic aims.

This memorandum offers a general perspective, not legal advice. Always consult a qualified attorney for specific legal counsel.

Kind Regards,

ChatGPT
}

\promptbox{Planning}{
\textbf{[User]} Im <REDACTED> and want to plan a trip to <REDACTED>

\textbf{[Assistant]} That sounds like an exciting adventure! <REDACTED> offers a variety of attractions, from its natural landscapes to its bustling cities. What interests you the most: outdoor activities, cultural experiences, or something else? This will help me tailor some recommendations for your trip.

\textbf{[User]} Cultural experiences

\textbf{[Assistant]} Great! For a cultural experience in <REDACTED>, you might consider the following:

<REDACTED>'s Old Market - A historic area with unique shops, restaurants, and cobblestone streets.

The Durham Museum - Located in <REDACTED>, offers a look at the region's history and culture.

The Joslyn Art Museum - Also in <REDACTED>, features a comprehensive collection of art.

Nebraska State Capitol - In Lincoln, this architectural marvel offers guided tours that delve into Nebraska's history.

The Sheldon Museum of Art - Located in <REDACTED>, houses an impressive collection of <REDACTED> art.

Stuhr Museum - In <REDACTED>, gives you a look at pioneer life with historical reenactments and exhibits.

<REDACTED> cultural sites - The state is home to several reservations and cultural centers where you can learn about indigenous history and culture.

Jazz in June - If you happen to visit in June, this music festival in Lincoln offers free outdoor concerts.

Czech Days in <REDACTED> <REDACTED> culture with music, dance, and food.

<REDACTED>'s Old Market - A historic area with unique shops, restaurants, and cobblestone streets.

The Durham Museum - Located in <REDACTED>, offers a look at the region's history and culture.

The Joslyn Art Museum - Also in <REDACTED>, features a comprehensive collection of art.

Nebraska State Capitol - In Lincoln, this architectural marvel offers guided tours that delve into Nebraska's history.

The Sheldon Museum of Art - Located in <REDACTED>, houses an impressive collection of <REDACTED> art.

Stuhr Museum - In <REDACTED>, gives you a look at pioneer life with historical reenactments and exhibits.

<REDACTED> cultural sites - The state is home to several reservations and cultural centers where you can learn about indigenous history and culture.

Jazz in June - If you happen to visit in June, this music festival in Lincoln offers free outdoor concerts.

Czech Days in <REDACTED> <REDACTED> culture with music, dance, and food.

Each city and attraction will give you a different perspective on Nebraskan and <REDACTED> culture. Would you like more information on any of these?

\textbf{[User]} How does that relate to me as a <REDACTED>?

\textbf{[Assistant]} As a <REDACTED>, experiencing cultural aspects of Nebraska could offer a fascinating contrast to your own cultural background. You'd have the opportunity to explore how <REDACTED> culture has evolved, and how it intersects with or diverges from <REDACTED> culture.

Music: While Nebraska doesn't have reggae or dancehall scenes, it does have a rich musical culture that includes jazz and blues, which have influenced music globally, including in <REDACTED>.

Community Events: Attending local events like festivals can be a fun way to engage with the community and compare cultural celebrations to your own.

Indigenous Culture: Learning about the <REDACTED> heritage could offer interesting parallels to <REDACTED>'s own indigenous and colonial history.

Food: Trying local cuisine, like Nebraska's famous beef, could be a gastronomic adventure to compare with <REDACTED> specialties like jerk chicken or ackee and <REDACTED>.

Music: While Nebraska doesn't have reggae or dancehall scenes, it does have a rich musical culture that includes jazz and blues, which have influenced music globally, including in <REDACTED>.

Community Events: Attending local events like festivals can be a fun way to engage with the community and compare cultural celebrations to your own.

Indigenous Culture: Learning about the <REDACTED> heritage could offer interesting parallels to <REDACTED>'s own indigenous and colonial history.

Food: Trying local cuisine, like Nebraska's famous beef, could be a gastronomic adventure to compare with <REDACTED> specialties like jerk chicken or ackee and <REDACTED>.

The trip could serve as a cultural exchange of sorts, giving you a chance to share your own <REDACTED> heritage while immersing yourself in a different environment.

\textbf{[User]} Give me a conversational example. Speak as me talking with a native nebraskan

\textbf{[Assistant]} Certainly! Imagine you're at a local café in <REDACTED>, <REDACTED>, and you strike up a conversation with someone sitting next to you. Here's how it might go:

You: "Hey there, the coffee smells amazing, doesn't it? I'm visiting from <REDACTED> and trying to get a taste of the local scene. Any recommendations?"

<REDACTED>: "Oh, welcome to <REDACTED>! Yeah, the coffee here is pretty good. If you're looking for a cultural experience, you should check out the Old Market. Lots of history and unique shops there."

You: "Ah, the Old Market sounds interesting! Back home in <REDACTED>, we have vibrant markets too, full of spices, fruits, and crafts. Is it similar?"

<REDACTED>: "Well, it's a bit different-more focused on art galleries, boutiques, and restaurants. But it's definitely a must-see. You get to feel the pulse of the city."

You: "I'm all for soaking up local culture. You know, in <REDACTED>, we've got a rich musical history with reggae and dancehall. What's the music scene like here?"

<REDACTED>: "We've got a diverse range of music, but jazz and blues are pretty big. There's even a Jazz in June festival in Lincoln if you're around then."

You: "Sounds like something I'd enjoy! Music really has a way of bringing people together, doesn't it? Just like our annual Reggae Sumfest in <REDACTED>."

<REDACTED>: "Absolutely, music is universal. If you're interested in history, don't miss the Durham Museum. It gives you a great overview of the region."

You: "I appreciate the tip!
}

\promptbox{Data analysis}
{
\textbf{[User]} Explain the difficulty of explaining a binary finding in terms of significance wrt degrees of freedom.

\textbf{[Assistant]} When dealing with a binary outcome or finding (i.e., a situation where there are only two possible results, such as "yes" or "no", "success" or "failure", etc.), the main statistical techniques used are based on binary or logistic regression models. One of the key aspects of these models is the concept of "degrees of freedom."

Degrees of freedom in a statistical model represent the number of values in a statistical calculation that have the freedom to vary. In other words, they are the number of independent pieces of information that go into the calculation of a statistic.

In a binary logistic regression, for example, the degrees of freedom typically equal the number of predictor variables you have in your model. The concept is related to the complexity of your model: the more predictor variables you have, the more complex your model is and the more degrees of freedom it has.

Now, the concept of "significance" refers to the probability that the observed relationship (e.g., between predictor variables and the binary outcome) occurred by chance. If the significance value (often denoted as 'p') is below a certain threshold (commonly 0.05), we say the relationship is statistically significant.

The difficulty in explaining a binary finding in terms of significance with respect to degrees of freedom arises from a few points:

Overfitting: If you have too many degrees of freedom (i.e., too many predictor variables) compared to the number of observations, your model might fit the data too closely. This can result in overfitting, where the model works well for the existing data but may not predict new data well. Overfit models can give misleadingly small p-values, suggesting significance where there may be none.

Underfitting: On the other hand, if you have too few degrees of freedom, your model might not capture the complexity of the data, resulting in underfitting. This could result in larger p-values, possibly failing to detect significant relationships.

Multicollinearity: This is when predictor variables in a model are highly correlated. This can inflate the variance of the coefficient estimates and make the estimates very sensitive to minor changes in the model. The result is that the model may indicate that a predictor is not statistically significant, even though it is.

Power of the Test: The power of a statistical test is the probability that it will reject a false null hypothesis. A higher number of degrees of freedom (with a corresponding increase in sample size) can increase the power of the test, making it more likely to detect a truly significant effect. However, if your degrees of freedom are limited (as they often are in practice), your test may not be powerful enough to detect a real effect.

Misinterpretation: The p-value only indicates the probability of getting the observed data (or something more extreme) assuming the null hypothesis is true (i.e., there is no effect or no difference). It doesn't tell you the probability that the null hypothesis is true or the size of the effect. This is a common misunderstanding.

Overfitting: If you have too many degrees of freedom (i.e., too many predictor variables) compared to the number of observations, your model might fit the data too closely. This can result in overfitting, where the model works well for the existing data but may not predict new data well. Overfit models can give misleadingly small p-values, suggesting significance where there may be none.

Underfitting: On the other hand, if you have too few degrees of freedom, your model might not capture the complexity of the data, resulting in underfitting. This could result in larger p-values, possibly failing to detect significant relationships.

Multicollinearity: This is when predictor variables in a model are highly correlated. This can inflate the variance of the coefficient estimates and make the estimates very sensitive to minor changes in the model. The result is that the model may indicate that a predictor is not statistically significant, even though it is.

Power of the Test: The power of a statistical test is the probability that it will reject a false null hypothesis. A higher number of degrees of freedom (with a corresponding increase in sample size) can increase the power of the test, making it more likely to detect a truly significant effect. However, if your degrees of freedom are limited (as they often are in practice), your test may not be powerful enough to detect a real effect.

Misinterpretation: The p-value only indicates the probability of getting the observed data (or something more extreme) assuming the null hypothesis is true (i.e., there is no effect or no difference). It doesn't tell you the probability that the null hypothesis is true or the size of the effect. This is a common misunderstanding.

In summary, while degrees of freedom, binary findings, and significance are all key concepts in statistical analysis, the relationship between them can be complex and nuanced, leading to potential issues in interpretation and explanation.

\textbf{[User]} Explain wrt to the often-used comparison to pirates.

\textbf{[Assistant]} I assume you're referring to the often-used humorous example about the number of pirates and global warming. The idea is that if you plot the global average temperature against the estimated number of pirates over the last few centuries, you'll see that as the number of pirates has decreased, the global temperature has increased. Thus, one might naively conclude that decreasing the number of pirates causes global warming.

This, of course, is a misunderstanding of the concept of correlation and causation in statistics. Just because two variables are correlated (i.e., they move together), it does not necessarily mean one causes the other. The pirate example illustrates this concept vividly.

This argument is often used to highlight that correlation does not imply causation, and to point out that sometimes the statistical significance found in a model could merely be coincidental or due to an omitted variable that is actually influencing the outcome.

In the context of a binary outcome (pirates or no pirates) and the global temperature, the "degrees of freedom" would theoretically be unlimited because there are numerous factors (variables) that can influence global temperature.

However, when you run a statistical test to see if the presence or absence of pirates significantly impacts global temperature (a binary logistic regression), you might find a statistically significant p-value due to the high correlation. But in reality, the correlation between pirates and global temperature is likely spurious (i.e., coincidental), and the "significant" finding does not provide any evidence of a causal relationship.

It's a playful example that highlights the importance of good statistical reasoning, understanding the underlying assumptions of your models, and the dangers of over-interpreting statistical significance.

\textbf{[User]} Place this in the context of a encryption algorithm with a known input and measured output, such as any strong encoding scheme with a key. This is not a binary result.

\textbf{[Assistant]} Encryption algorithms transform plain text into a ciphered message using a key. With a known input (plain text and key) and measured output (ciphered text), the goal of encryption is not a binary result, but rather to maintain the secrecy and integrity of the transmitted data.

Let's understand this with respect to degrees of freedom and significance.

Degrees of Freedom: In the context of an encryption algorithm, the degrees of freedom could be interpreted as the range of possible outputs that the algorithm could produce for a given input, given a specific key. For a strong encryption algorithm, this would ideally be extremely large even with minor changes in input or key, making it nearly impossible to predict the output without knowing the exact key.

Significance: The concept of significance in the context of encryption might not align with its traditional statistical meaning. In encryption, a significant result could be considered as one where the ciphered text is uncorrelated with the plain text and key. In other words, even small changes in the input or key should result in large, unpredictable changes in the output, a property known as the avalanche effect.

From this perspective, an encryption algorithm with high degrees of freedom and significant unpredictability would be considered strong and secure. If you could predict the output with a certain level of accuracy based on the input, that would indicate a weakness in the encryption algorithm, as it would imply a potential attacker could guess the key or decipher the text.

Now, if you were to compare the properties of different encryption algorithms (binary or not), you might measure "significance" in terms of their ability to resist various types of attacks (such as brute force attacks, side-channel attacks, etc.). "Degrees of freedom" could refer to the number of possible keys or the complexity of the algorithm. However, it's important to note that these terms are being used somewhat metaphorically, as the traditional statistical concepts of significance and degrees of freedom don't directly apply to encryption algorithms.

\textbf{[User]} Can you contrast the weakness of a "number of Pirates" DoF correlation without causation with the strength of a encryption algorithm where the only "cause" is obtaining the correct decoding key?

\textbf{[Assistant]} The concepts of correlation, causation, and degrees of freedom in the context of statistical analyses and encryption algorithms are quite different, but can be contrasted to understand their respective strengths and weaknesses.

In the "number of pirates" correlation example, the argument is made in jest to highlight the fallacy of confusing correlation with causation. The number of pirates decreasing while global temperatures increase is simply a spurious correlation - there's no meaningful causative relationship. Here, the degrees of freedom would refer to the various factors that might affect global temperatures, and "pirates" is just one possible variable among many. The inherent weakness in this correlation is that pirates don't have any scientifically plausible impact on global temperatures.

In contrast, an encryption algorithm has a direct causative relationship between the input (plaintext and key) and the output (ciphertext). If you have the correct key, you can decrypt the ciphertext back into the original plaintext. It's not a matter of correlation; the correct key directly causes the decryption to occur correctly.

In the context of encryption, degrees of freedom could metaphorically represent the number of possible keys. A strong encryption algorithm has a large key space (i.e., a large number of possible keys), making it practically impossible to guess the key through a brute force attack.

In encryption, the equivalent of a "causal relationship" is the deterministic process by which the encryption algorithm uses the key to transform the plaintext into ciphertext. This is a fundamental strength of encryption - if you have the correct key, you can reliably decrypt the message. In contrast to the "pirates" example, this causative relationship is not only meaningful, but it is the fundamental basis for the operation of encryption algorithms.

In conclusion, the "number of pirates" example is a humorous demonstration of the weakness of false correlations, while the operation of encryption algorithms shows the strength of a true causative relationship - the key "causes" the plaintext to be encrypted or decrypted correctly.
}

\promptbox{Planning}{
\textbf{[User]} How many water bodies are there in <REDACTED>, and what are their names and types? You can also browse the web to find out.

\textbf{[Assistant]} <REDACTED>, a region in <REDACTED>, is renowned for its numerous and scenic lakes. Some of the most notable and largest lakes within <REDACTED> include:

Lake Como- Also known as <REDACTED>, this lake is famous for its deep waters and picturesque surroundings, often attracting celebrities and tourists alike.

Lake Garda- The largest lake in <REDACTED>, it stretches across several regions, including <REDACTED>. It is a popular destination for water sports and scenic beauty.

Lake Maggiore- While part of this lake lies within the neighboring region of <REDACTED>, it also forms a significant part of <REDACTED> in <REDACTED>.

Lake Iseo- Known for its clear waters and charming islands, including the largest lake island in <REDACTED>, <REDACTED>.

<REDACTED> Surrounded by rolling hills and quiet towns, this lake offers a more tranquil retreat.

<REDACTED> and less crowded, providing a peaceful escape with opportunities for fishing and windsurfing.

Lake Lugano- Although mainly in <REDACTED>, parts of this lake extend into <REDACTED>, offering stunning views and luxurious resorts.

Lake Como- Also known as <REDACTED>, this lake is famous for its deep waters and picturesque surroundings, often attracting celebrities and tourists alike.

Lake Garda- The largest lake in <REDACTED>, it stretches across several regions, including <REDACTED>. It is a popular destination for water sports and scenic beauty.

Lake Maggiore- While part of this lake lies within the neighboring region of <REDACTED>, it also forms a significant part of <REDACTED> in <REDACTED>.

Lake Iseo- Known for its clear waters and charming islands, including the largest lake island in <REDACTED>, <REDACTED>.

<REDACTED> Surrounded by rolling hills and quiet towns, this lake offers a more tranquil retreat.

<REDACTED> and less crowded, providing a peaceful escape with opportunities for fishing and windsurfing.

Lake Lugano- Although mainly in <REDACTED>, parts of this lake extend into <REDACTED>, offering stunning views and luxurious resorts.

Additional smaller lakes such as <REDACTED>, Lake Pusiano, <REDACTED>, and <REDACTED> also contribute to the region's reputation as a lake district. These lakes vary in size and are popular for both their natural beauty and recreational activities. Each lake has its own unique charm, ranging from bustling tourist hubs with water sports to serene retreats ideal for nature walks and picnics.

These lakes make <REDACTED> a favorite destination for those seeking the beauty and recreational opportunities provided by these expansive water bodiesitalia.

\textbf{[User]} What is the concentration of reactive silicates observed for the lake Endine? You can browse the web if you don't have the answer.

\textbf{[Assistant]} I couldn't find specific information on the concentration of reactive silicates in <REDACTED>. The search results primarily provide general information about the lake's location, environmental status, and activities around it, but do not detail the chemical composition of its waters, including silicates. If you're conducting research or need this data for environmental studies, I recommend contacting local environmental agencies or universities in the area that might conduct regular water quality assessments of <REDACTED>.

\textbf{[User]} What are the observations generated by Sensor 14530? This sensor is in the weather station <REDACTED>, Varano borghi

\textbf{[Assistant]} I wasn't able to find specific data for sensor 14530 at weather station <REDACTED> in <REDACTED>. It seems like detailed observations from individual sensors are not readily available or explicitly listed on the common weather reporting websites. For more precise and specific sensor data, you might need to access specialized meteorological databases or contact the station directly if possible. The available information generally includes broader weather conditions such as temperature, wind speed, and humidity for the area, but not down to the level of individual sensor outputswunderground.

\textbf{[User]} What is the average number of accesses recorded by Hospital Macchi during 2016 for digestive tract diseases?

\textbf{[Assistant]} I wasn't able to find specific data regarding the average number of accesses for digestive tract diseases at <REDACTED> in <REDACTED> during 2016. The information available online about Hospital Macchi generally describes the facility and its services but doesn't provide detailed statistics or annual reports accessible through common search methods. For detailed statistics like these, you might need to contact the hospital directly or access regional health databases that track hospital admissions and treatments for specific conditions.

}
\promptbox{Random or tangential topic (excluded from single-topic task set)}{
\textbf{[User]} How many water bodies are there in <REDACTED>, and what are their names and types? You can also browse the web to find out.

\textbf{[Assistant]} <REDACTED>, a region in <REDACTED>, is renowned for its numerous and scenic lakes. Some of the most notable and largest lakes within <REDACTED> include:

Lake Como- Also known as <REDACTED>, this lake is famous for its deep waters and picturesque surroundings, often attracting celebrities and tourists alike.

Lake Garda- The largest lake in <REDACTED>, it stretches across several regions, including <REDACTED>. It is a popular destination for water sports and scenic beauty.

Lake Maggiore- While part of this lake lies within the neighboring region of <REDACTED>, it also forms a significant part of <REDACTED> in <REDACTED>.

Lake Iseo- Known for its clear waters and charming islands, including the largest lake island in <REDACTED>, <REDACTED>.

<REDACTED> Surrounded by rolling hills and quiet towns, this lake offers a more tranquil retreat.

<REDACTED> and less crowded, providing a peaceful escape with opportunities for fishing and windsurfing.

Lake Lugano- Although mainly in <REDACTED>, parts of this lake extend into <REDACTED>, offering stunning views and luxurious resorts.

Lake Como- Also known as <REDACTED>, this lake is famous for its deep waters and picturesque surroundings, often attracting celebrities and tourists alike.

Lake Garda- The largest lake in <REDACTED>, it stretches across several regions, including <REDACTED>. It is a popular destination for water sports and scenic beauty.

Lake Maggiore- While part of this lake lies within the neighboring region of <REDACTED>, it also forms a significant part of <REDACTED> in <REDACTED>.

Lake Iseo- Known for its clear waters and charming islands, including the largest lake island in <REDACTED>, <REDACTED>.

<REDACTED> Surrounded by rolling hills and quiet towns, this lake offers a more tranquil retreat.

<REDACTED> and less crowded, providing a peaceful escape with opportunities for fishing and windsurfing.

Lake Lugano- Although mainly in <REDACTED>, parts of this lake extend into <REDACTED>, offering stunning views and luxurious resorts.

Additional smaller lakes such as <REDACTED>, Lake Pusiano, <REDACTED>, and <REDACTED> also contribute to the region's reputation as a lake district. These lakes vary in size and are popular for both their natural beauty and recreational activities. Each lake has its own unique charm, ranging from bustling tourist hubs with water sports to serene retreats ideal for nature walks and picnics.

These lakes make <REDACTED> a favorite destination for those seeking the beauty and recreational opportunities provided by these expansive water bodiesitalia.

\textbf{[User]} What is the concentration of reactive silicates observed for the lake Endine? You can browse the web if you don't have the answer.

\textbf{[Assistant]} I couldn't find specific information on the concentration of reactive silicates in <REDACTED>. The search results primarily provide general information about the lake's location, environmental status, and activities around it, but do not detail the chemical composition of its waters, including silicates. If you're conducting research or need this data for environmental studies, I recommend contacting local environmental agencies or universities in the area that might conduct regular water quality assessments of <REDACTED>.

\textbf{[User]} What are the observations generated by Sensor 14530? This sensor is in the weather station <REDACTED>, Varano borghi

\textbf{[Assistant]} I wasn't able to find specific data for sensor 14530 at weather station <REDACTED> in <REDACTED>. It seems like detailed observations from individual sensors are not readily available or explicitly listed on the common weather reporting websites. For more precise and specific sensor data, you might need to access specialized meteorological databases or contact the station directly if possible. The available information generally includes broader weather conditions such as temperature, wind speed, and humidity for the area, but not down to the level of individual sensor outputswunderground.

\textbf{[User]} What is the average number of accesses recorded by Hospital Macchi during 2016 for digestive tract diseases?

\textbf{[Assistant]} I wasn't able to find specific data regarding the average number of accesses for digestive tract diseases at <REDACTED> in <REDACTED> during 2016. The information available online about Hospital Macchi generally describes the facility and its services but doesn't provide detailed statistics or annual reports accessible through common search methods. For detailed statistics like these, you might need to contact the hospital directly or access regional health databases that track hospital admissions and treatments for specific conditions.
}

\begin{figure}[p]
    \centering

    \begin{subfigure}[t]{\linewidth}
        \centering
        \includegraphics[width=\linewidth]{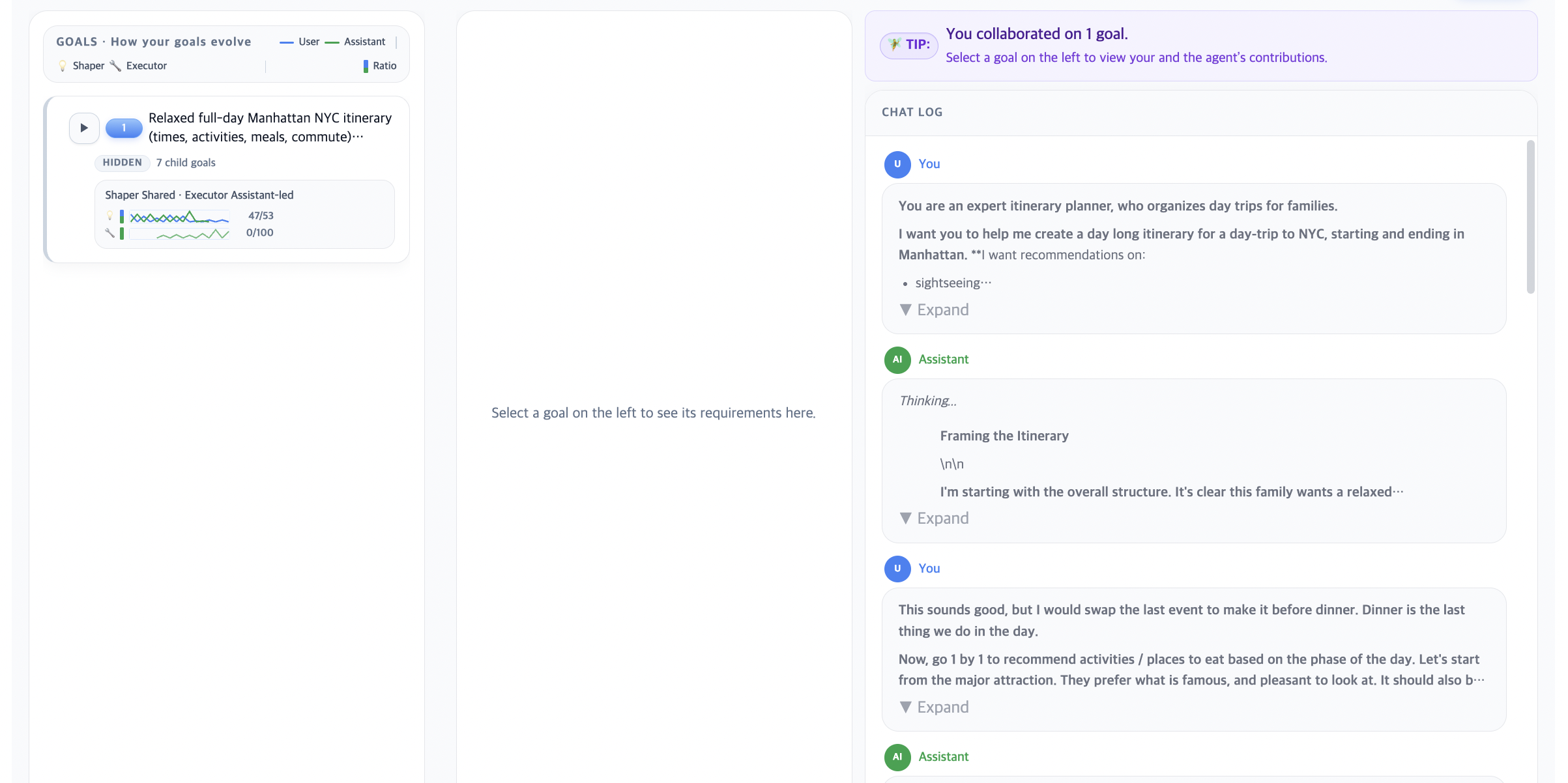}
        \caption{First Screen}
        \label{fig:UI_1}
    \end{subfigure}

    \vspace{1em}

    \begin{subfigure}[t]{\linewidth}
        \centering
        \includegraphics[width=\linewidth]{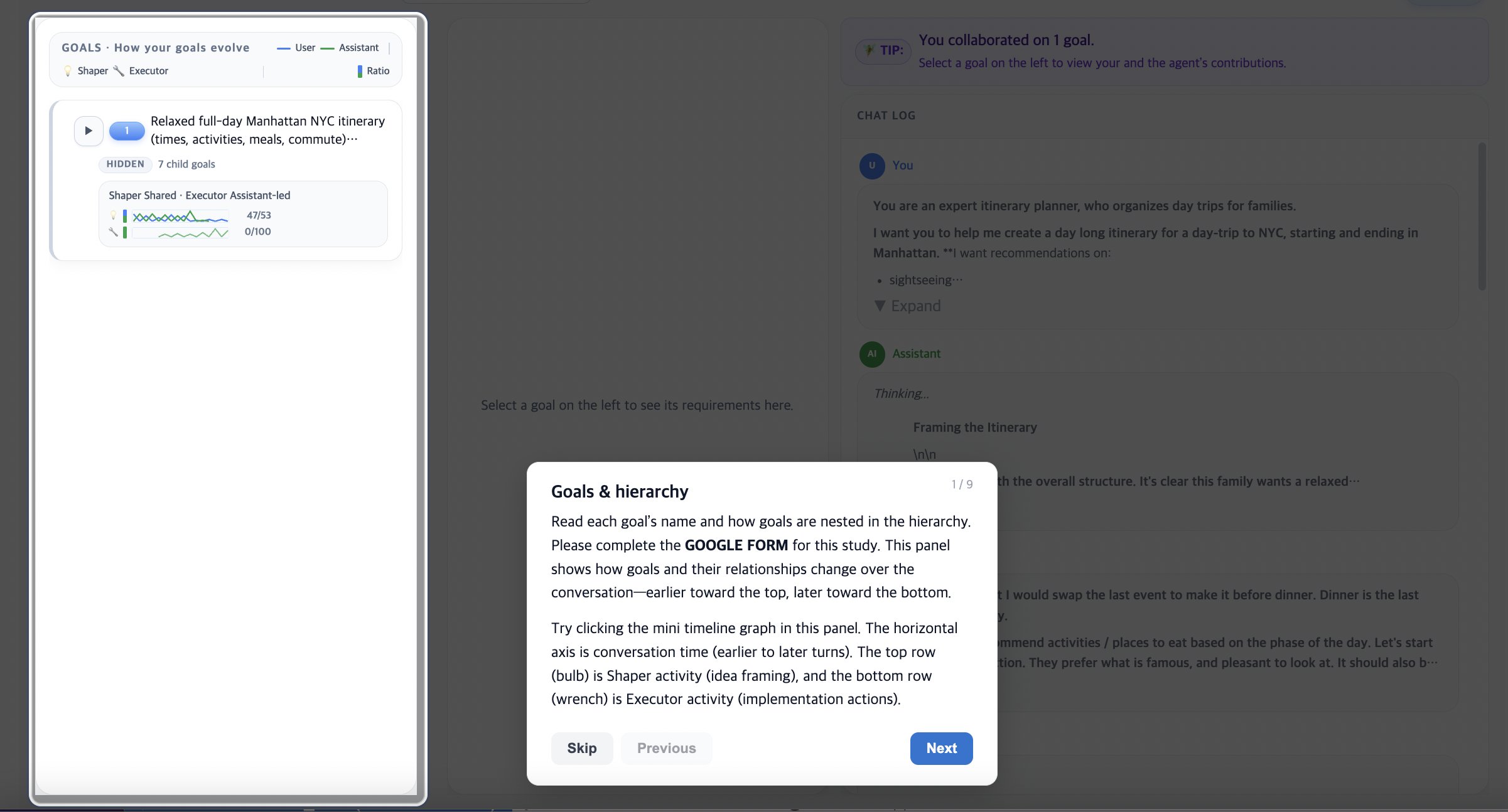}
        \caption{Second Screen}
        \label{fig:UI_2}
    \end{subfigure}

    \caption{Screenshots of UI and Tutorial we used for Human Study}
    \label{fig:UI}
\end{figure}

\clearpage

\begin{figure}[p]\ContinuedFloat
    \centering

    \begin{subfigure}[t]{\linewidth}
        \centering
        \includegraphics[width=\linewidth]{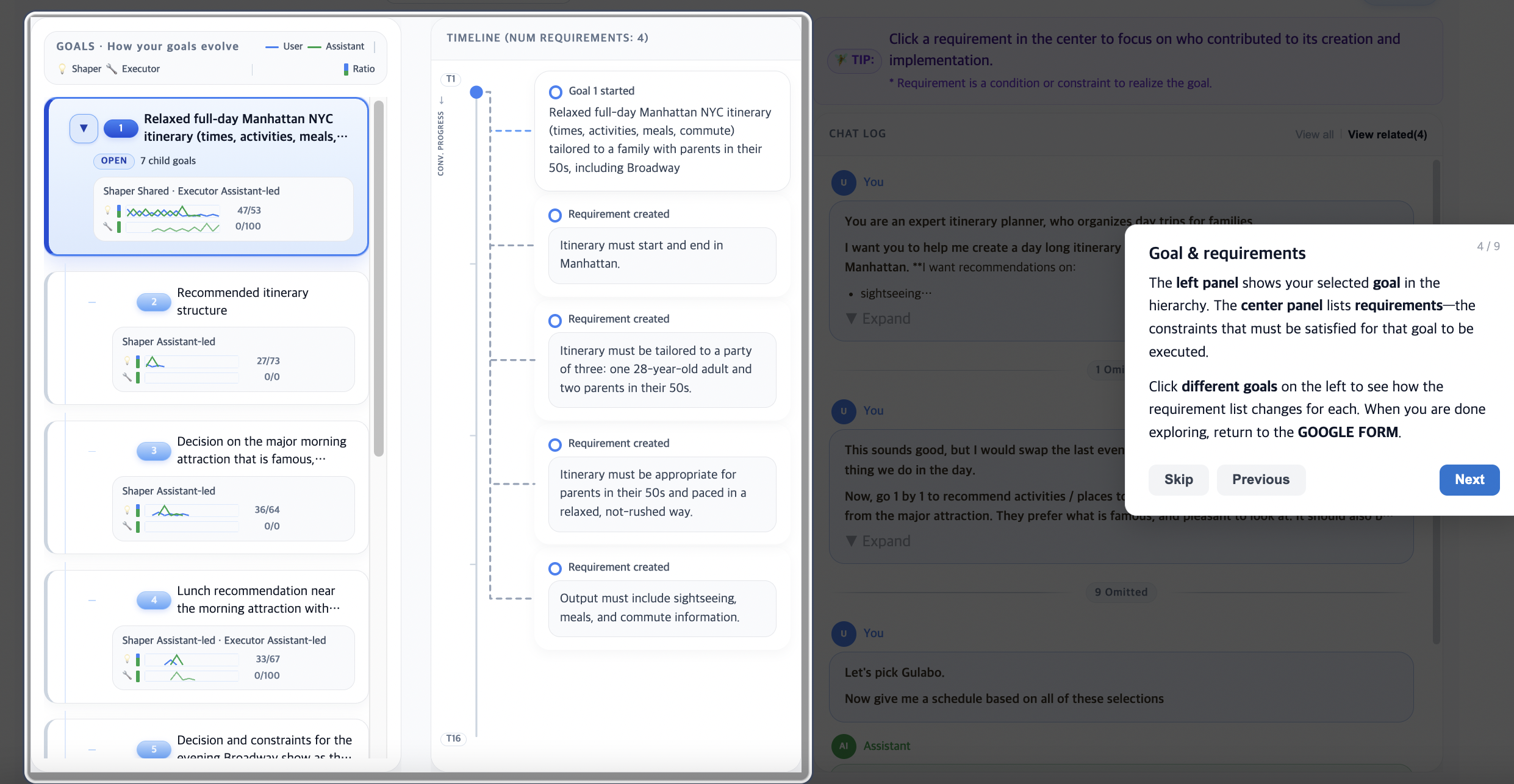}
        \caption{Third Screen}
        \label{fig:UI_3}
    \end{subfigure}

    \vspace{1em}

    \begin{subfigure}[t]{\linewidth}
        \centering
        \includegraphics[width=\linewidth]{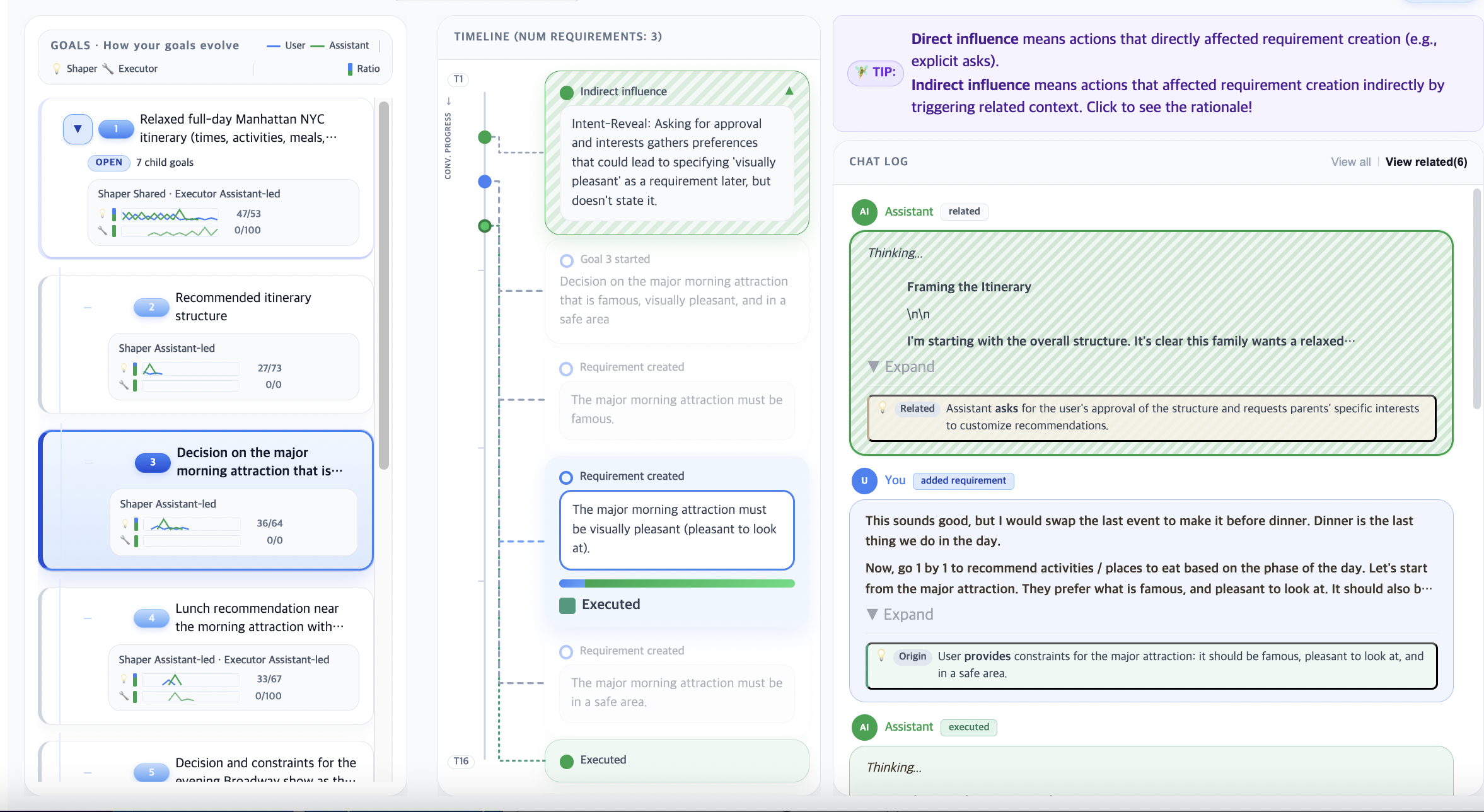}
        \caption{Fourth Screen}
        \label{fig:UI_4}
    \end{subfigure}

    \caption{Screenshots of UI and Tutorial we used for Human Study (continued)}
    \label{fig:UI-2}
\end{figure}

\end{document}